%% file: paper.tex
\documentclass[]{bytedance_seed}

\usepackage[toc,page,header]{appendix}

\usepackage{minitoc}
\usepackage{inconsolata}
\usepackage{url}
\usepackage[linesnumbered,ruled,vlined]{algorithm2e}
\usepackage{multirow}
\usepackage{booktabs}
\usepackage{caption}
\usepackage{subcaption}
\usepackage{xspace}
\usepackage{amsmath}
\usepackage{amssymb}
\usepackage{newtxmath}
\usepackage{bbm}
\usepackage{graphicx}
\usepackage{tabularx}
\usepackage{bbding}
\usepackage{pifont}
\usepackage{diagbox}
\usepackage{makecell}
\usepackage{colortbl}
\usepackage{bbm}
\usepackage{color}
\usepackage{hhline}
\usepackage{soul}
\usepackage{float}
\usepackage[most]{tcolorbox}

\newcommand{\paratitle}[1]{\vspace{1.5ex}\noindent\textbf{#1}}
\newcommand{\ie}{\textit{i.e.,}\xspace}

\newcommand{\eg}{\textit{e.g.,}\xspace}

\newcommand{\ignore}[1]{}

\definecolor{takeaway}{RGB}{165, 209, 216}
\definecolor{takeawayTitle}{RGB}{57, 89, 163}

\title{ Pass@k Training for Adaptively Balancing Exploration and Exploitation of Large Reasoning Models  }

\author[1,2,*]{Zhipeng Chen}
\author[2]{Xiaobo Qin}
\author[2]{Youbin Wu}
\author[2]{Yue Ling}
\author[2]{Qinghao Ye}
\author[1,2, \dagger]{\\Wayne Xin Zhao}
\author[2, \dagger]{Guang Shi}

\affiliation[1]{Renmin University of China}
\affiliation[2]{ByteDance Seed}

\contribution[*]{Work done at ByteDance Seed}
\contribution[\dagger]{Corresponding authors}

\abstract{
Reinforcement learning with verifiable rewards (RLVR), which typically adopts Pass@1 as the reward, has faced the issues in balancing exploration and exploitation, causing policies to prefer conservative actions, converging to a local optimum. Identifying an appropriate reward metric is therefore crucial. Regarding the prior work, although Pass@k has been used in evaluation, its connection to LLM exploration ability in RLVR remains largely overlooked. To investigate this, we first use Pass@k as the reward to train the policy model (\ie \textbf{Pass@k Training}), and observe the improvement on its exploration ability. Next, we derive an analytical solution for the advantage of Pass@k Training, leading to an efficient and effective process. Building on this, our analysis reveals that exploration and exploitation are not inherently conflicting objectives, while they can mutually enhance each other. Moreover, Pass@k Training with analytical derivation essentially involves directly designing the advantage function. Inspired by this, we preliminarily explore the advantage design for RLVR, showing promising results and highlighting a potential future direction.
}

\date{\today}
\correspondence{Zhipeng Chen at \email{zhipeng\_chen@ruc.edu.cn}}

\checkdata[Project Page]{\url{https://github.com/RUCAIBox/Passk_Training}}

\begin{document}
\maketitle

\input{sections/1-introduction}
\input{sections/2-HowPassK}
\input{sections/3-WhatBenefits}
\input{sections/4-WhyPassK}
\input{sections/5-relatedwork}
\input{sections/6-conclusion}

\section*{Acknowledgement}
We sincerely thank Enigmata Team~\cite{30:journals/corr/abs-2505-19914} to provide the training and validation sets of Enigmata and share experiences of RLVR training on logic puzzles. We appreciate Songhua Cai and other contributors of the Seed Infrastructure team for infrastructure support.

\clearpage

\bibliographystyle{plainnat}
\bibliography{main}

\clearpage

\beginappendix

\input{sections/appendix}

\end{document}

%% file: sections/1-introduction.tex
\vspace{-30pt}
\begin{figure}[h]
    \centering
    \includegraphics[width=0.9\linewidth]{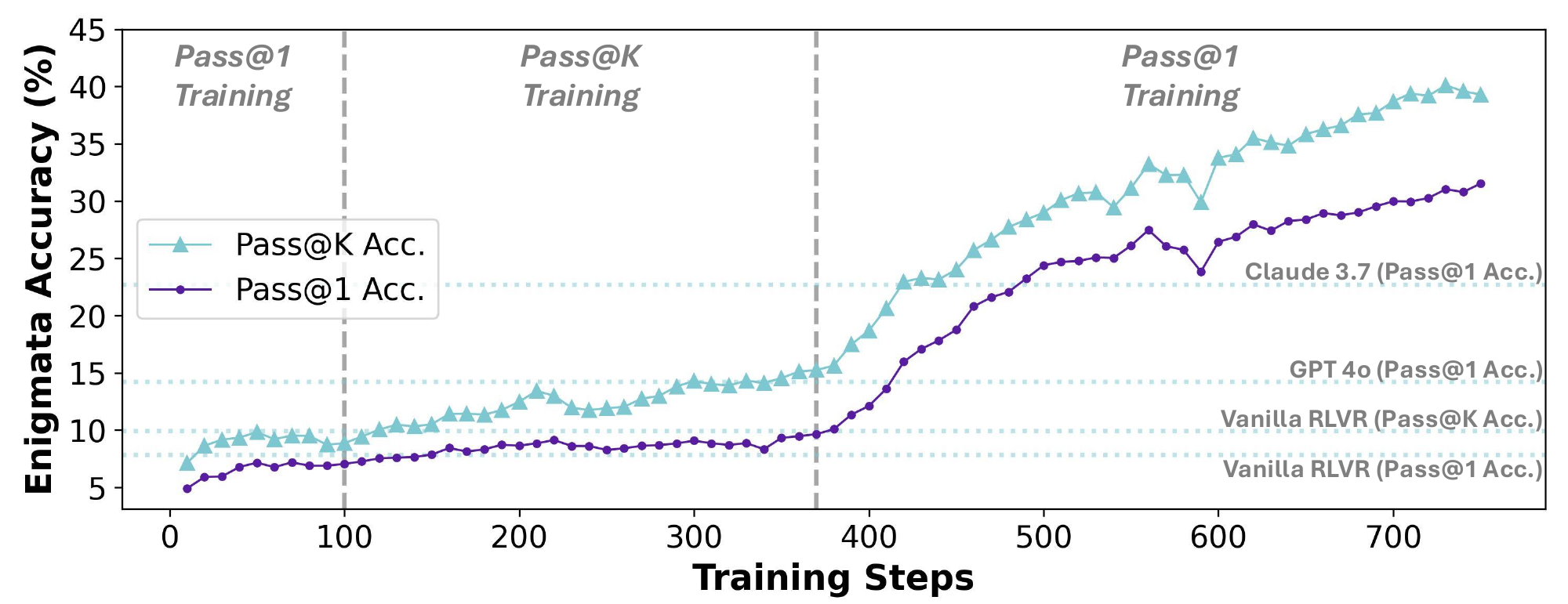}
    \caption{Enigmata scores (Validation Set) of \textbf{Pass@k Training} on Qwen2.5-7B-Ins, which boosts its exploration ability, leading to continuous improvements in following training, surpassing native RLVR and powerful LLMs.}
    \label{fig:overview}
\end{figure}

\newpage

\section{Introduction}
\label{sec:intro}

Recently, reinforcement learning with verifiable rewards (RLVR) has emerged to solve complex reasoning tasks and dramatically boost the reasoning capabilities of large language models (LLMs)~\cite{1:conf/acl/TrungZJSJL24,2:journals/corr/abs-2411-15124,3:journals/corr/abs-2412-16849}. During RLVR training, LLMs generate various responses based on the given prompt and receive rewards for responses~\cite{4:journals/corr/abs-2402-03300,5:journals/corr/abs-2501-12599,6:journals/corr/abs-2505-07062}. LLMs can possess the ability to generate a more comprehensive reasoning process by learning from outcome-level supervision~\cite{7:journals/corr/abs-2501-12948,8:journals/corr/abs-2503-04548}, thereby achieving higher performance on downstream tasks. The success of these large reasoning models (LRMs), such as OpenAI o1~\cite{9:journals/corr/abs-2412-16720} and DeepSeek R1~\cite{7:journals/corr/abs-2501-12948}, suggests that RLVR training pushes the limits of the capacities of LLMs.

The current RLVR training that typically optimizes the Pass@1 objective, also known as \textit{Pass@1 Training}, trains LLMs to learn from their exploration and generate the most confident response for the given prompt~\cite{10:journals/corr/abs-2502-14768,11:journals/corr/abs-2503-18892,12:journals/corr/abs-2503-24290}, leading to a major challenge of the balance of exploration and exploitation~\cite{13:journals/corr/abs-2506-14758,14:journals/corr/abs-2501-11651}. 
Typically, exploration refers to performing novel and various actions~\cite{15:conf/acl/WangLSXDLCWS24}, while exploitation requires LLMs to invoke reliable actions that the verifier prefers among the known behaviours~\cite{16:journals/tmlr/SinghCAAPGLH0XP24}. 
During the Pass@1 Training process, LLMs tend to imitate the behaviours that can bring an increase of reward scores in previous attempts, and prevent the behaviours that receive low rewards~\cite{17:journals/corr/abs-2502-06781,18:conf/acl/ChenZZWZZW24}. 
However, in outcome-supervision, which is the popular Pass@1 Training setting~\cite{7:journals/corr/abs-2501-12948,10:journals/corr/abs-2502-14768,11:journals/corr/abs-2503-18892}, the erroneous solution with correct answer will receive positive rewards, while the correct solution with wrong answer will be assigned negative rewards~\cite{18:conf/acl/ChenZZWZZW24,71:journals/corr/abs-2503-21380}.
In this case, the unsuccessful explorations, which contains the correct idea, will be associated with the high cost, as it often yields no reward, resulting in an imbalance between exploitation and exploration~\cite{19:journals/corr/abs-2505-22617}, might leading policy to abandon exploration and converge on a local optimum~\cite{20:journals/tist/LiLSCYY21,21:journals/corr/abs-2109-00157}. 
Limited by the suboptimal nature of the reward under reinforcement learning methods (\eg PPO and GRPO)~\cite{64:journals/focm/NesterovS17,65:journals/siamjo/GhadimiL13a,66:conf/iclr/Zhang025}, LLMs face the challenge of further learning, restricting the effectiveness and advancement potential of scaling RLVR process.

Based on the above discussion, to mitigate the issue of impaired LLM exploration ability in Pass@1 Training, we advocate for an optimization-centric approach with a higher tolerance for incorrect responses, since they might contain useful ideas or reasoning actions, preventing the model from becoming trapped in a local optimum, thereby extending the upper limit of its capabilities and enabling it to approach a global optimum gradually. 
Fortunately, with the development of LLM technologies, Pass@k has emerged to assess whether policy can generate correct responses within $k$ attempts, which is a common metric for evaluating the boundaries of LLM capabilities~\cite{25:journals/corr/abs-2407-21787}. 
Compared with the Pass@1 metric, the Pass@k metric allows the policy to generate several incorrect responses.
Thus, we consider whether the Pass@k metric can be utilized in the RLVR process to push the boundaries of LLM abilities.
Unlike Pass@1 metric, in Pass@k evaluation, to maximize the probability that at least one of the $k$ samples is successful, a ``smart'' policy will generate $k$ candidate solutions that differ from each other and cover different regions of the solution space, rather than $k$ highly similar samples.
The stronger exploration ability enables the model to acquire more comprehensive knowledge and stronger robustness.

Building on this idea, we leverage the Pass@k metric as the reward to continually train a model that has already undergone Pass@1 Training (named as \textbf{Pass@k Training}). We find that the model trained on this approach can achieve higher Pass@k scores on the test set, and maintain its Pass@1 scores. Since the naive implementation of Pass@k Training faces several critical issues, we further employ the bootstrap sampling~\cite{26:journals/technometrics/Ziegel06a,27:grunkemeier2004bootstrap} and analytical derivation to optimize the training procedure, achieving effective and efficient Pass@k Training (Section~\ref{sec:how_passk}). To further understand the feature and inner mechanism of Pass@k Training, we proposed five research questions to investigate how Pass@k Training balances the exploration and exploitation abilities of LLMs during RLVR training, observing the natural prevention of the decrease in the entropy of policy distribution, which is also a critical metric to indicate the exploration ability of policy~\cite{13:journals/corr/abs-2506-14758,14:journals/corr/abs-2501-11651,19:journals/corr/abs-2505-22617} (Section~\ref{sec:what_benefits}). Furthermore, we consider whether Pass@k Training can provide guidance and inspiration for the future development of RLVR training. From the perspective of implicit reward design, we analyze the key factors contributing to the effectiveness of Pass@k Training and explore several possible avenues for its optimization (Section~\ref{sec:why_passk}).


Overall, the vital takeaways of our work can be summarized as follows:
\begin{itemize}
    \item Compared to Pass@1 Training, Pass@k Training significantly enhances the exploration ability of LLMs, improving Pass@k performance while not harming Pass@1 scores. Among its three progressive variants, bootstrap sampling offers higher training efficiency than full sampling, and analytical derivation serves as its theoretical asymptotic form that mitigates the variance introduced by sampling. (Section~\ref{sec:how_passk})
    \item Compared to Pass@1 Training and its variants, Pass@k Training is both robust to different values of $k$ and generalizable across domains and tasks. Moreover, the enhancement of LLM exploration ability is helpful to improve their exploitation through continual training, leading 7B LLM to surpass the powerful LLMs (\eg GPT-4o and Claude-3.7), highlighting the practical value of Pass@k Training. (Section~\ref{sec:what_benefits})
    \item Pass@k Training with analytical derivation, which directly designs the advantage function, can be viewed as a form of implicit reward design. Following this idea, empirical experiments suggest that implicit reward design allows finer-grained control over optimization, such as focusing on harder problems or improving training efficiency, without complex theoretical derivations, making it a promising direction for future RLVR development.  (Section~\ref{sec:why_passk}) 
\end{itemize}

%% file: sections/2-HowPassK.tex
\begin{figure}[t]
    \centering
    \includegraphics[width=0.95\linewidth]{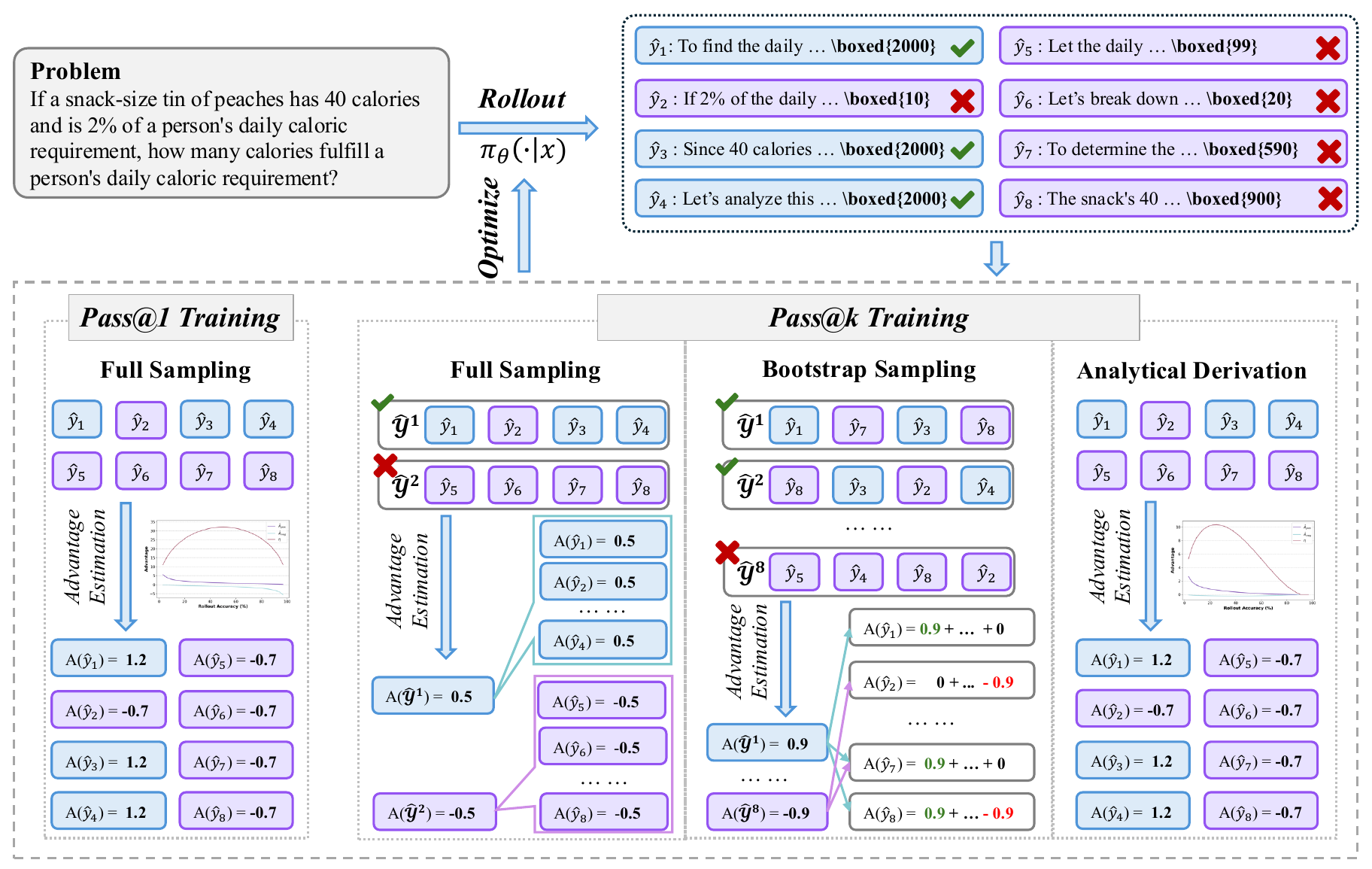}
    \caption{The overview and comparison between Pass@1 Training and Pass@k Training. The major difference between these training paradigms is in the reward calculation and advantage estimation process. Besides, full sampling, bootstrap sampling, and analytical derivation are three progressive enhancements for the Pass@k Training.
    To better demonstrate the Pass@k Training pipeline, we present the pseudo code in Appendix~\ref{app:code}.}
    \label{fig:framework}
\end{figure}

\section{Pass@k as Reward in RLVR Training}
\label{sec:how_passk}
In this section, we first formulate the reasoning tasks and provide the review of traditional Pass@1 Training (Section~\ref{sec:formulation_pass1}). Next, we introduce how to implement Pass@k as reward in RLVR training process (Section~\ref{sec:passk_w_full_sampling}), and then propose two progressive enhancements to improve the training efficiency and effectivenss (Section~\ref{code:bootstrap_sampling} and Section~\ref{sec:passk_w_analytical_derivation}).
To better illustrate Pass@k Training, we present an overview in Figure~\ref{fig:framework} and the pseudo code in Appendix~\ref{app:code}, demonstrating the implementation details of Pass@k Training.

\subsection{Formulation of Reasoning Tasks and Pass@1 Training}
\label{sec:formulation_pass1}

The complex reasoning tasks can assess the reasoning and logical abilities of LLMs. Typically, a problem from the whole dataset $D$ contains a description $x$ and a ground truth answer $y$, and the policy $\pi_\theta$ (\ie LLM with the parameters $\theta$) needs to generate a response $\hat{y}=\{t_1, t_2, \dots, t_l\}$ based on the $x$, where $t_i$ and $l$ refer to the $i$-th token and the length of the response $\hat{y}$. After obtaining the generated response $\hat{y}$, verifiers are used to verify the correctness of the LLM response and provide a reward $R(y,\hat{y})\in \{R_\text{neg},R_\text{pos}\}$ ($R_\text{neg}<R_\text{pos}$), where $R_\text{neg}$ is for negative responses and $R_\text{pos}$ is for positive responses. To simplify the notation, we use $R$ to represent $R(y,\hat{y})$. In our experiment, we adopt $R_\text{neg}=0$ and $R_\text{pos}=1$.

Based on the above formulation of reasoning tasks, in the Pass@1 Training process (\eg GRPO~\cite{4:journals/corr/abs-2402-03300}),  the advantage is estimated through the average value and standard deviation of the response rewards within the same group, which can be shown as follows,
\begin{equation}
\label{eq:grpo_avg}
    \bar{R} = \frac{1}{N_\text{rollout}}\sum_{i=1}^{N_\text{rollout}}{R_i},
\end{equation}
\begin{equation}
\label{eq:grpo_std}
    \sigma=\frac{1}{N_\text{rollout}}\sqrt{\sum_{i=1}^{N_\text{rollout}}{(R_i-\bar{R})^2}},
\end{equation}
\begin{equation}
\label{eq:grpo_adv}
    \hat{A}_{i,1}=\hat{A}_{i,2}=\dots=\hat{A}_{i,|\hat{y}_i|} = \frac{R_i - \bar{R}}{\sigma},
\end{equation}
where $N_\text{rollout}$ denotes the number of the rolled-out responses for the corresponding question, and $R_i$ and $\hat{y}_i$ refer to the rewards and the generated response of the $i$-th response, respectively. After obtaining the advantage values, GRPO utilizes the following equation to calculate the objective function $\mathcal{J}(\theta)$ that is leveraged to perform gradient descent and optimize the parameters of the model,
\begin{equation}
    \mathcal{J}(\theta)=\mathbb{E}_{(q,a)\sim D,\{o_i\}_{i=1}^G\sim \pi_\theta(\cdot|q)}\left[\frac{1}{G}\sum_{i=1}^{G}\frac{1}{|\hat{y}_i|}\sum_{t=1}^{|\hat{y}_i|}\left(\min\left(r_{i,t}\hat{A}_{i,t},\text{clip}\left(r_{i,t}, 1-\varepsilon, 1+\varepsilon \right)\hat{A}_{i,t} \right) -\beta D_{\text{kL}} \right) \right].
\end{equation}
Since each token shares the same advantage value in GRPO, we will no longer distinguish at the token level in the following discussion, and use $\hat{A}_i$ to represent the advantage value of the $i$-th response, instead. 

To enhance the effectiveness and efficiency of the RLVR training process, we employ a variant of GRPO (\ie DAPO~\cite{47:journals/corr/abs-2503-14476}) in our following experiments, only retaining the clip-higher and token-level policy gradient loss.

\subsection{Pass@k Training}
\label{sec:passk_w_full_sampling}

As discussed in previous work~\cite{41:journals/corr/abs-2503-19595,42:journals/corr/abs-2505-15201}, the behaviour of LLMs can be adjusted by the corresponding rewards. Following this idea, we consider whether the Pass@k metric can be adopted as a reward to push the boundary of LLM abilities, since the Pass@k can reflect LLM exploration ability. Thus, in this part, we first introduce the definition of the Pass@k metric and then incorporate the Pass@k metric into reward function of RLVR.

\paratitle{Definition of Pass@k Metric.}
Given the question $x$, the policy model is utilized to rollout the $k$ responses through a specific decoding strategy or searching algorithm (\eg sampling-based decoding strategy or Monte Carlo Tree Search). The $i$-th sampled response $\hat{y}_i$ will receive a reward $R_i$, which is provided by the verifier. Based on this, the value of the Pass@k metric is defined as the expected maximum reward obtained from the $k$ sampled responses. Formally, the Pass@k metric can be computed using the following equation,
\begin{equation}
\label{eq:passk}
    \text{Pass@k} = \mathbb{E}_{(x,y)\sim D,\{\hat{y}_i\}_{i=1}^k\sim \pi_\theta(\cdot|x)}\left[\max\left(R_1, \dots, R_k)\right)\right].
\end{equation}

\paratitle{Pass@k Implementation: Full Sampling.}
To integrate the Pass@k metric into the RLVR process, we propose a basic implementation through the full sampling mechanism. First, we leverage the policy $\pi_\theta$ to rollout the $N_\text{rollout}$ responses $\mathcal{\hat{Y}}=\{\hat{y}_1, \dots, \hat{y}_{N_\text{rollout}}\}$ for the given question. In this situation, these responses are separated into $N^\text{group}=\lfloor\frac{N_\text{rollout}}{k}\rfloor$ groups, and the redundant responses are discarded, where the $j$-th group contains the $k$ responses $\mathcal{\hat{Y}}^j=\{\hat{y}_{k\times (j-1)+1},\dots,\hat{y}_{k\times(j-1)+k}\}$. Next, we assign the reward scores to each group based on its Pass@k value. Concretely, the verifier will provide a reward for each response, and the group reward is computed by the maximum among the rewards of the responses belonging to it. Following the advantage estimation approach in the DAPO algorithm, the advantage value of $j$-th group $\hat{A}^{j}$ can be calculated. After that, we divide the group advantage to the responses it contains, \ie $\hat{A}_{k\times(j-1)+1}=\dots=\hat{A}_{k\times(j-1)+k}=\hat{A}^j$. Finally, we can utilize the sampled responses and their advantage value to optimize the model parameters.

\begin{figure}[t]
    \centering
    \subfloat[Pass@1 Performance of Maze Tasks.]{
        \includegraphics[width=0.48\linewidth]{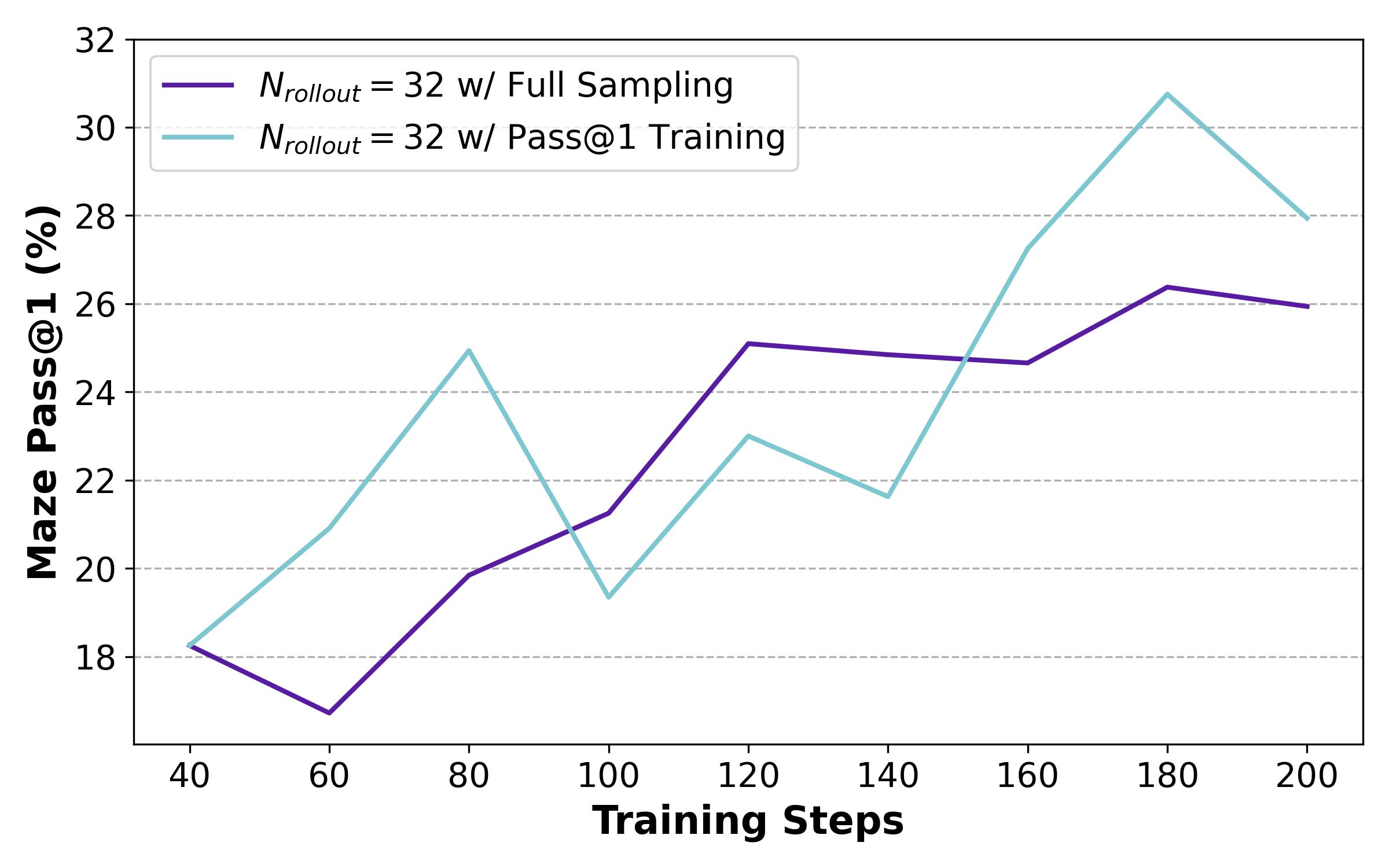}
        \label{fig:sec2.2-pass1}
    }
    \subfloat[Pass@k Performance of Maze Tasks.]{
        \includegraphics[width=0.48\linewidth]{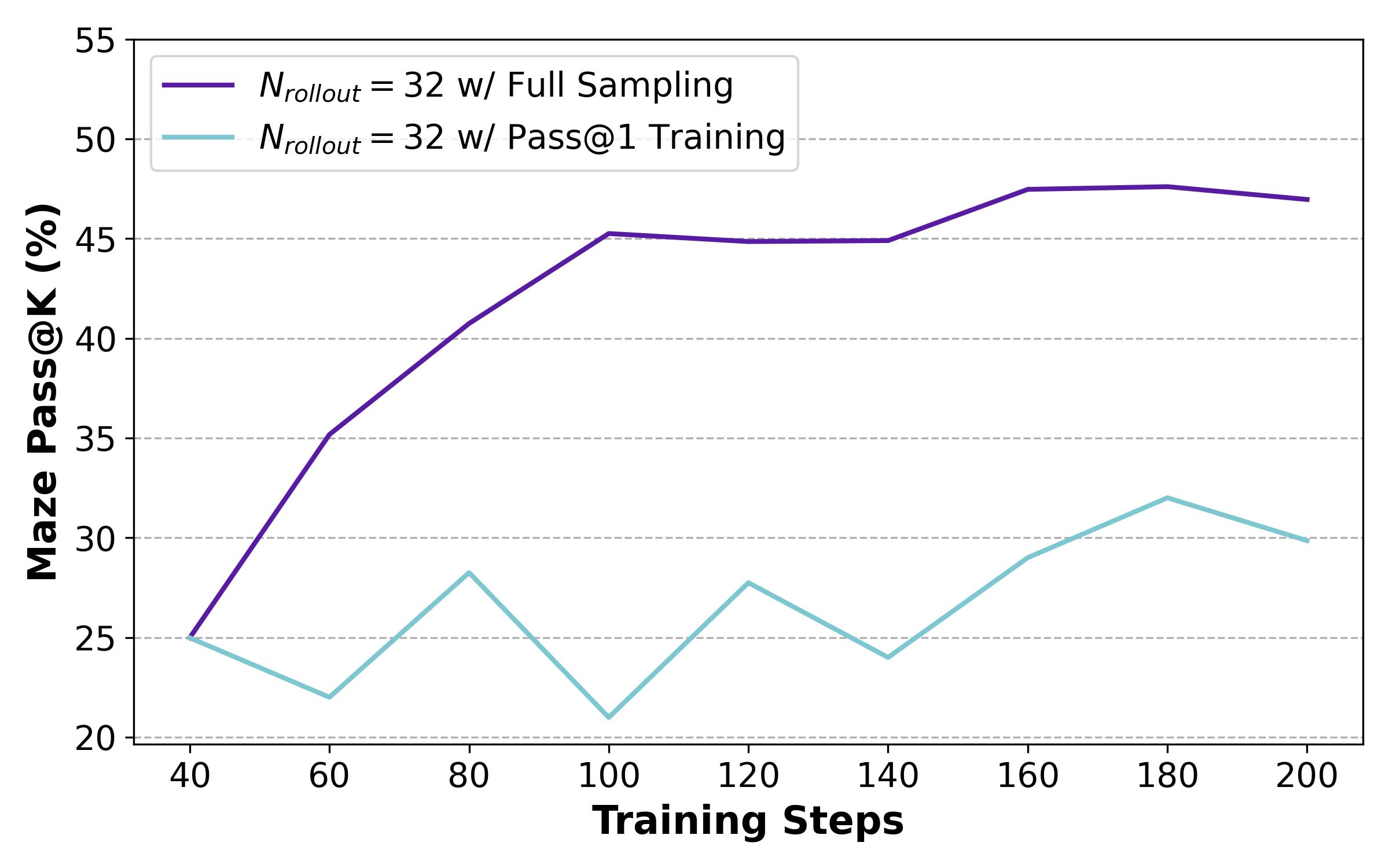}
        \label{fig:sec2.2-passk}
    }
    \caption{Training progress of Pass@1 Training and Pass@k Training with Full Sampling on baseline setting.}
    \label{fig:sec2.2}
\end{figure}

\paratitle{Empirical Insight: Improving Exploration.}
To evaluate the effectiveness of employing Pass@k as a reward, we compare the performance between Pass@k Training with full sampling and the vanilla Pass@1 Training, as shown in Figure~\ref{fig:sec2.2}. We observe that during the Pass@1 Training, Pass@k performance on downstream tasks remains stable with slight improvement. As a result, while the Pass@1 metric improves in the early stages of training, it stagnates in the later stages, indicating that the model has fallen into a local optimum. In contrast, employing Pass@k as the reward during the RLVR process, the Pass@k performance of LLM on downstream tasks achieves continual improvement, and more training steps or a larger number of rollouts consistently bring further performance improvements of LLMs, demonstrating that Pass@k Training is scalable. Moreover, Pass@k Training does not compromise the model’s Pass@1 performance and even results in Pass@1 performance gains, suggesting that Pass@k Training and Pass@1 Training share a similar optimization objective and direction, and they can be improved together.

\begin{tcolorbox}[
    colframe=takeaway,
    colback=white,
    coltitle=takeawayTitle,
]
\textcolor{takeawayTitle}{\textbf{Takeaway from Section~\ref{sec:passk_w_full_sampling}}}

Compared with the traditional RLVR training method that uses Pass@1 as the reward function, using Pass@k as the reward function for RLVR training can effectively improve the model's Pass@k performance on downstream tasks without compromising its Pass@1 performance.
\end{tcolorbox}

\subsection{Efficient Pass@k Training via Bootstrap Sampling}
\label{sec:passk_w_bootstrap_sampling}

Although Pass@k Training can push the limit of LLM abilities, the rollout times increases significantly with the increase of $N^\text{group}$, costing a higher computational resources. Thus, in this part, we consider utilizing the bootstrap sampling mechanism to reduce the rollout times while maintaining a constant number of groups.

To achieve the efficient Pass@k Training, during the rollout process, we first use the policy model $\pi_\theta$ to generate $N_{\text{rollout}}$ responses $\mathcal{\hat{Y}}=\{\hat{y}_1,\dots,\hat{y}_{N_{\text{rollout}}}\}$ for the given question $x$. Next, to construct the $N^{\text{group}}$ groups for the following optimization process, we randomly sample the $k$ responses from the previously generated response set $\mathcal{\hat{Y}}$, and these sampled responses collectively constitute a group. Specifically, to construct the $j$-th group, we select $k$ distinct values from the range $1$ to $N_\text{rollout}$, obtaining a set $\mathcal{P}=\{p_{j,1},\dots,p_{j,k}\}$. Then the responses whose indices are in the set $\mathcal{P}$ constitute the current group $\hat{\mathcal{Y}}^j=\{\hat{y}_{p_{j,1}},\dots,\hat{y}_{p_{j,k}}\}$. This procedure will be repeated to $N^{\text{group}}$ times, collecting $N^{\text{group}}$ groups of responses. Once the groups are obtained, we can estimate the advantage value for each group and assign it to the responses. Since we use a bootstrap sampling strategy to construct groups, some responses may appear in multiple groups. Therefore, for each response, we compute its final advantage by summing the advantages of all groups to which it belongs, \ie
\begin{equation}
\label{eq:bootstrap_sampling_group_to_response}
    \hat{A}_{i}=\sum_{j=1}^{N^\text{group}}\hat{A}^j\cdot \mathbb{I}[\hat{y}_i \in \hat{\mathcal{Y}}^j],
\end{equation}
where $\mathbb{I}[\hat{y}_i\in \hat{\mathcal{Y}}^j]$ is an indicator function, which returns $1$ if and only if the $i$-th response $\hat{y}_i$ belongs to the $j$-th group $\hat{\mathcal{Y}}^j$, while returns $0$ for others. In practice, we set $N^{\text{group}}=N_{\text{rollout}}$ for an efficient RLVR process.

\begin{figure}[t]
    \centering
    \subfloat[Pass@1 Performance of Maze Tasks.]{
        \includegraphics[width=0.48\linewidth]{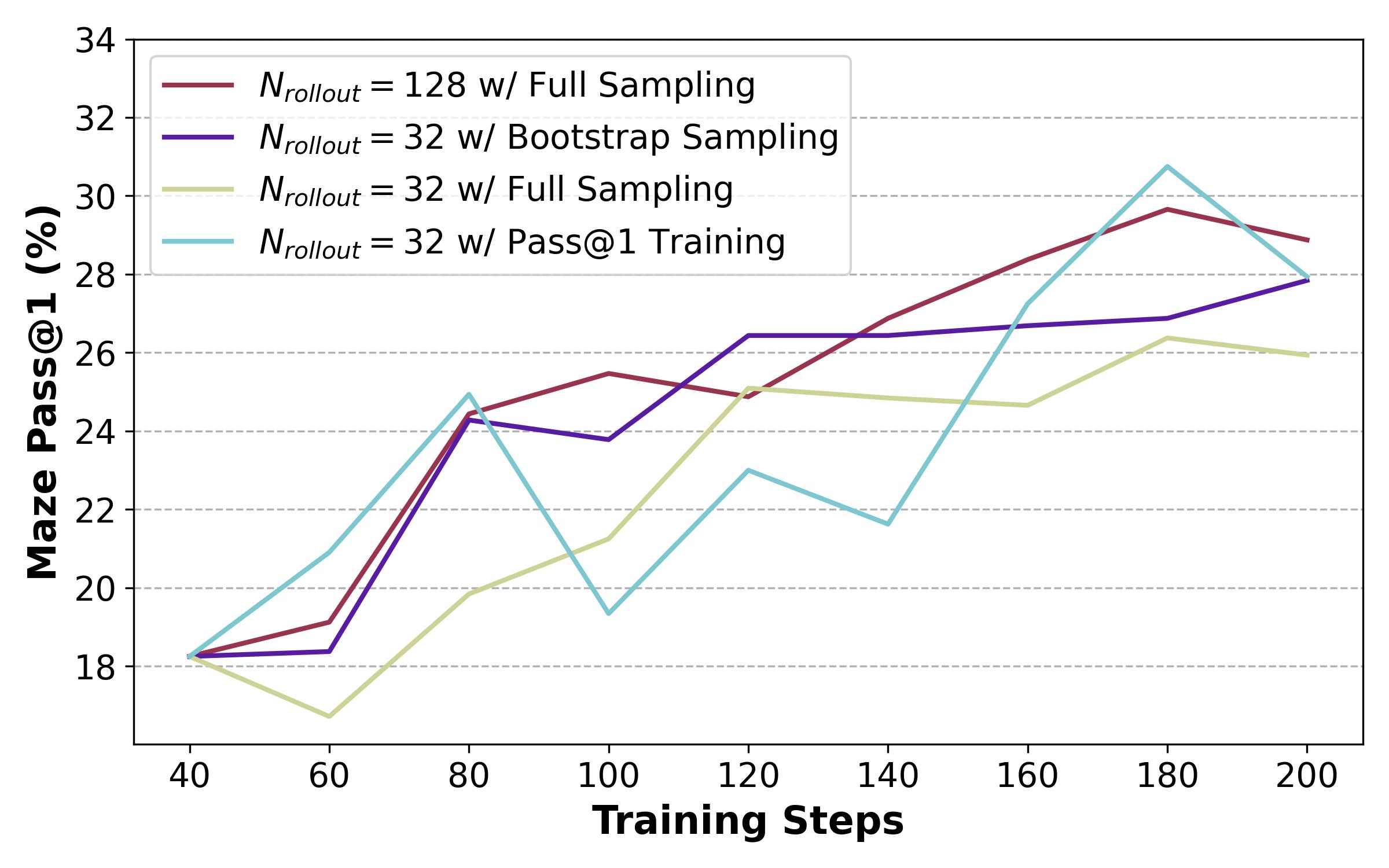}
        \label{fig:sec2.3-pass1}
    }
    \subfloat[Pass@k Performance of Maze Tasks.]{
        \includegraphics[width=0.48\linewidth]{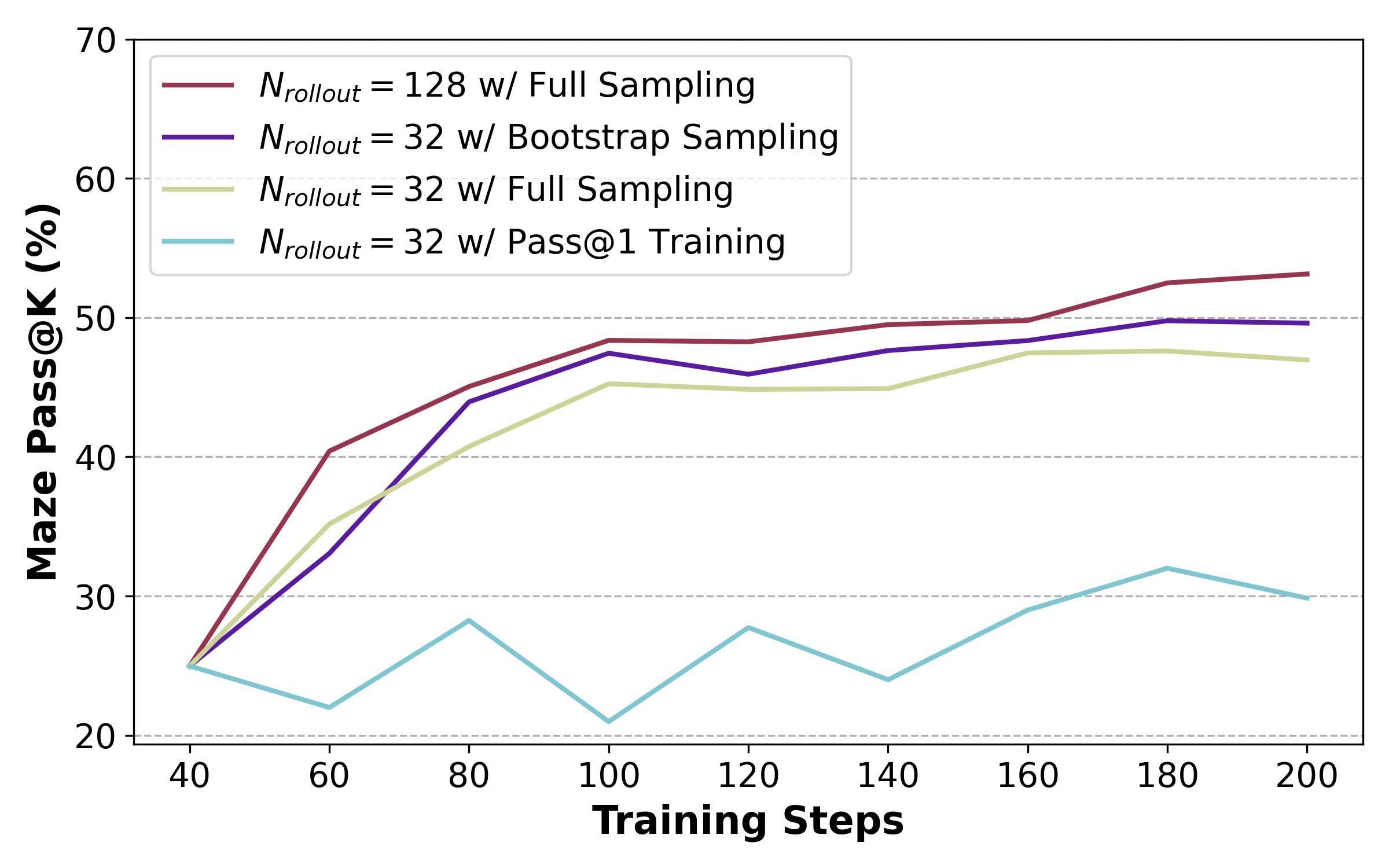}
        \label{fig:sec2.3-passk}
    }
    \caption{Training progress of Pass@1 Training and Pass@k Training with Bootstrap Sampling under various $N_\text{rollout}$.}
    \label{fig:sec2.3}
\end{figure}

\paratitle{Empirical Insight: Reduction in Training Budget.}
To assess the effectiveness of bootstrap sampling for Pass@k Training, we conduct Pass@1 Training and Pass@k Training with full sampling (described in Section~\ref{sec:passk_w_full_sampling}) with different rollout times as the baseline methods, and present the evaluation in Figure~\ref{fig:sec2.3}. Given the same number of rollouts $N_\text{rollout}$ (\ie ``$N_\text{rollout}=32$ w/ Full Sampling'' v.s. ``$N_\text{rollout}=32$ w/ Bootstrap Sampling''), bootstrap sampling outperforms full sampling. This improvement arises from the fact that bootstrap sampling generates a larger number of groups, which in turn reduces the variance of the sampled reward distribution relative to the true distribution, leading to more stable and effective training. With the same number of groups $N^\text{group}$, bootstrap sampling does not lead to significant performance degradation on the Pass@k metric compared to full sampling (\ie ``$N_\text{rollout}=128$ w/ Full Sampling''), and it requires only one-fourth of the theoretical computational cost, resulting in higher training efficiency. Additionally, it achieves comparable performance to full sampling on the Pass@1 metric. In conclusion, Pass@k Training with bootstrap sampling outperforms Pass@1 Training and enhances the efficiency of the training process with full sampling.

\begin{tcolorbox}[
    colframe=takeaway,
    colback=white,
    coltitle=takeawayTitle,
]
\textcolor{takeawayTitle}{\textbf{Takeaway from Section~\ref{sec:passk_w_bootstrap_sampling}}}

Compared with the full sampling-based Pass@k Training method, the bootstrap sampling-based training method can achieve better training results with the same number of rollouts. With the same number of groups, it can reduce computational overhead while achieving comparable performance.
\end{tcolorbox}

\subsection{Analytical Derivation of Efficient and Effective Pass@k Training}
\label{sec:passk_w_analytical_derivation}

Following the idea of the bootstrap sampling mechanism described in Section~\ref{sec:passk_w_bootstrap_sampling}, we derive the analytical solution of the response advantage (\ie \textbf{$\hat{A}_\text{pos}$} and \textbf{$\hat{A}_\text{neg}$}) to remove the variance brought by the sampling operation for constructing the groups. The details of the derivation are presented in Appendix B.

To deduce the analytical formula of the advantage, we start by analyzing the advantage reward and standard deviation of the groups, \ie $\bar{R}^\text{group}$ and $\sigma^\text{group}$. The group that contains at least one positive response (named as positive group) will be assigned the positive reward $R_{\text{pos}}$, while the other groups (named as negative group) will be endowed with the negative reward $R_{\text{neg}}$. Following the advantage estimation method of DAPO, it is critical to calculate the average and standard variance of the reward scores of the groups. First, the average reward of the group can be formulated as the following equation,
\begin{equation}
\label{eq:def_group_avg}
    \bar{R}^\text{group}=\frac{1}{N^\text{group}_\text{total}}\times\left(N^\text{group}_\text{pos} \times R_\text{pos} + N^\text{group}_\text{neg} \times R_\text{neg}\right),
\end{equation}
where $N^\text{group}_\text{total}$ refers to the total number of groups, and $N^\text{group}_\text{pos}$ and $N^\text{group}_\text{neg}$ denote the number of positive and negative groups, respectively. To calculate the number of positive and negative groups, we first define the number of positive responses as $N_\text{pos}$ and the number of negative responses as $N_\text{neg}$. Typically, $N_\text{pos} + N_\text{neg}=N_\text{rollout}$. Based on the above definition, as each group is constructed by selecting $k$ responses, we can obtain the total number of the group $N^\text{group}_\text{total}$ as follows,
\begin{equation}
\label{eq:n_group_total}
    N^\text{group}_\text{total}=\binom{N_{\text{rollout}}}{k}=N^\text{group}_\text{pos}+N^\text{group}_\text{neg}.
\end{equation}
Since negative groups do not contain the positive responses, when and only when randomly sampling $k$ negative responses from the whole responses $\mathcal{O}$, these sampled responses can construct a negative group. Thus, the number of negative groups can be calculated as follows,
\begin{equation}
\label{eq:n_group_neg}
    N^\text{group}_\text{neg} = \binom{N_\text{neg}}{k}.
\end{equation}
According to Eq.~\ref{eq:n_group_total} and Eq.~\ref{eq:n_group_neg}, we can obtain the number of the positive groups,
\begin{equation}
\label{eq:n_group_pos}
    N^\text{group}_\text{pos}=N^\text{group}_\text{total}-N^\text{group}_\text{neg}=\binom{N_{\text{rollout}}}{k}-\binom{N_\text{neg}}{k}.
\end{equation}
Substitute Eq.~\ref{eq:n_group_total}, Eq.~\ref{eq:n_group_neg}, and Eq.~\ref{eq:n_group_pos} into Eq.~\ref{eq:def_group_avg}, we can obtain the average rewards of the group $\bar{R}^\text{group}$,
\begin{equation}
\label{eq:group_avg}
   \bar{R}^\text{group}=1-\frac{\binom{N_\text{neg}}{k}}{\binom{N_\text{rollout}}{k}}. 
\end{equation}
Based on the average rewards of the group $\bar{R}^\text{group}$, the standard variance can be calculated as follows,
\begin{equation}
\label{eq:group_std}
    \sigma^\text{group}=\sqrt{\bar{R}^\text{group}\times\left(1-\bar{R}^\text{group}\right)}.
\end{equation}
Based on the average (Eq.~\ref{eq:group_avg}) and the standard variance (Eq.~\ref{eq:group_std}) of reward scores, we can finally deduce the advantage of the positive group $\hat{A}^\text{group}_\text{pos}$ and the negative group $\hat{A}^\text{group}_\text{neg}$ as follows,
\begin{equation}
    \hat{A}^\text{group}_\text{pos}=\frac{R_\text{pos}-\bar{R}^\text{group}}{\sigma^\text{group}}=\frac{1-\bar{R}^\text{group}}{\sigma^\text{group}},~~\hat{A}^\text{group}_\text{neg}=\frac{R_\text{neg}-\bar{R}^\text{group}}{\sigma^\text{group}}=-\frac{\bar{R}^\text{group}}{\sigma^\text{group}}.
\end{equation}
To transfer the group-relative advantage $\hat{A}^\text{group}_\text{pos}$and $\hat{A}^\text{group}_\text{neg}$ obtained in the previous section to the response-relative advantage $\hat{A}_{pos}$ and $\hat{A}_{neg}$, we need to consider, for each response, the correctness of the group it belongs to and compute the advantage value proportionally. Typically, a response will belong to $\binom{N_\text{rollout}-1}{k-1}$ groups, because a group can be formed with the current response if and only if $k-1$ responses are selected from the remaining $N_\text{rollout}-1$ responses. Further, for a positive response, the groups that it belongs to can always pass the Pass@k verification (\ie positive group). Thus, \textbf{the advantage of a positive response $\hat{A}_\text{pos}$} can be calculated as follows,
\begin{equation}
\label{eq:response_pos}
    \hat{A}_\text{pos}=\frac{1-\bar{R}^\text{group}}{\sigma^\text{group}}.
\end{equation}
Then, considering a negative response, the group that it belongs to is the negative group if and only if the other $k-1$responses are all negative responses. In this case, the number of required groups is $\binom{N_\text{neg}-1}{k-1}$, \ie the current response can form a negative group with any $k-1
$ responses selected from the remaining $N_\text{neg}-1$ negative responses. Based on the number of negative groups, we can compute the number of positive groups by subtracting the number of negative groups from the total number of groups, \ie $\binom{N_\text{rollout}-1}{k-1}-\binom{N_\text{neg}-1}{k-1}$. Thus, \textbf{the advantage of a negative response $\hat{A}_\text{neg}$} can be calculated as follows,
\begin{equation}
\label{eq:response_neg}
    \hat{A}_\text{neg}=\left(1-\bar{R}^\text{group}-\frac{\binom{N_\text{neg}-1}{k-1}}{\binom{N_\text{rollout}-1}{k-1}}\right)\times\left(\sigma^\text{group}\right)^{-1}.
\end{equation}
After obtaining the analytical solutions of response-relative advantage $\hat{A}_\text{pos}$ and $\hat{A}_\text{neg}$, we directly employ them in the advantage estimation process and then optimize the model parameters. 
By examining the analytical solutions of the advantage value, we observe that it depends only on the total number of sampled responses $N_\text{rollout}$, the number of positive responses $N_\text{pos}$, the number of negative responses $N_\text{neg}$, and the value of $k
$. Therefore, after the rollout procedure, we can directly compute the advantage value of each response for RLVR training without going through the previously described cumbersome reward calculation process. 

\begin{figure}[t]
    \centering
    \subfloat[Pass@1 Performance of Maze Tasks.]{
        \includegraphics[width=0.48\linewidth]{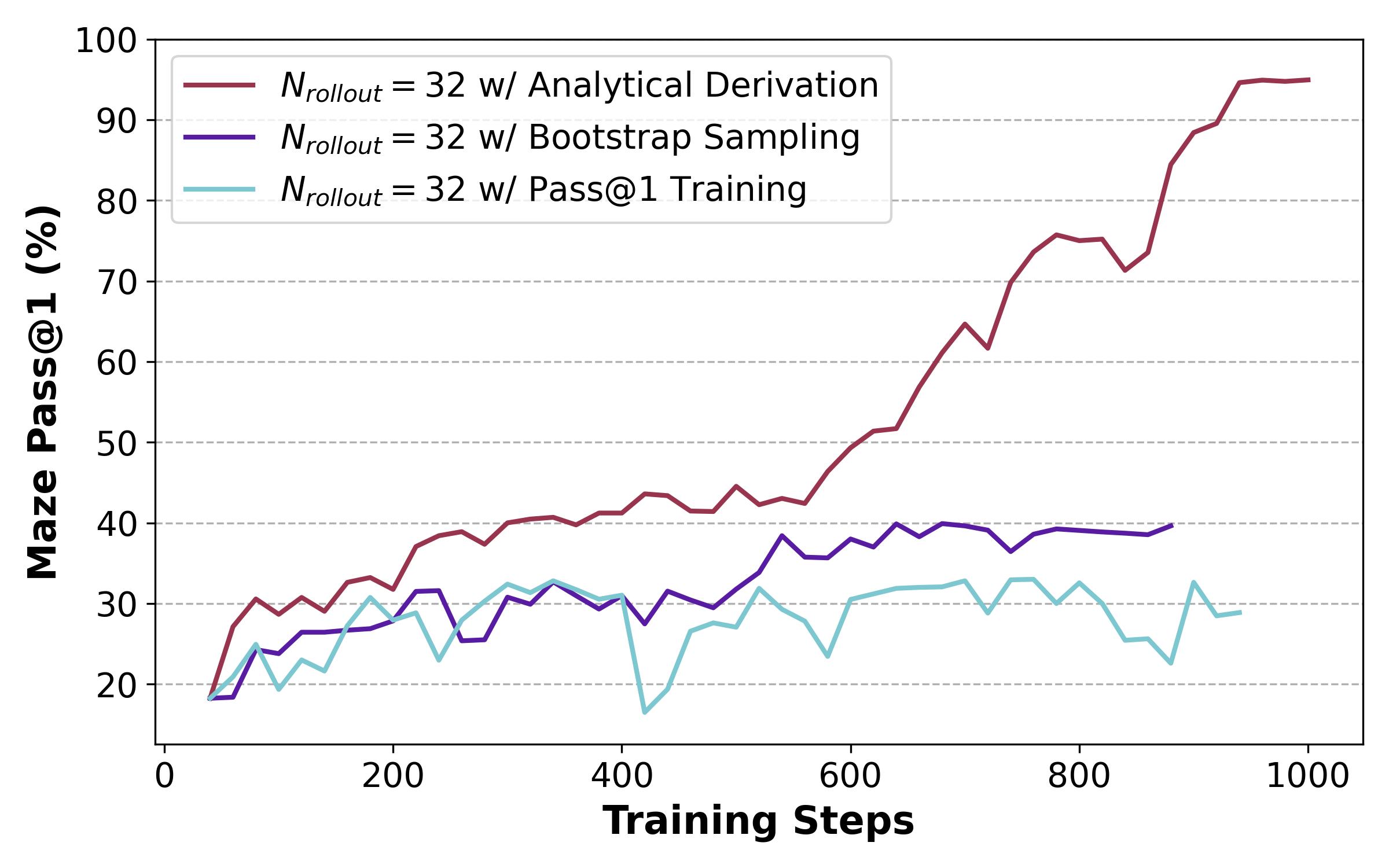}
        \label{fig:sec2.4-pass1}
    }
    \subfloat[Pass@k Performance of Maze Tasks.]{
        \includegraphics[width=0.48\linewidth]{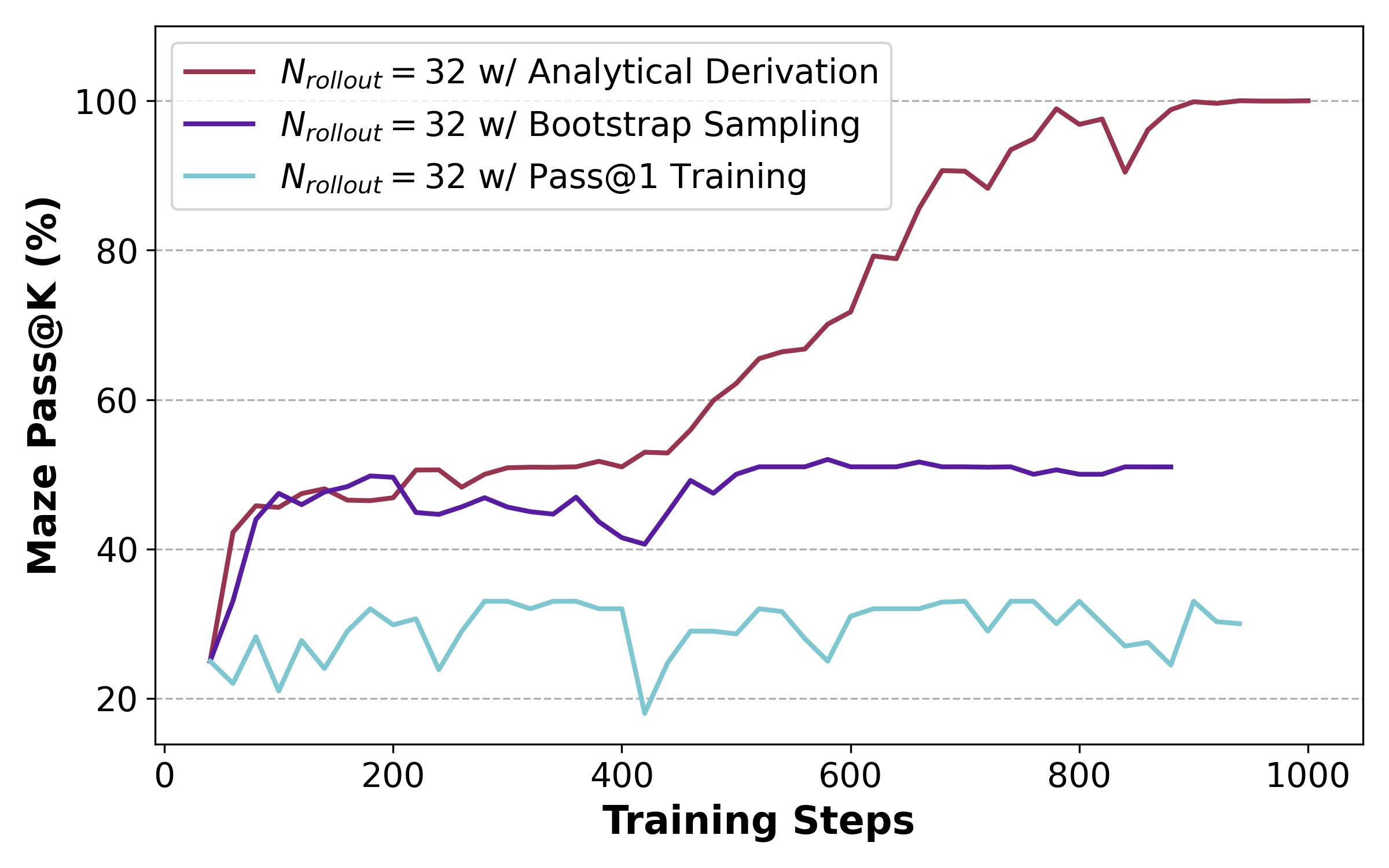}
        \label{fig:sec2.4-passk}
    }
    \caption{Training progress of Pass@1 Training and Pass@k Training with Analytical Derivation and Bootstrap Sampling on baseline setting.}
    \label{fig:sec2.4}
\end{figure}

\paratitle{Empirical Insight: Further Improvement on Pass@k.}
For the evaluation and comparison, we unified the number of rollouts $N_\text{rollout}$ to 32 and compared the training effects of Pass@1 Training and Pass@k Training with bootstrap sampling and analytical derivation. The experimental results are presented in Figure~\ref{fig:sec2.4}.
To present the comprehensive evaluation, we also conduct the external experiments about different LLMs on various tasks and present the results in Appendix~\ref{sec:app_exp}.
In the experiment, we can observe that both types of Pass@k Training achieve better results than Pass@1 Training, which further confirms the effectiveness of Pass@k Training. When the number of training steps increases, the Pass@k Training based on bootstrap sampling experiences a relatively sharp performance fluctuation at 400 steps, with the Pass@k performance declining, which indicates that this method has certain instability. In contrast, for the method based on bootstrap sampling, Pass@k Training with analytical derivation eliminates the sampling process required for constructing groups. It directly reduces the variance caused by the sampling process through the calculation of analytical solutions, thereby providing a more stable training process. Thus, the method of Pass@k Training with analytical derivation can reduce fluctuations during the training process and bring about continuous performance improvements as the number of training steps increases.

\begin{tcolorbox}[
    colframe=takeaway,
    colback=white,
    coltitle=takeawayTitle,
]
\textcolor{takeawayTitle}{\textbf{Takeaway from Section~\ref{sec:passk_w_analytical_derivation}}}

Pass@k Training with analytical derivation not only avoids the computational overhead caused by a large number of rollouts in full sampling, but also eliminates the variance introduced by sampling in bootstrap sampling. This makes the RLVR training process more efficient and effective, and can guide the model's exploration ability to continuously improve as the number of training steps scales.
\end{tcolorbox}

%% file: sections/3-WhatBenefits.tex
\section{Balancing Exploration and Exploitation with Pass@k Training}
\label{sec:what_benefits}

In this section, we further investigate the features and effectiveness of Pass@k Training. First, we compare Pass@k Training with commonly used methods for enhancing model exploration ability~\cite{14:journals/corr/abs-2501-11651,69:journals/corr/abs-2506-10947} to further verify its effectiveness (Section~\ref{sec:noise_entropy}). Next, to further understand how Pass@k Training influences its exploration capability, we examine the diversity of model’s responses and the entropy of policy distribution (Section~\ref{sec:exploration_ability}). After that, we wonder whether the improvement brought by Pass@k Training can be transferred to other domains or tasks, and then assess the generalization performance of it (Section~\ref{sec:generalization}). Moreover, as RLVR stability and robustness are widely concerned~\cite{22:journals/corr/abs-2505-22312,28:hong2025glm,67:journals/tmlr/CasperDSGSRFKLF23}, we analyze how the value of $k$ affects the Pass@k Training process (Section~\ref{sec:robustness}). Finally, since Pass@1 is a more important metric in practical applications, we explore how to transfer the benefits of Pass@k Training to the model's Pass@1 performance, and the experiment results demonstrate the high practical value of Pass@k Training (Section~\ref{sec:passk_to_pass1}).

\subsection{How does Pass@k Training Compare to Noise Rewards or Entropy Regularization?}
\label{sec:noise_entropy}

\begin{figure}[h]
    \centering
    \subfloat[Pass@k Performance of Noise Rewards.]{
        \includegraphics[width=0.48\linewidth]{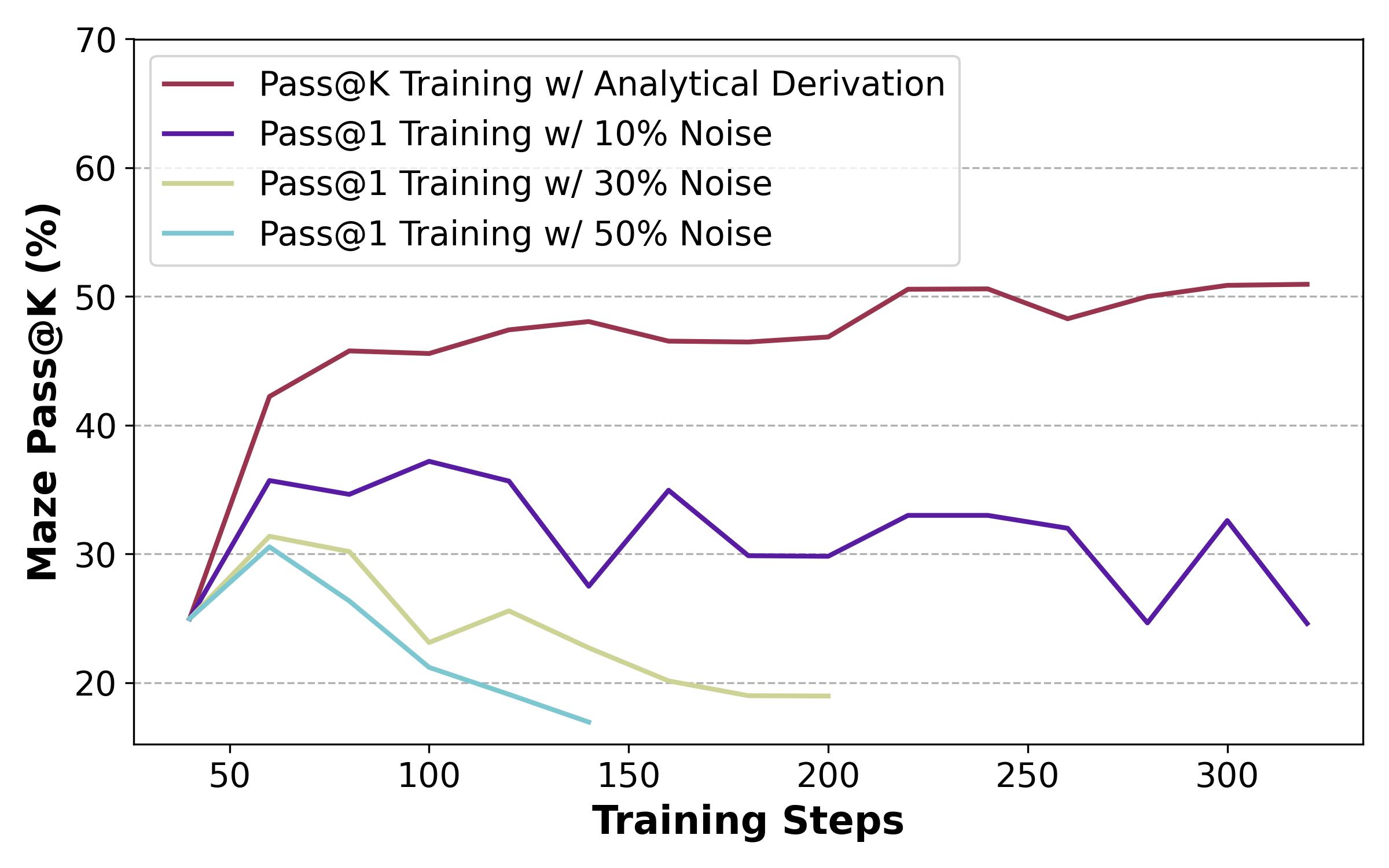}
        \label{fig:sec3.1-noise}
    }
    \subfloat[Pass@k Performance of Entropy Regularization.]{
        \includegraphics[width=0.48\linewidth]{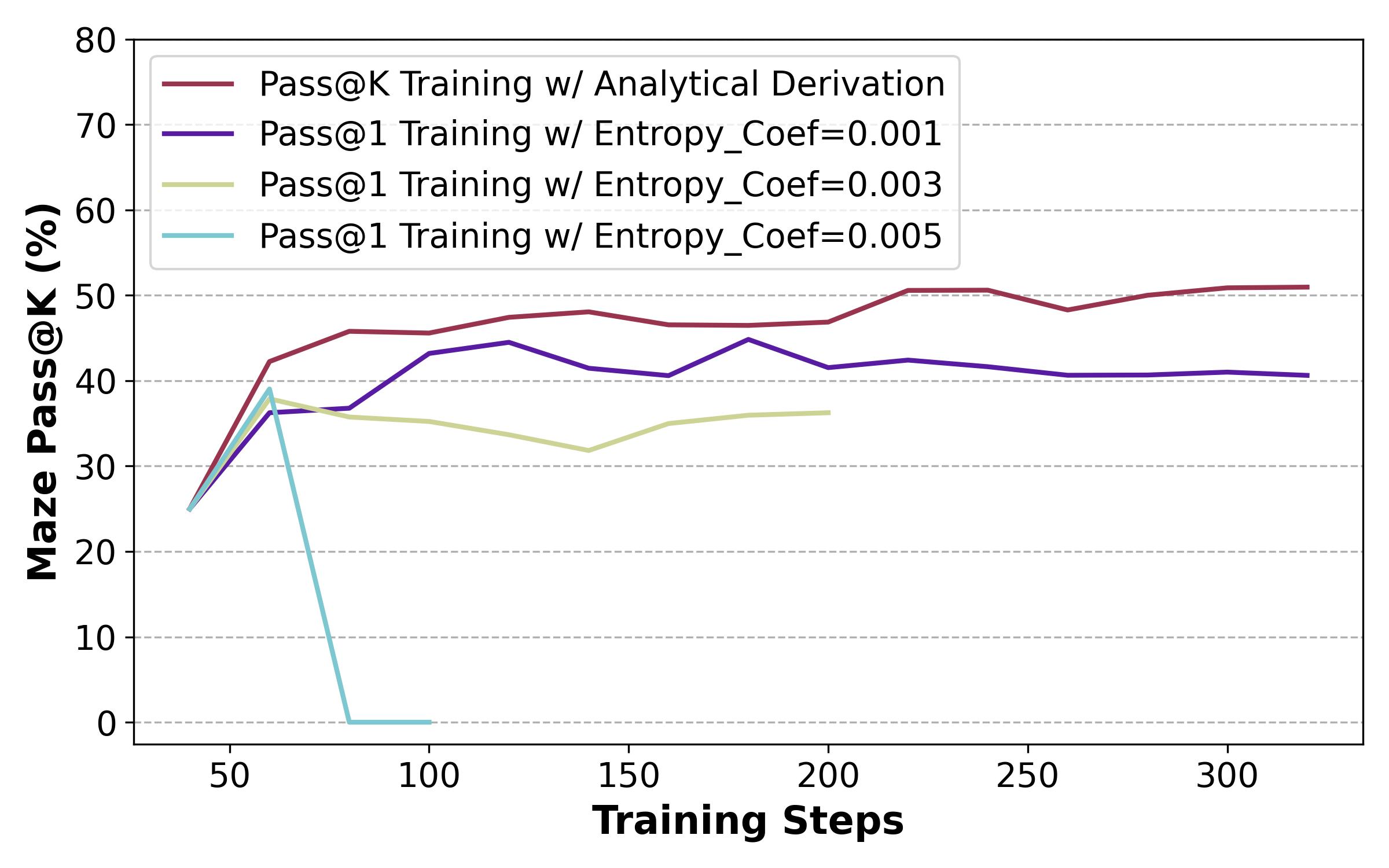}
        \label{fig:sec3.1-entropy}
    }
    \caption{Training progress of Noise Rewards and Entropy Regularization on baseline setting.}
    \label{fig:sec3.1}
\end{figure}

Inspired by the Pass@k Training procedure (Section~\ref{sec:passk_w_full_sampling}) and the previous work~\cite{14:journals/corr/abs-2501-11651}, we conduct a comparison between Pass@k Training and two baseline approaches, \ie Noise Rewards and Entropy Regularization.

\paratitle{Noise Rewards.}
Reviewing the RLVR procedure that leverages the Pass@k metric as the reward (as described in Section~\ref{sec:passk_w_full_sampling}), we note that some negative responses may receive the positive reward $R_{\text{pos}}$ if they belong to a positive group. This raises the question of whether the improvement in Pass@k scores is partially driven by learning from these negative responses with counterfactual positive rewards. To investigate this, we conduct an experiment in which a certain proportion (\ie 10\%, 30\%, and 50\%) of the rewards assigned to negative responses are flipped. The results are presented in Figure~\ref{fig:sec3.1-noise}. Empirical results indicate that encouraging LLMs to learn from negative responses does not contribute to improving their reasoning abilities. On the contrary, introducing a higher proportion of noise into the reward significantly degrades model performance. As the proportion of flipped rewards increases, the model's performance deteriorates progressively on both Pass@1 and Pass@k metrics. Furthermore, performance continues to decline with additional training steps. These findings suggest that naively incorporating noise into the reward does not enhance the reasoning capabilities of LLMs. Instead, the proportion of noise must be carefully controlled, such as through the structured design of the Pass@k metric, which can guide LLMs beyond the limitations of their existing reasoning abilities.

\paratitle{Entropy Regularization.}
A surge of studies~\cite{13:journals/corr/abs-2506-14758,14:journals/corr/abs-2501-11651,19:journals/corr/abs-2505-22617,62:journals/corr/abs-2506-01939} have pointed out that entropy can indicate the exploration ability of LLMs and can be incorporated into the objective function to preserve their exploration ability. Thus, following the previous work~\cite{14:journals/corr/abs-2501-11651}, we employ entropy regularization with the coefficient of $\{0.001, 0.003, 0.005\}$ in the RLVR training process and present the results in the right part of Figure~\ref{fig:sec3.1-entropy}. According to the results, we can find that a high coefficient of entropy regularization might cause the collapse of the model, like setting the coefficient to 0.005. Although the small coefficient of entropy regularization does not make LLM crush, it still cannot outperform Pass@k Training, and even lead to the decrease of the performance of LLMs with the increase of the training steps. The above phenomenon indicates that entropy regularization might affect training effectiveness and stability.

\paratitle{Discussion about Other Entropy-guided Approaches.}
We compare the effectiveness between Pass@k Training and the naive implementation of Entropy-guided Approach (\ie Entropy Regularization).
Moreover, there are several other methods, such as integrating the entropy into the advantage function~\cite{13:journals/corr/abs-2506-14758} or focusing on tokens with high covariance~\cite{19:journals/corr/abs-2505-22617}.
Similarly, these methods might introduce a new trade-off: overly strict constraints may lead to underfitting and insufficient model training, while overly loose constraints can result in instability during training, potentially affecting the training effectiveness and model performance~\cite{22:journals/corr/abs-2505-22312,28:hong2025glm,67:journals/tmlr/CasperDSGSRFKLF23}, since entropy conflicts with the Pass@1 metric.
Therefore, the hyper-parameters should be carefully selected during the above methods to bring the performance improvement of LLMs.
Actually, these methods are orthogonal to Pass@k Training, meaning that we can also combine these methods with Pass@k Training to achieve better training results.
To verify this, we conduct the experiments in Section~\ref{sec:adaptive_training} to assess the effectiveness of Pass@k Training based on the guidance of policy entropy, demonstrating significant improvements.

\begin{tcolorbox}[
    colframe=takeaway,
    colback=white,
    coltitle=takeawayTitle,
]
\textcolor{takeawayTitle}{\textbf{Takeaway from Section~\ref{sec:noise_entropy}}}

Pass@k Training outperforms Noise Rewards and Entropy Regularization: randomly flipping the reward of negative responses might degrade the performance of LLMs, and incorporating Entropy Regularization brings new trade-off issues, which hardly achieve continuous improvement.
\end{tcolorbox}

\subsection{Does Pass@k Training Really Improve the Exploration Ability of LLMs?}
\label{sec:exploration_ability}

\begin{figure}[h]
    \centering
    \begin{subfigure}{0.48\textwidth}
        \centering
        \includegraphics[width=\textwidth]{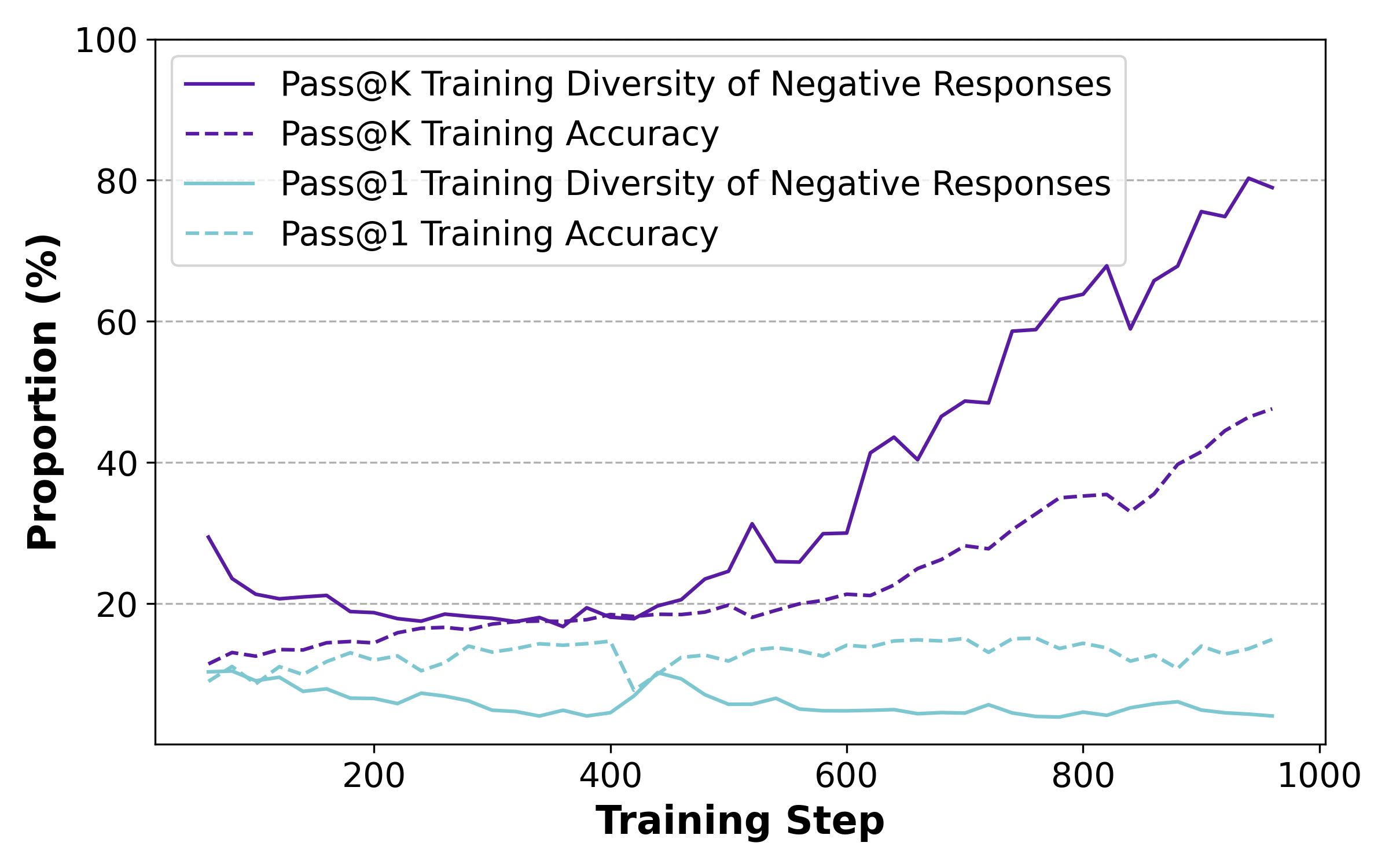}
        \caption{Diversity of Negative Responses on Maze Tasks.}
        \label{fig:sec3.2-answer_diversity}
    \end{subfigure}
    \begin{subfigure}{0.48\textwidth}
        \centering
        \includegraphics[width=\textwidth]{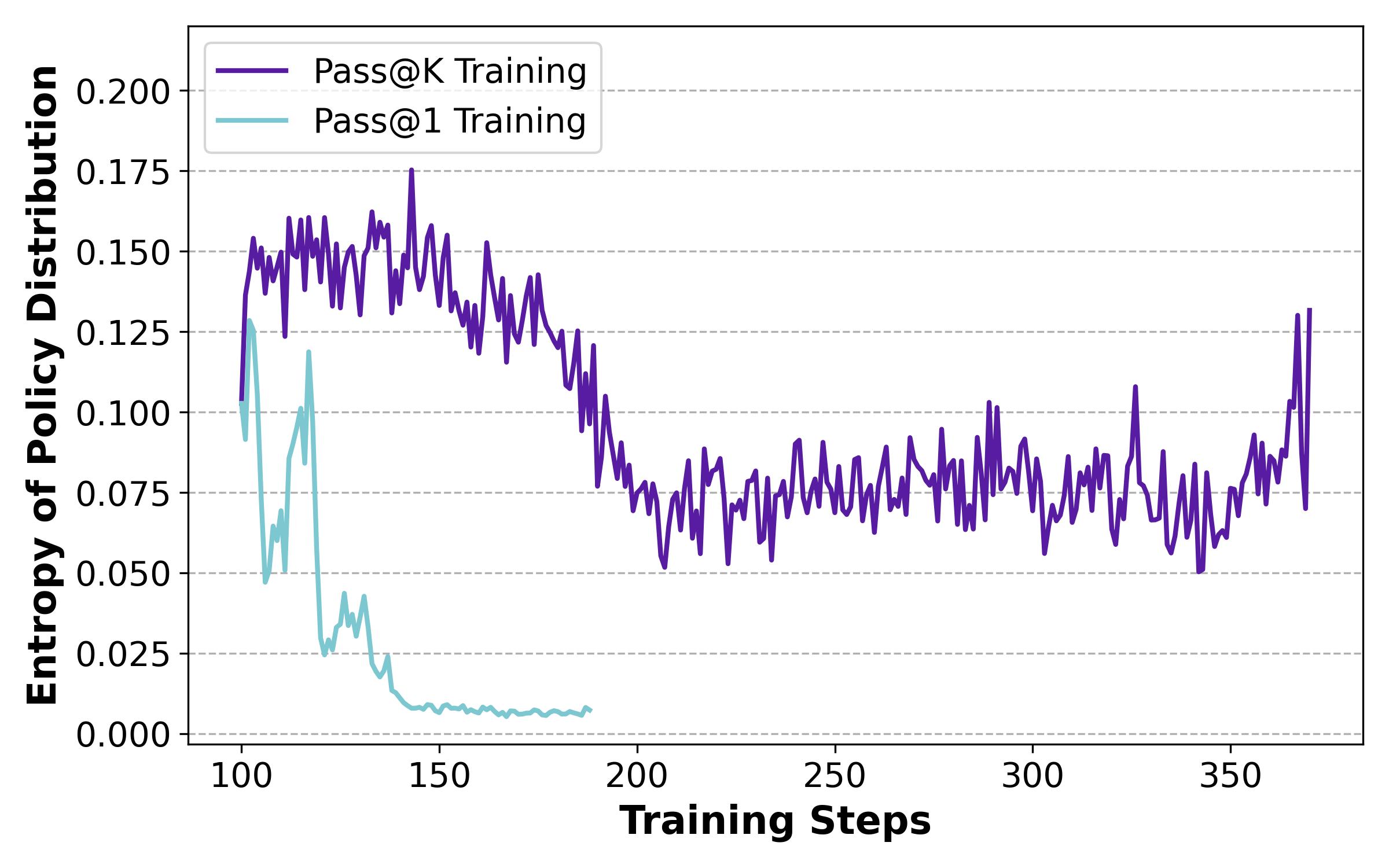}
        \caption{Entropy of Policy Distribution on Puzzle Tasks.}
        \label{fig:sec3.2-policy_entropy}
    \end{subfigure}
    \caption{Training progress of Pass@1 Training and Pass@k Training on baseline setting.}
    \label{fig:sec3.2}
\end{figure}

To analyze the changes in exploration of LLMs during the RLVR training process, from the perspective of answer diversity and entropy of policy distribution, we conduct the related empirical study and show the corresponding results in Figure~\ref{fig:sec3.2}.

\paratitle{Answer Diversity of Negative Responses.}
We counter the accuracy and the ratio of different answers among the negative responses of Pass@k and Pass@1 Training, which is presented in Figure~\ref{fig:sec3.2-answer_diversity}, aiming to assess the exploration ability of LLMs on the uncertain answer. According to the results, we observe that the answer diversity of the negative response stays at the same level during the RLVR training process, indicating that the LLMs try to select the ``safe'' actions and tends to generate similar answers in the exploration procedure, limiting the scope of exploration and constrain the effectivenss of RLVR. Differently, in Pass@k Training, the model is encouraged to achieve a higher Pass@k score and naturally learn the strategy to generate diverse answers when the model is not confident about this question. In this case, the exploration ability of LLM is enhanced and thereby improves its exploitation ability (\ie Pass@1 score).

\paratitle{Entropy of Policy Distribution.}
In Figure~\ref{fig:sec3.2-policy_entropy}, the results show a similar conclusion to our previous discussion on answer diversity. Pass@k Training keeps the entropy of policy distribution at a relatively high level, while Pass@1 Training induces entropy to converge to a low value. This phenomenon suggests that LLMs can keep their exploration ability during Pass@k Training, but will lose their exploration ability during Pass@1 Training. On the other hand, we can also observe that Pass@k Training leads to an increase in entropy, starting from step 200 of the RLVR procedure. This phenomenon validates our hypothesis that using Pass@k as the training objective can encourage the model to conduct more exploration, thereby naturally increasing entropy.

In conclusion, exploration and exploitation do not conflict with each other, and they can improve mutually. Pass@k Training can achieve this goal.

\begin{tcolorbox}[
    colframe=takeaway,
    colback=white,
    coltitle=takeawayTitle,
]
\textcolor{takeawayTitle}{\textbf{Takeaway from Section~\ref{sec:exploration_ability}}}

Pass@k Training can encourage the model to conduct more exploration, generating diverse answers that naturally leads to an increase in entropy, when the model does not have sufficient confidence to generate the correct answer.
\end{tcolorbox}

\subsection{What is the Generalization Ability of LLMs After Pass@k Training?}
\label{sec:generalization}

\begin{table}[h]
    \centering
    \small
    \caption{Pass@1/Pass@k Performance of Qwen2.5-7B-Instruct trained on different RLVR approaches.}
      \begin{tabular}{lcccc}
      \toprule
      \multirow{2.5}*{\textbf{Pass@1/Pass@k}} & \multicolumn{2}{c}{\textbf{In-Domain Tasks}} & \multicolumn{2}{c}{\textbf{Out-of-Domain Tasks}} \\
      \cmidrule(r){2-3}\cmidrule(r){4-5}
      & ARC-AGI 1 & Enigmate & KORBench & AIME 2025 \\
      \midrule
      Qwen2.5-7B-Instruct & 2.4/4.8 & 4.8/10.1 & 36.5/45.9 & 4.2/15.8 \\
      + Pass@1 Training & 3.3/3.8 & 12.9/21.3 & 37.7/45.6 & 5.4/19.1 \\
      + Pass@k Training & \textbf{4.0}/\textbf{5.3} & \textbf{17.9}/\textbf{29.8} & \textbf{47.7}/\textbf{63.5} & \textbf{7.1}/\textbf{22.4} \\
      \bottomrule
      \end{tabular}
      \label{tab:generalization}
\end{table}

To analyze the generalization ability of Pass@k Training, we conduct the corresponding experiments and present the results in Table~\ref{tab:generalization}. We can observe that Pass@1 and Pass@k Training can enhance the model's capacities on in-domain and out-of-domain tasks, suggesting the strong generalization ability of the RLVR training process. Further, comparing the performance between these two training approaches, the model trained through Pass@k outperforms the model trained on Pass@1. The reason behind it is that Pass@k Training encourages models to explore better solutions, which can be easily generalized to other tasks. In contrast, Pass@1 Training makes LLMs behave conservatively, thereby affecting LLMs' OOD performance.

\begin{tcolorbox}[
    colframe=takeaway,
    colback=white,
    coltitle=takeawayTitle,
]
\textcolor{takeawayTitle}{\textbf{Takeaway from Section~\ref{sec:generalization}}}

Pass@k Training exhibits stronger generalization ability than Pass@1 Training, achieving greater improvements over the base model in both in-domain and out-of-domain testing.
\end{tcolorbox}

\subsection{How does the Value of k Affect Pass@k Training?}
\label{sec:robustness}

\begin{figure}[h]
    \centering
    \subfloat[Training Reward.]{
        \includegraphics[width=0.48\linewidth]{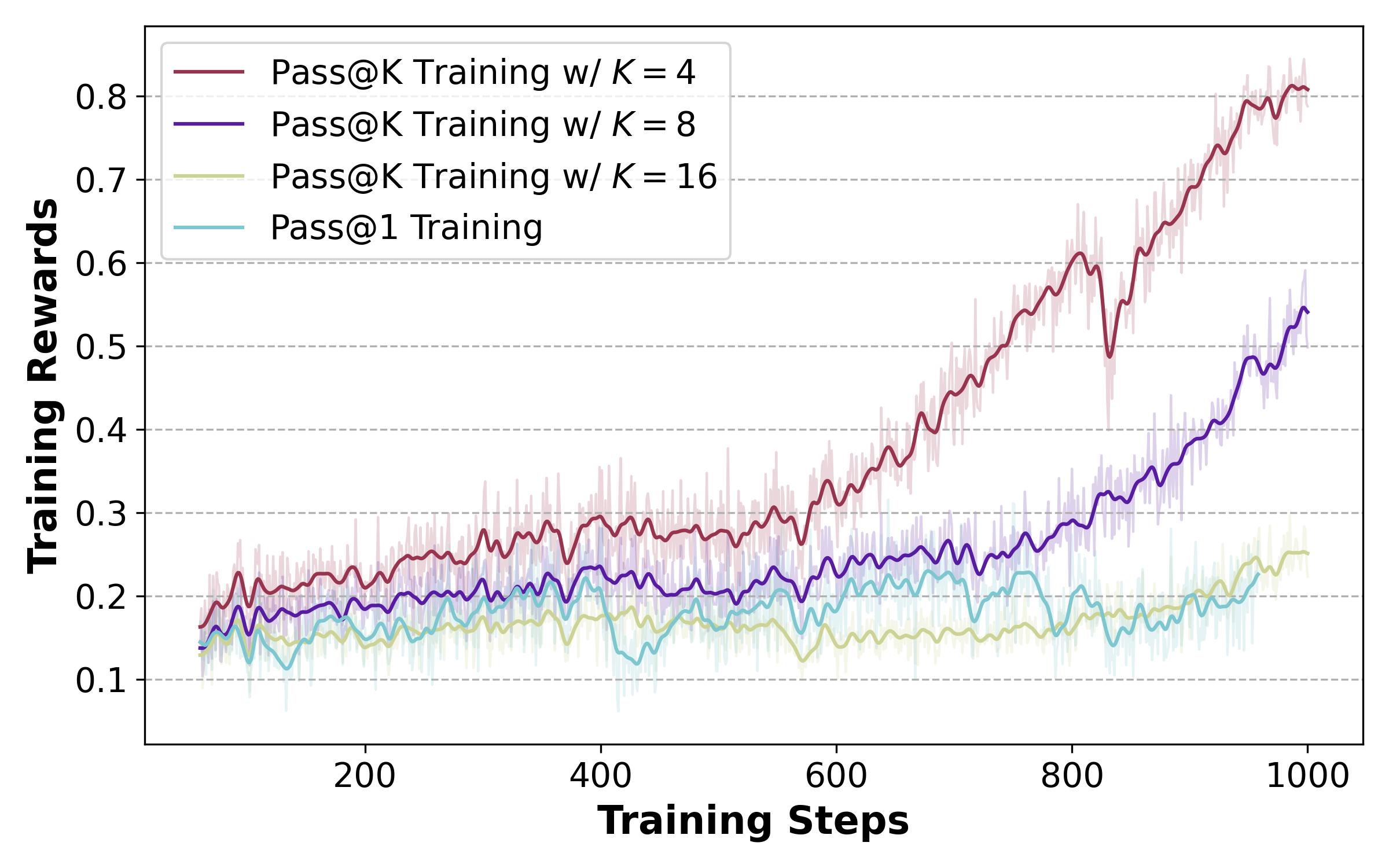}
        \label{fig:sec3.4-train_k}
    }
    \subfloat[Pass@k Performance on Test set.]{
        \includegraphics[width=0.48\linewidth]{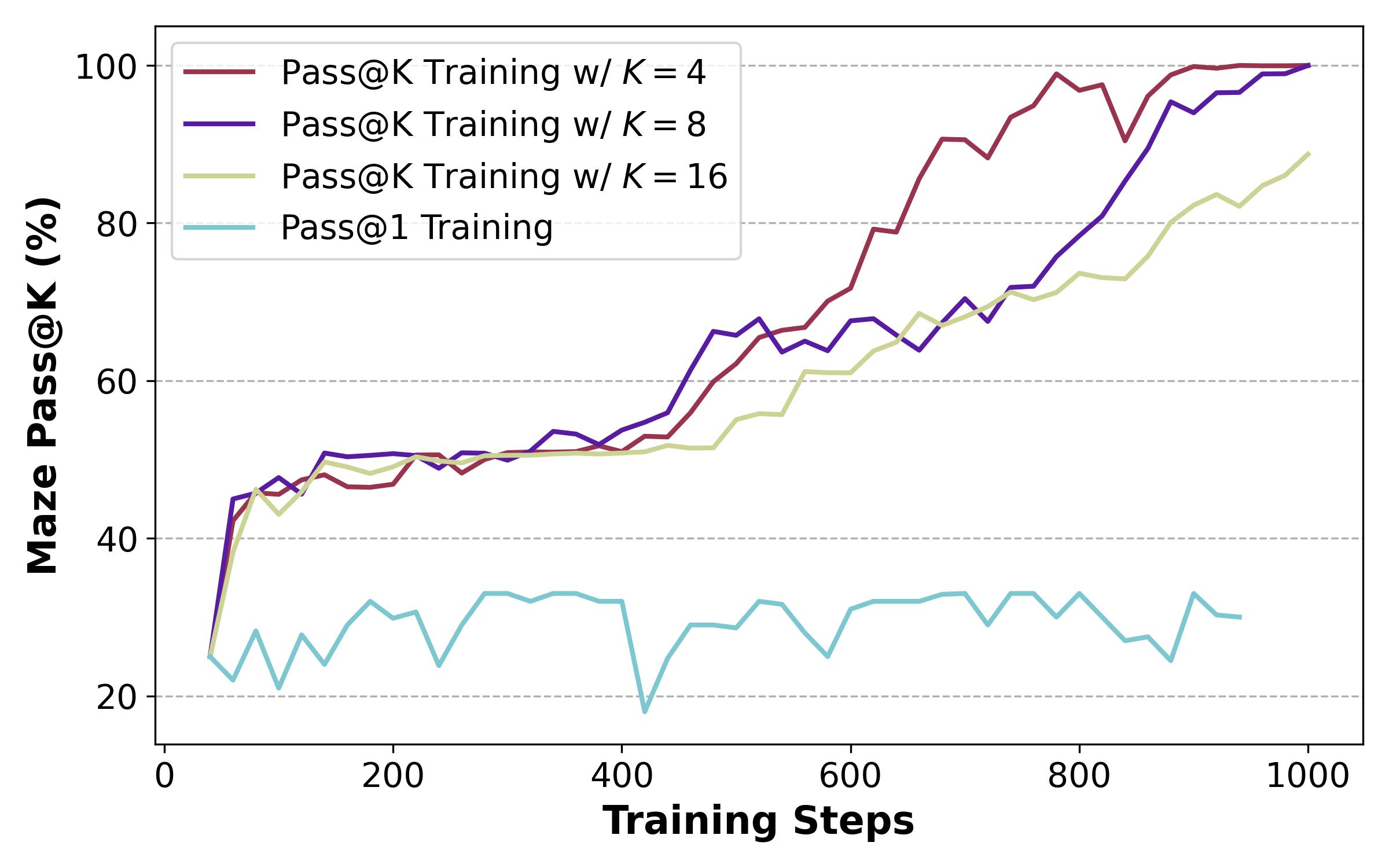}
        \label{fig:sec3.4-test_k}
    }
    \\
    \subfloat[Training Reward.]{
        \includegraphics[width=0.48\linewidth]{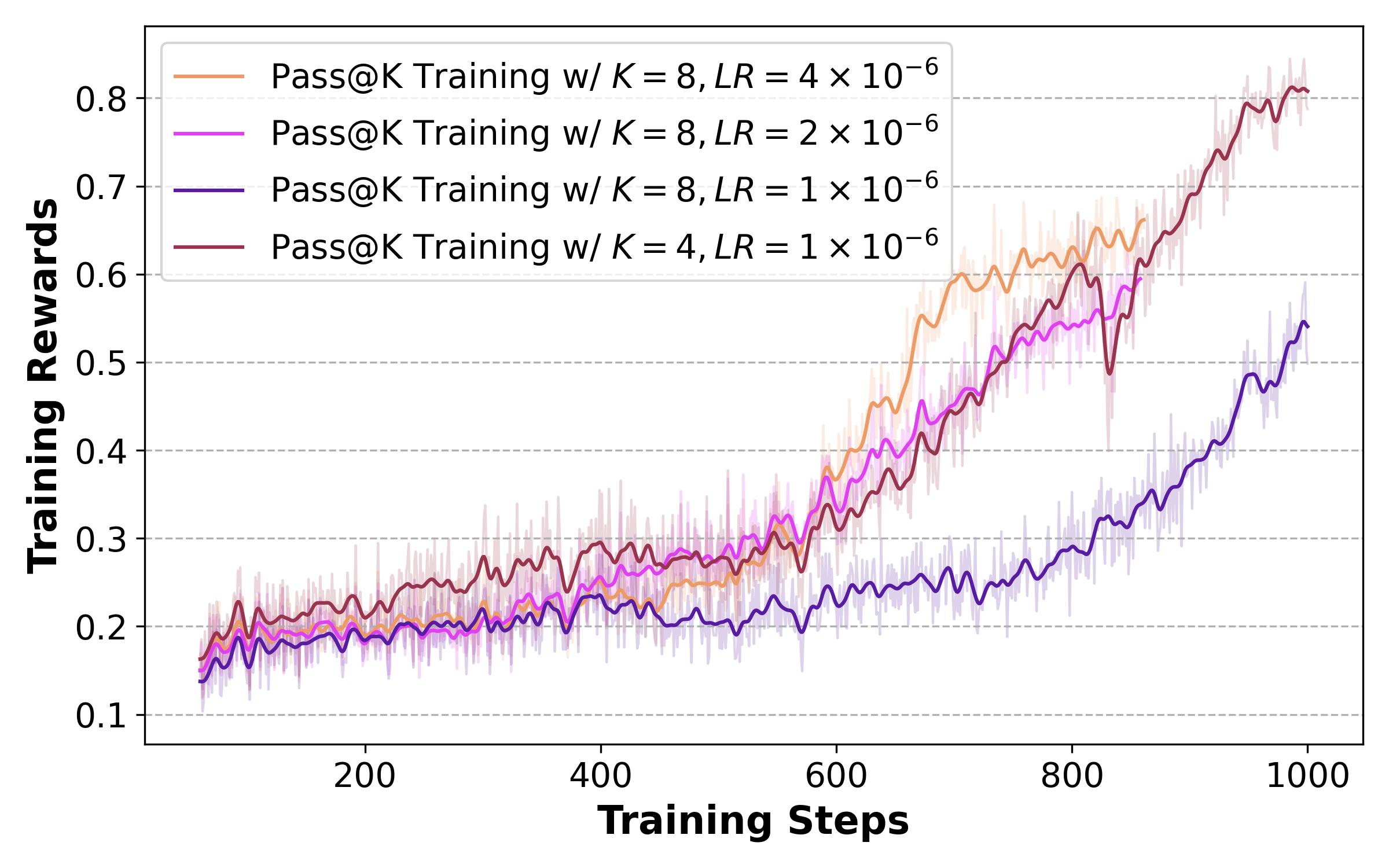}
        \label{fig:sec3.4-train_lr}
    }
    \subfloat[Pass@k Performance on Test set.]{
        \includegraphics[width=0.48\linewidth]{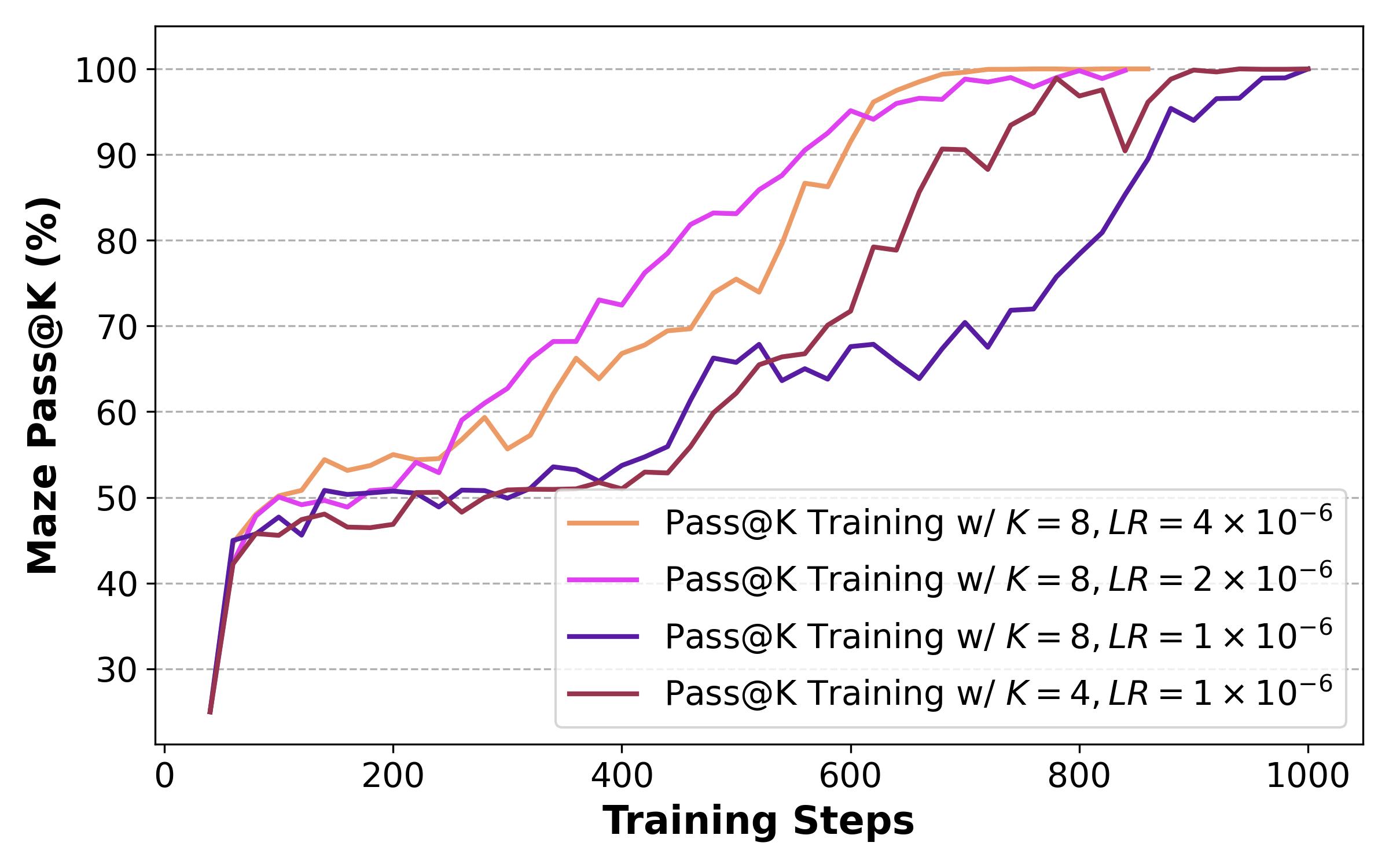}
        \label{fig:sec3.4-test_lr}
    }
    \caption{Training progress of Pass@k Training under various $k$ and learning rate (LR).}
    \label{fig:sec3.4}
\end{figure}

To analyze the robustness of Pass@k Training, we adjust the value of $k$ in $\{4,8,16\}$to perform RLVR training on Maze tasks and present the training reward and Pass@k performance of the testset in Figure~\ref{fig:sec3.4-train_k} and Figure~\ref{fig:sec3.4-test_k}, respectively. Whatever the value of $k$, the training rewards can be improved to a relatively high level as the training progresses, indicating the value of $k$ is not a vital factor that can help LLMs escape from the local optimum of Pass@1 Training. However, with the increase of $k$, the improvement slows down, affecting training efficiency. Through analyzing the analytical solutions of advantage values (\ie Eq.~\ref{eq:response_pos} and Eq.~\ref{eq:response_neg}), we can realize that a larger value of $k$ will bring a smaller value of advantage, resulting to a shorter optimization step, causing lower training efficiency.

Based on this phenomenon, we investigate whether scaling the learning rate (LR) to enlarge the optimization step can improve training efficiency. Following this idea, we employ the learning rate in $\{1\times 10^{-6},~2\times 10^{-6},~4\times 10^{-6}\}$ on the setting of $N=32$ and $k=8$, and present the results in Figure~\ref{fig:sec3.4-train_lr} and Figure~\ref{fig:sec3.4-test_lr}. With the increase of learning rate, the inflection point appears earlier, indicating higher training efficiency. The training efficiency of Pass@8 training even outperforms Pass@4 training when we employ  $4\times 10^{-6}$ as learning rate. These results have shown that the issues of training efficiency can be easily mitigated.

\begin{tcolorbox}[
    colframe=takeaway,
    colback=white,
    coltitle=takeawayTitle,
]
\textcolor{takeawayTitle}{\textbf{Takeaway from Section~\ref{sec:robustness}}}

Pass@k Training exhibits strong robustness to the choice of the value of $k$, leading to a stable and effective training process. Although there is a decline in the model's optimization efficiency as $k$ increases, this issue can be easily addressed by enlarging the learning rate.
\end{tcolorbox}

\subsection{Can the Benefits from Pass@k Training Be Transferred to Pass@1 Performance?}
\label{sec:passk_to_pass1}

\begin{table}[t]
    \centering
    \small
    \setlength{\tabcolsep}{4pt}
    \caption{Enigmata Pass@1/Pass@k Performance of Qwen2.5 models trained on different RLVR approaches. ``P@1 T.'' and ``P@k T.'' denote the Pass@1 Training and Pass@k Training with analytical derivation, respectively.}
      \begin{tabular}{lcccccccc}
      \toprule
       & \textbf{Crypto} & \textbf{Arithmetic} & \textbf{Logic} & \textbf{Grid} & \textbf{Graph} & \textbf{Search} & \textbf{Sequential} & \textbf{Overall} \\
      \midrule
      \multicolumn{9}{c}{\textbf{Closed-source LLMs (Pass@1)}} \\
      Grok-2-1212 & 10.1 & 9.4 & 50.0 & 12.8 & 17.6 & 3.9 & 6.4 & 13.6 \\
      GPT-4o-1120 & 26.2 & 1.9 & 34.5 & 17.8 & 19.3 & 6.0 & 3.9 & 14.2 \\
      Claude-3.7-Sonnet & 38.1 & 16.7 & 60.0 & 22.9 & 22.4 & 7.8 & 15.0 & 22.7 \\
      \midrule
      \multicolumn{9}{c}{\textbf{RLVR on Qwen2.5-7B-Instruct (Pass@1/Pass@k)}} \\
      Baseline & 0.1/0.7 & 1.0/3.3 & 28.1/48.4 & 3.7/9.2 & 3.0/11.6 & 0.3/1.0 & 2.7/5.4 & 4.7/10.1 \\
      + P@1 T. & 1.2/5.7 & 6.6/28.0 & 41.1/68.4 & 14.8/22.8 & 14.5/20.4 & 3.3/6.8 & 9.6/12.7 & 12.9/21.3 \\
      + P@k T. & 14.0/39.7 & 27.4/63.0 & 46.0/74.0 & 18.8/27.8 & 15.2/21.5 & 5.3/12.3 & 14.1/18.3 & 17.9/29.8 \\
      + P@k T. + P@1 T. & 96.9/98.3 & 36.2/67.7 & 49.3/71.8 & 30.9/37.5 & 20.3/30.7 & 25.8/37.5 & 10.6/12.9 & \textbf{30.8}/\textbf{40.6} \\
      \midrule
      \multicolumn{9}{c}{\textbf{RLVR on Qwen2.5-32B-Instruct (Pass@1/Pass@k)}} \\
      Baseline & 1.5/5.3 & 4.6/16.0 & 45.7/71.1 & 11.6/20.6 & 8.0/26.7 & 2.3/6.4 & 7.7/16.3 & 10.9/21.6 \\
      + P@1 T. & 95.8/99.7 & 53.0/85.0 & 76.6/92.2 & 38.4/47.4 & 44.4/57.8 & 47.0/58.8 & 21.8/25.8 & 45.2/56.0 \\
      + P@k T. & 93.8/99.3 & 51.1/86.3 & 74.8/92.4 & 39.0/49.6 & 42.7/61.3 & 45.9/59.9 & 21.5/26.6 & 44.5/57.4 \\
      + P@k T. + P@1 T. & 95.9/99.3 & 49.6/84.3 & 82.0/94.9 & 40.0/51.0 & 48.2/60.2 & 48.8/60.8 & 22.2/26.2 & \textbf{46.8}/\textbf{57.9} \\
      \bottomrule
      \end{tabular}
      \label{tab:main_qwen}
\end{table}

\begin{table}[t]
    \centering
    \small
    \caption{Pass@1/Pass@k Performance of Seed1.5-VL-Small (Internal Version) trained on different RLVR approaches. Seed1.5-VL-Small (Internal Version) is an MoE model that contains fewer parameters than Seed1.5-VL~\cite{6:journals/corr/abs-2505-07062}.}
      \begin{tabular}{lccc}
      \toprule
      \textbf{Pass@1/Pass@k} & \textbf{MathVision} & \textbf{MMMU} & \textbf{Avg.} \\
      \midrule
      Seed1.5-VL-Small (Internal Version) & 54.6/72.5 & 71.2/80.2 & 62.9/76.4 \\
      + Pass@1 Training & 55.3/74.0 & 72.0/83.7 & 63.7/78.9 \\
      + Pass@k Training & 53.9/75.6 & 72.0/84.3 & 63.0/80.0 \\
      + Pass@k Training + Pass@1 Training & 56.4/76.8 & 72.3/84.5 & \textbf{64.4}/\textbf{80.7} \\
      \bottomrule
      \end{tabular}
      \label{tab:main_seedvl}
\end{table}

To transfer the benefits brought by Pass@k Training to LLM Pass@1 performance, a native implementation is continually performing Pass@1 Training on the model, which is trained through Pass@k Training. We employ this approach in the RLVR training process and present the results of Qwen models on Puzzle tasks and Seed1.5-VL-Small (Internal Version) on multi-modal reasoning tasks in Table~\ref{tab:main_qwen} and Table~\ref{tab:main_seedvl}, respectively.
Moreover, to present the comprehensive evaluation, we also conduct the external experiments about different LLMs on Engimata and mathematical tasks (\eg AIME 2024~\cite{70:aime24} and AIME 2025~\cite{33:aime25}) in Appendix~\ref{sec:app_exp}.

First, the Pass@1 Training following the Pass@k Training can significantly improve the reasoning ability of LLMs, achieving remarkable Pass@1 performance. According to the results, we can observe that even the 7B model can surpass the powerful closed-source LLMs, including Grok-2, GPT-4o, and Claude-3.7-Sonnet. This might be because Pass@k Training enhances the exploration ability of the LLM, guiding it to escape from the local optimum and unleashing the potential of the LLM in the subsequent RLVR training.

Second, either the small-scale or large-scale LLMs (\eg Qwen2.5 with 7B or 32B parameters) can benefit from Pass@k Training. Besides, the model architecture and the model family do not influence the improvement of continual Pass@1 Training. The Qwen model is the dense model, while Seed1.5-VL-Small (Internal Version) is the MoE model. Their Pass@1 performance can be further improved after Pass@k Training.

Third, the domain and form of downstream tasks also do not affect the transfer from LLM Pass@k performance to their Pass@1 performance. Our evaluation includes synthetic puzzle tasks that are expressed in natural language, and multi-modal reasoning tasks that contain pictures in the problem description. These tasks require LLMs to possess different categories of abilities, and our Pass@k Training can specifically enhance the corresponding capabilities, showing strong effectiveness.

\begin{tcolorbox}[
    colframe=takeaway,
    colback=white,
    coltitle=takeawayTitle,
]
\textcolor{takeawayTitle}{\textbf{Takeaway from Section~\ref{sec:passk_to_pass1}}}

The benefits brought by Pass@k Training can be transferred to Pass@1 performance of LLMs, which is not affected by the scale of model parameters (e.g., 7B or 32B), model architecture (e.g, dense model or MoE model), model family (i.e., Qwen model or Seed model), or downstream tasks (natural language tasks or multi-modal tasks).
\end{tcolorbox}

%% file: sections/4-WhyPassK.tex
\section{Generalizing Pass@k Training via Implicit Reward Design}
\label{sec:why_passk}

As introduced in Section~\ref{sec:how_passk}, we achieve effective and efficient Pass@k Training by deriving the analytical form of the advantage function. In this section, we further investigate the key factors contributing to the success of Pass@k Training from an advantage perspective (Section~\ref{sec:difference_pass1_passk}). Moreover, the advantage function design can be viewed as a form of \textit{implicit reward design}. Motivated by this, we explore how to design advantage functions directly based on optimization objectives in scenarios where it is difficult to derive analytical solutions from the reward function (Section~\ref{sec:implicit_reward_design}).

\subsection{Difference Between Pass@1 and Pass@k Training}
\label{sec:difference_pass1_passk}

\subsubsection{Analysis Based on Advantage Value Curves}

\begin{figure}[t]
    \centering
    \subfloat[Pass@1 Training.]{
        \includegraphics[width=0.48\linewidth]{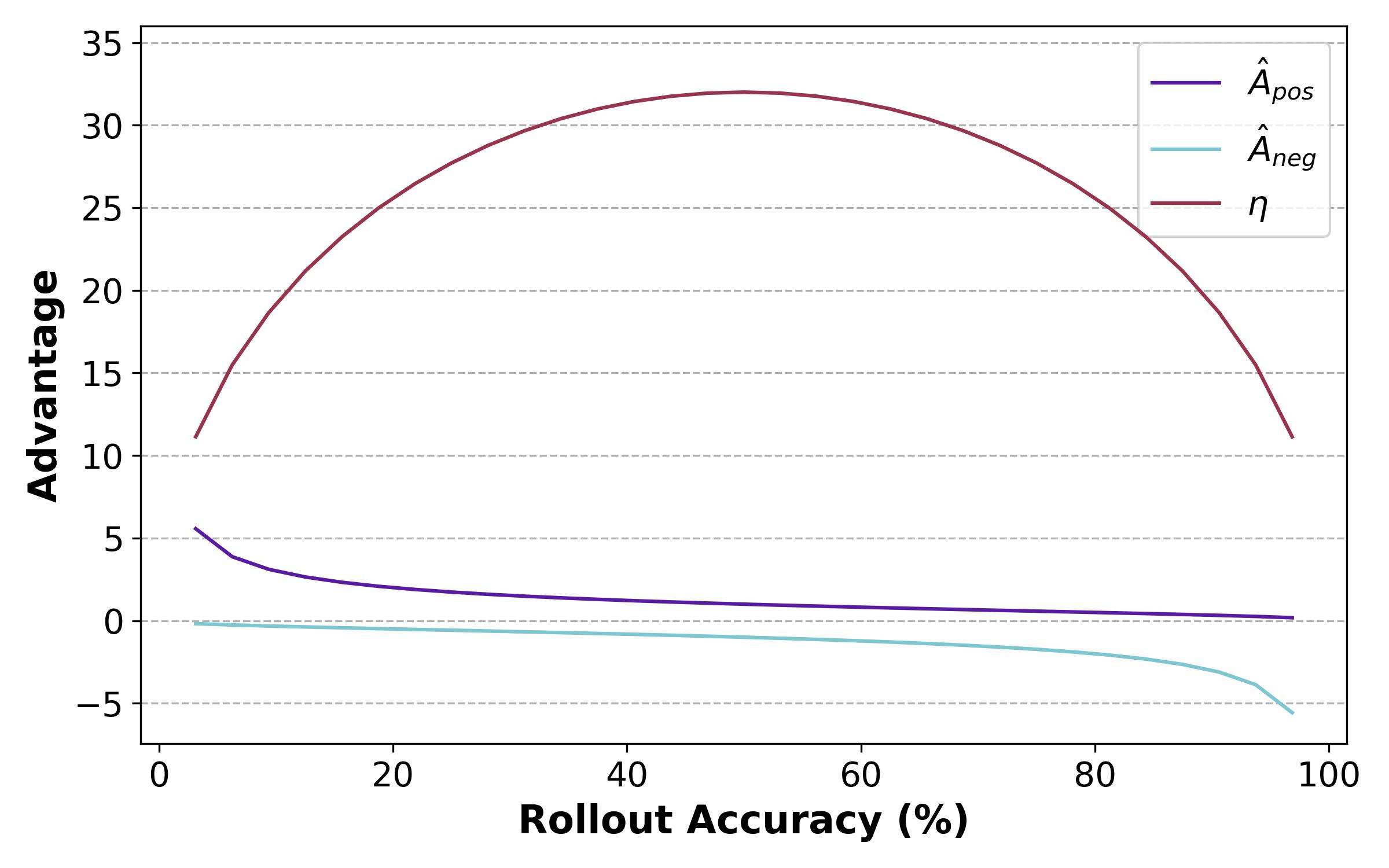}
        \label{fig:sec4-adv-vanilla}
    }
    \subfloat[Pass@k Training.]{
        \includegraphics[width=0.48\linewidth]{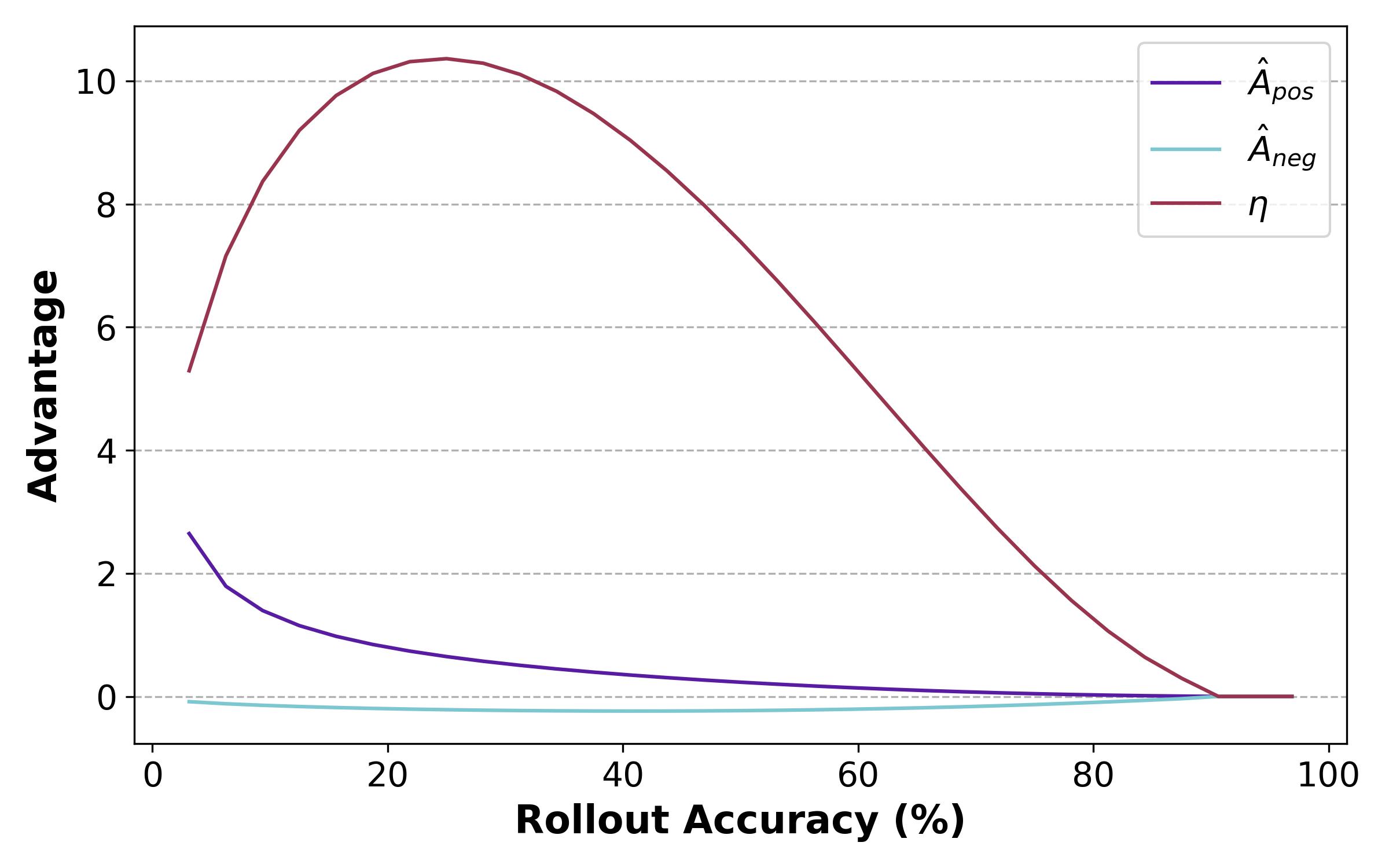}
        \label{fig:sec4-adv-pre_norm}
    }
    \caption{The curves of advantage function of Pass@1 Training and Pass@k Training on the setting of $N_\text{rollout}=32$.}
    \label{fig:sec4-adv}
\end{figure}

To analyze why Pass@k Training can help LLMs escape the local optimum, we first visualize the advantage curves of Pass@1 Training and Pass@k Training across responses with different correctness levels, as in GRPO and its variants, the advantage value depends solely on the correctness of the model's response. Furthermore, we observe that during the optimization process, the advantage value is directly multiplied by the gradient and can be interpreted as a scaling factor for the gradient. In this context, a larger absolute value of the advantage indicates a greater scaling of the gradient, and thus a larger update step for the corresponding sample. This implies that the model places greater optimization effort on samples with higher advantage magnitudes. Therefore, we argue that the absolute value of the advantage is also an important aspect worthy of investigation. Based on this insight, and to simplify the analysis, we compute the sum of the absolute advantage values across all responses, as defined below:
\begin{equation}
    \eta=N_\text{pos}\times|\hat{A}_\text{pos}|+N_\text{neg}\times|\hat{A}_\text{neg}|,
\end{equation}
The curves of $\eta$ (named as \textit{Sum of Absolute Advantage}) are added to our visualization and presented in Figure~\ref{fig:sec4-adv}. Comparing the curves of $\eta$of Pass@1 Training and Pass@k Training, we can observe that there are three major differences. 

\paratitle{Maximum of Sum of Absolute Advantage $\eta$.}
The maximum of $\eta$ of the Pass@1 Training approach is much higher than ones of Pass@k Training approach. As we discussed in Section~\ref{sec:robustness}, the maximum advantage values might affect training efficiency, and adding the coefficient on the loss function to adjust the advantage values can mitigate this issue. Thus, the maximum is not the critical factor that helps Pass@k Training outperform Pass@1 Training. 

\paratitle{Argmax of Sum of Absolute Advantage $\eta$.}
According to the curves in Figure~\ref{fig:sec4-adv}, the argmax of $\eta$ are significantly different between Pass@1 and Pass@8 Training. For Pass@1 Training, the maximum of $\eta$ appears at the position of 50\% accuracy (\ie $N_\text{pos}=0.5 \times N_\text{rollout}$), while the position of maximum of $\eta$ is 25\% accuracy (\ie $N_\text{pos}=0.25\times N_\text{rollout}$). This phenomenon suggests that Pass@k Training focuses on optimizing harder problems, while Pass@1 Training focuses on problems with medium difficulty. This further demonstrates that Pass@k Training tends to guide the model toward solving previously unsolved or difficult problems, rather than overfitting to those it has already mastered.

\paratitle{Trend of Sum of Absolute Advantage $\eta$.}
Another key difference between the function curves of Pass@1 and Pass@k Training lies in the trend of the function itself. In the $\eta$ curve of Pass@k Training, the value increases until it reaches a peak, and then gradually decreases to zero. Under this setting, when the problem is relatively easy (\ie the correctness is higher than 60\%), the optimization strength applied by the model (as indicated by the value of $\eta$) becomes smaller than that for harder problems. This further demonstrates that Pass@k Training focuses more on optimizing problems the model has not yet mastered. In contrast, during Pass@1 Training, the $\eta$ curve is symmetric around the point of maximum value, indicating that the training process allocates equal attention to both easy and hard problems.

\subsubsection{Analysis Based on Model Performance}

\begin{figure}[t]
    \centering
    \subfloat[Pass@1 Performance of Maze Tasks.]{
        \includegraphics[width=0.48\linewidth]{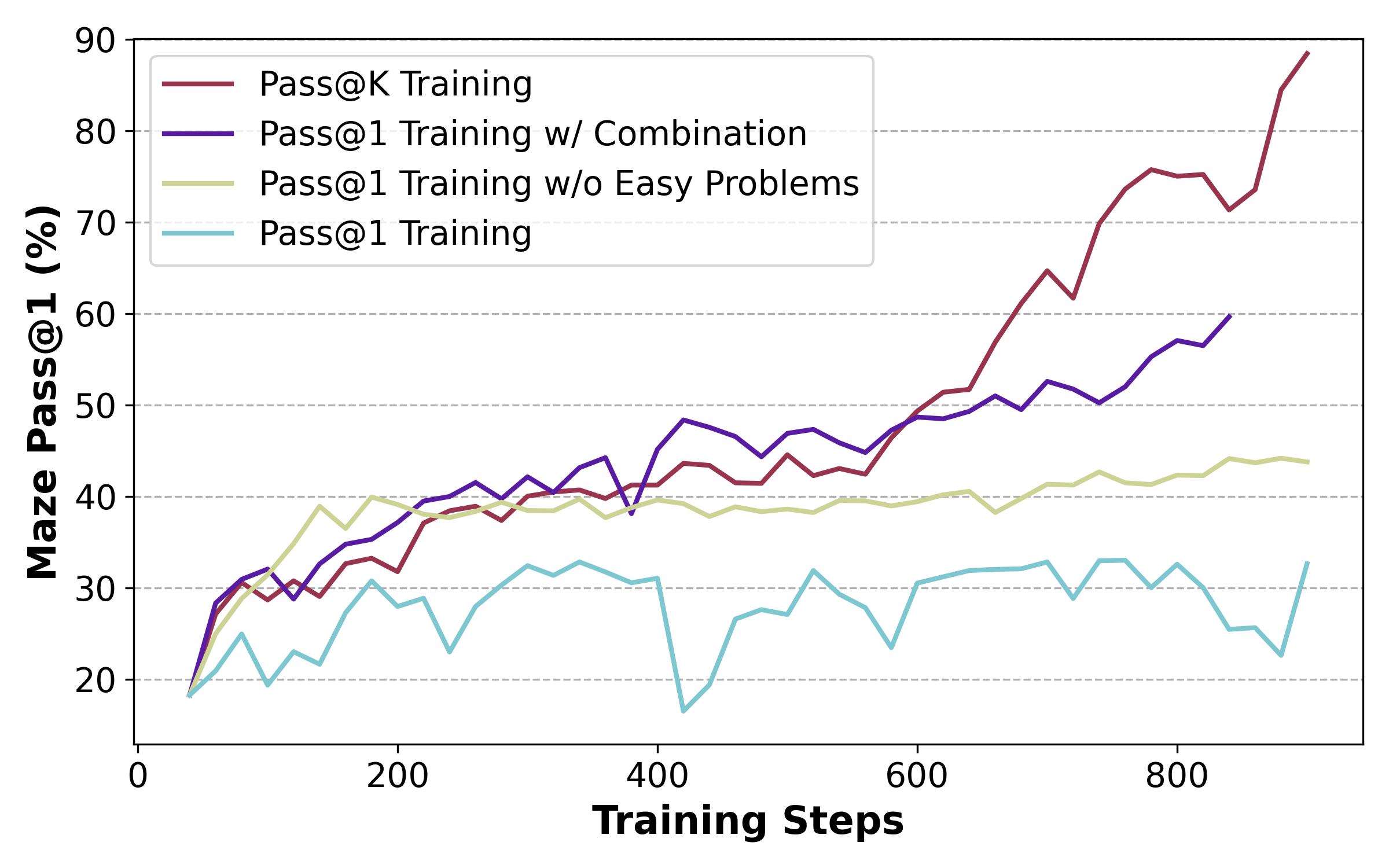}
        \label{fig:sec4.1.2-pass1}
    }
    \subfloat[Pass@k Performance of Maze Tasks.]{
        \includegraphics[width=0.48\linewidth]{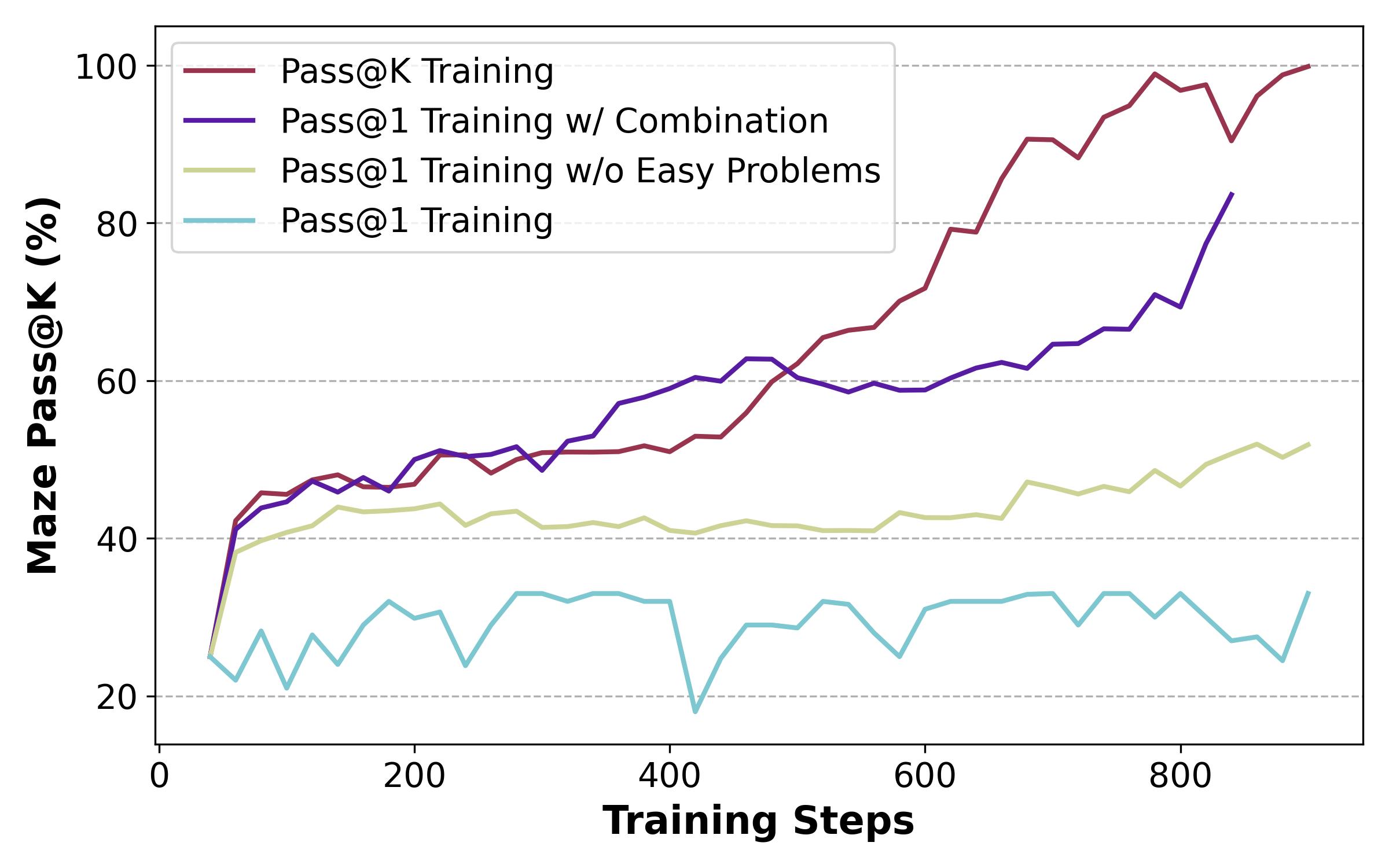}
        \label{fig:sec4.1.2-passk}
    }
    \caption{Training progress of various of Pass@1 Training and Pass@k Training on baseline setting.}
    \label{fig:sec4.1.2}
\end{figure}

As we discussed in the previous section, the effectiveness of the argmax and the trend of the sum of absolute advantage $\eta$ still remain unclear. Thus, in this section, we design the corresponding experiments to analyze their effectiveness, based on model performance. Additionally, we designed two training methods that serve as intermediates between Pass@1 and Pass@k Training, \ie removing the advantage values of the easy problems and combining the advantage estimation approaches of Pass@1 and Pass@k based on the accuracy of the current prompt. The curves of $\hat{A}_\text{pos}$, $\hat{A}_\text{neg}$, and $\eta$ of these four training approaches are presented in Figure~\ref{fig:adv-wo_easy} and Figure~\ref{fig:adv-segment}.

First, when the correctness of a response is high, we design the advantage function to decrease gradually toward zero. This setting allows the training reward to increase steadily during the optimization process, indicating that the model avoids getting stuck in a local optimum (\ie the blue line and purple line). When this optimization is removed, the reward on the training set fails to continue increasing, suggesting that the model has already converged to a local optimum and is no longer learning new knowledge during the RLVR process (\ie the red line and green line). This phenomenon suggests that excessive learning from easy examples is a key factor causing the model to fall into local optima. Therefore, reducing the degree of learning from easy questions can help prevent the model from getting trapped in such suboptimal solutions.

Second, simply setting the reward for easy questions to zero is not sufficient to effectively prevent the model from over-optimizing on them; it merely delays the point at which the model falls into a local optimum. As shown in Figure~\ref{fig:sec4.1.2}, removing the optimization for easy questions (represented by the red line) leads to higher training rewards and better test performance compared to the baseline (represented by the green line). However, both curves exhibit similar trends: after an initial phase of improvement, model performance gradually plateaus, making further progress difficult.

Third, regarding the choice of the argmax position of the $\eta$ function, a comparison of the curves in Figure~\ref{fig:sec4.1.2} reveals that shifting the argmax forward leads to higher optimization efficiency. Specifically, the model is able to escape from local optima more quickly, and a turning point in training reward appears earlier. This phenomenon suggests that hard problems contribute more significantly to model improvement and yield better optimization effects. Assigning greater optimization strength to harder problems can thus effectively enhance training efficiency, allowing the model to achieve better performance with fewer training steps.

Based on the above results and discussions, several preliminary conclusions can be made, \ie the argmax of $\eta$influences the training efficiency and the trend of $\eta$ prevents model from falling into local optimum. Besides, it is important to note that this is only our initial conclusion. More comprehensive experiments tailored to specific tasks and scenarios are required for further validation.

\begin{tcolorbox}[
    colframe=takeaway,
    colback=white,
    coltitle=takeawayTitle,
]
\textcolor{takeawayTitle}{\textbf{Takeaway from Section~\ref{sec:difference_pass1_passk}}}

In the RLVR training process, simple problems might easily lead to overfitting. Appropriately reducing the optimization strength for these problems helps prevent the model from getting stuck in local optima, thereby achieving better overall performance.
\end{tcolorbox}

\subsection{RLVR Training Through Implicit Reward Design}
\label{sec:implicit_reward_design}

Building upon the analysis of the curve properties of advantage values in the previous section, we now explore preliminary modifications to the advantage function in this section, \ie implicit reward design. Our goal is to explore the potential of implicit reward design and to propose several promising directions for future research.

\subsubsection{Exceeding Pass@k Training}

\begin{figure}[t]
    \centering
    \subfloat[Pass@1 Performance of Maze Tasks.]{
        \includegraphics[width=0.48\linewidth]{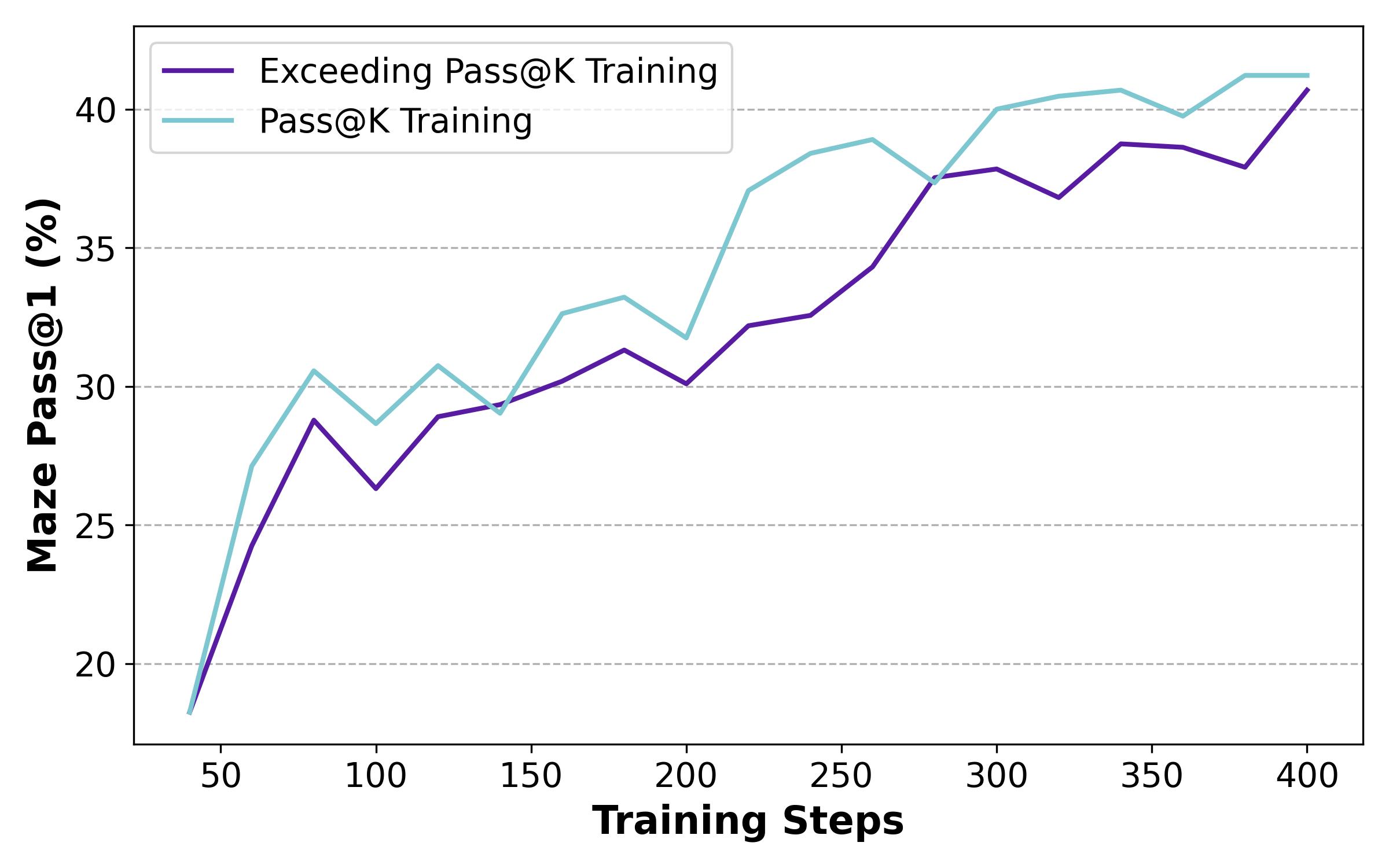}
        \label{fig:sec4.2.1-pass1}
    }
    \subfloat[Pass@k Performance of Maze Tasks.]{
        \includegraphics[width=0.48\linewidth]{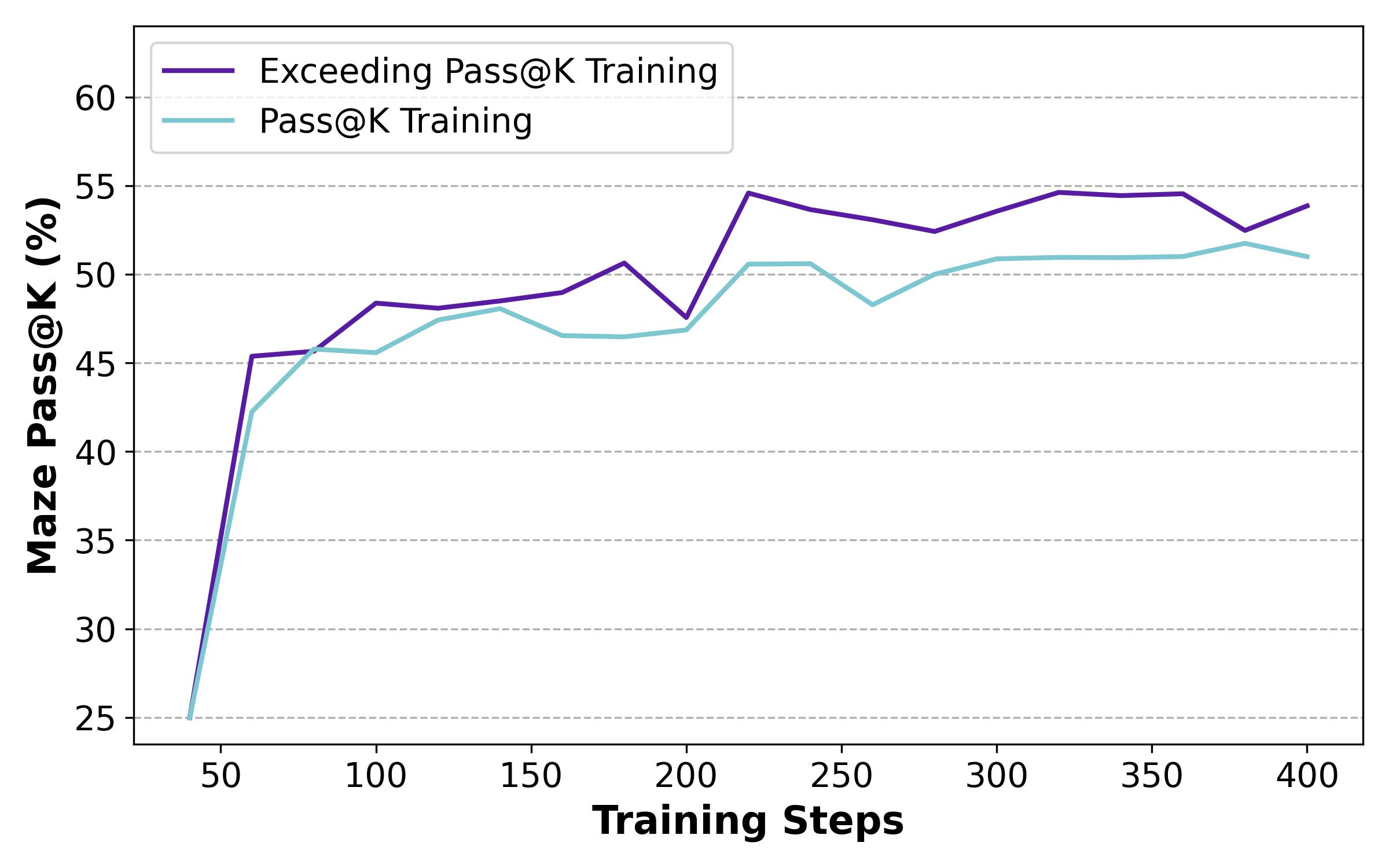}
        \label{fig:sec4.2.1-passk}
    }
    \caption{Training progress of Pass@k Training and Exceeding Pass@k Training on baseline setting.}
    \label{fig:sec4.2.1}
\end{figure}

In previous discussion, we have found that the position of maximum value of $\eta$will influence the training objective (focus on Pass@1 or Pass@k). Based on these observations and conclusions, we hypothesize that an earlier peak in the $\eta$ function leads to better optimization performance in Pass@k Training. To test this hypothesis, we design a transformation function as follows:
\begin{equation}
    f(N_\text{pos}) = \frac{4}{10\log(N_\text{pos}+0.5)},~~\hat{A}'=f(N_\text{pos})\times \hat{A}.
\end{equation}
The advantage value curve after applying the transformation function is shown in Figure~\ref{fig:adv-log}. We observe that, in the transformed curve, the peak of the $\eta$ function is shifted forward to the position where the correctness is $\frac{1}{32}$. According to our hypothesis, such a modification of the advantage function is expected to result in better optimization performance for Pass@k Training. We integrate this transformed function into the RLVR training process (named as \textit{Exceeding Pass@k Training}), and the corresponding training results are presented in Figure ~\ref{fig:sec4.2.1}.

From the experimental results, we observe that Exceeding Pass@k Training can effectively improve the model's Pass@k performance during the early training stage. However, since this method places excessive emphasis on difficult problems, the improvement in Pass@1 performance on downstream tasks progresses more slowly.  Based on these observations and analyses, we hypothesize that the computation of advantage values could be adaptively adjusted according to the model's current state. We leave this as a direction for future work.

\subsubsection{Combination of Pass@1 and Pass@k Training}

\begin{figure}[t]
    \centering
    \subfloat[Pass@1 Performance of Puzzle Tasks.]{
        \includegraphics[width=0.48\linewidth]{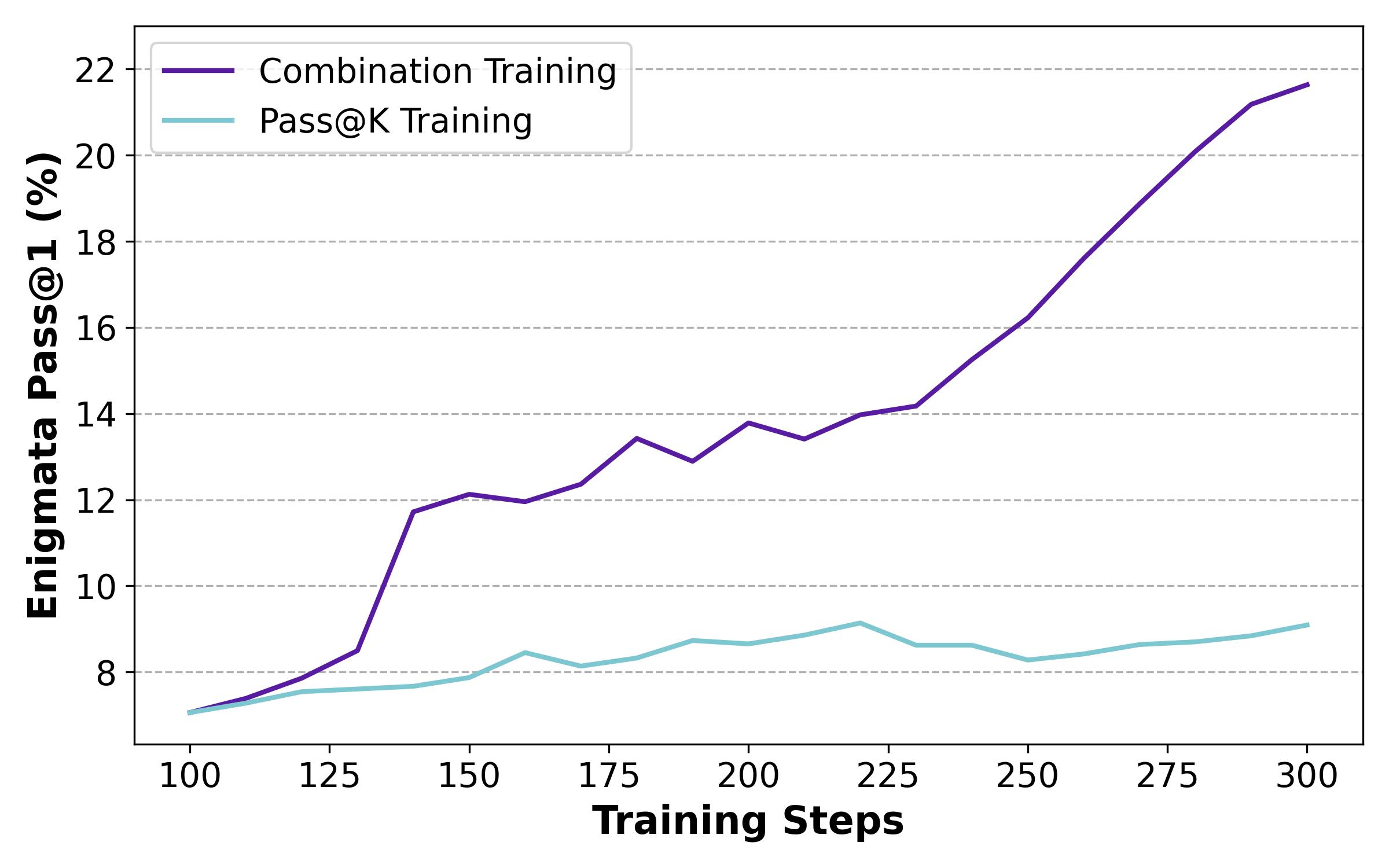}
        \label{fig:sec4.2.2-pass1}
    }
    \subfloat[Pass@k Performance of Puzzle Tasks.]{
        \includegraphics[width=0.48\linewidth]{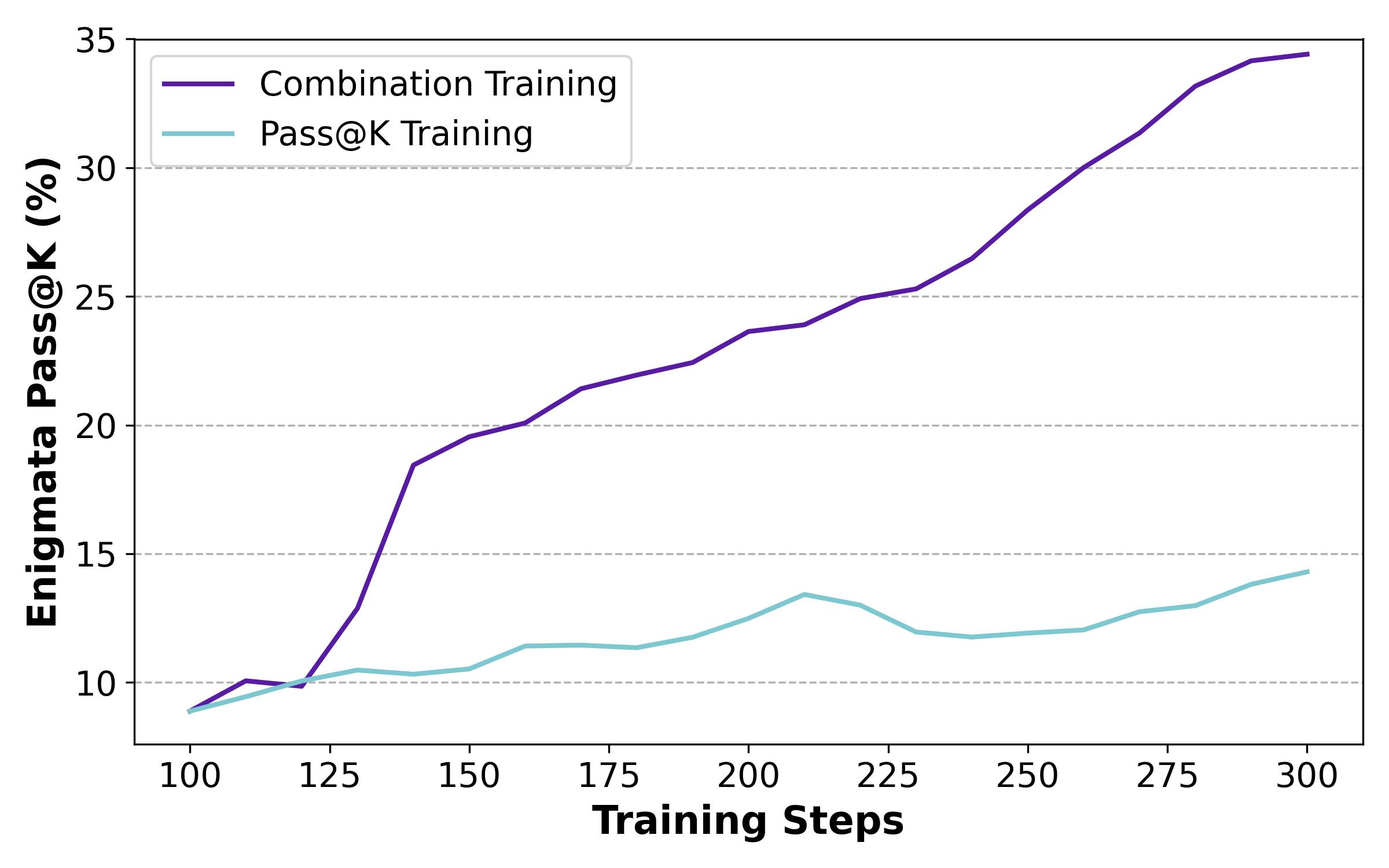}
        \label{fig:sec4.2.2-passk}
    }
    \caption{Training progress of Pass@k Training and Combination Training on baseline setting.}
    \label{fig:sec4.2.2}
\end{figure}

From the previous analysis, we observe that Pass@k Training focuses more on optimizing harder problems, and prevents the model from overfitting to the easy problems. Motivated by this observation, we consider whether combining Pass@1 and Pass@k Training could be beneficial. Thus, we design the following formula to estimate the final advantage value:
\begin{equation}
    \hat{A}=\frac{N_\text{pos}}{N}\times \hat{A}_\text{Pass@k}+(1-\frac{N_\text{pos}}{N})\times \hat{A}_\text{Pass@1},
\end{equation}
Where $\hat{A}_\text{Pass@k}$ and $\hat{A}_\text{Pass@k}$ denote the advantage values estimated by Pass@k and Pass@1 Training approach, respectively. In the above formula (named as \textit{Combination Training}), when the sampled response has a low correctness score, the advantage value from Pass@1 Training receives a higher weight and dominates the training process, leading to high training efficiency. Conversely, when the correctness is high, the advantage value from Pass@k Training is assigned a greater weight, thereby avoiding LLMs from overfitting to the problems that it has already mastered.

In Figure~\ref{fig:sec4.2.2}, we present the training results of the Qwen series models on the Enigmata benchmark. We observe that, for both Pass@1 and Pass@8 metrics, models trained with Combination Training consistently outperform those trained with standard Pass@k Training. During the Combination Training process, model performance improves rapidly and maintains a high growth rate. In contrast, Pass@k Training leads to slower performance gains. This is because difficult problems require extensive exploration for the model to learn effectively, making rapid improvement challenging. At the same time, easy problems receive low but sufficient optimization strength during training. These two factors together contribute to the lower optimization efficiency of Pass@k Training compared to Combination Training. The above analysis further supports the idea that adapting the advantage function based on the model's current state can effectively enhance model performance.

\subsubsection{Adaptive Training based on Policy Entropy}
\label{sec:adaptive_training}

\begin{figure}[t]
    \centering
    \subfloat[Pass@1 Performance of Puzzle Tasks.]{
        \includegraphics[width=0.48\linewidth]{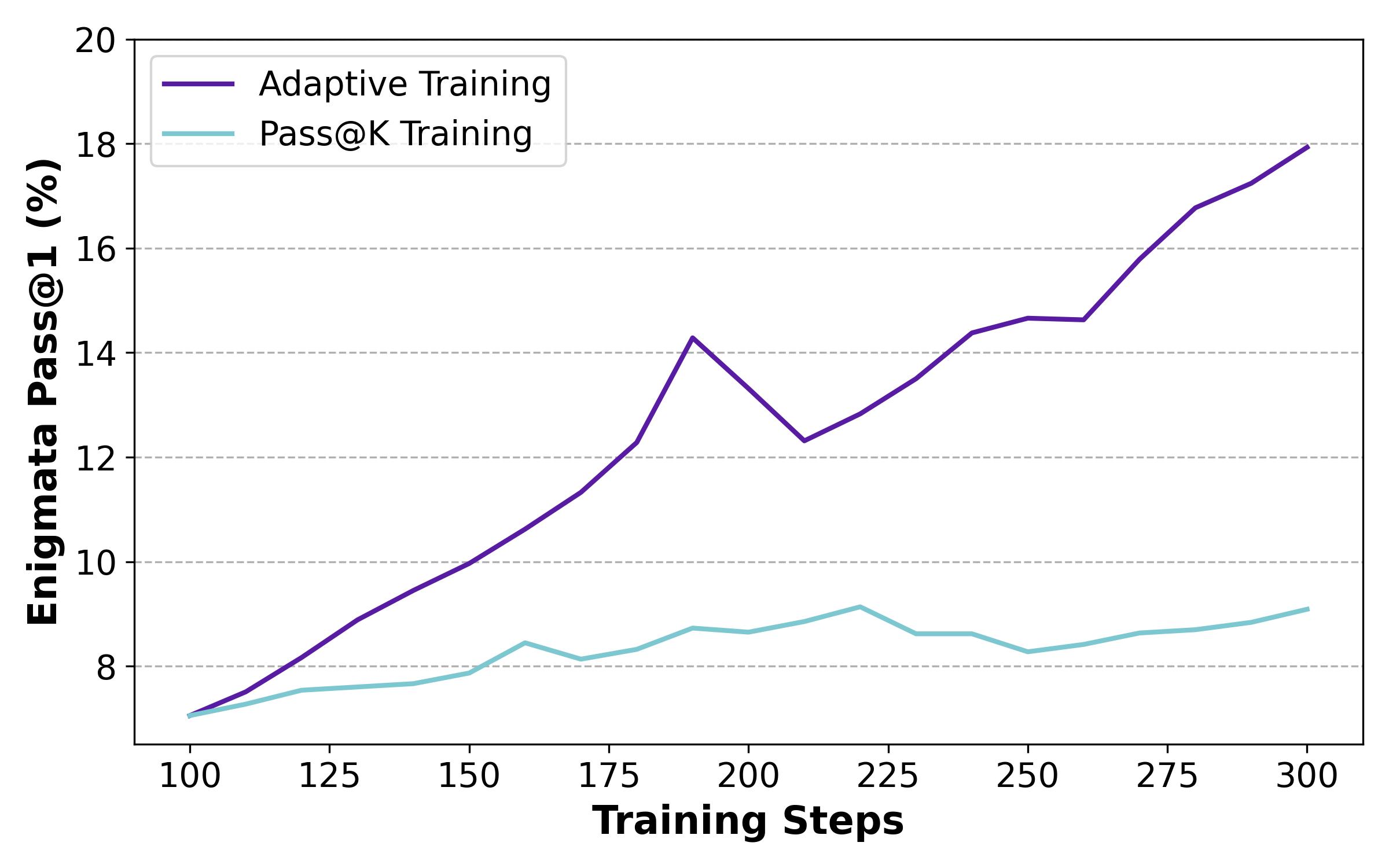}
        \label{fig:sec4.2.3-pass1}
    }
    \subfloat[Pass@k Performance of Puzzle Tasks.]{
        \includegraphics[width=0.48\linewidth]{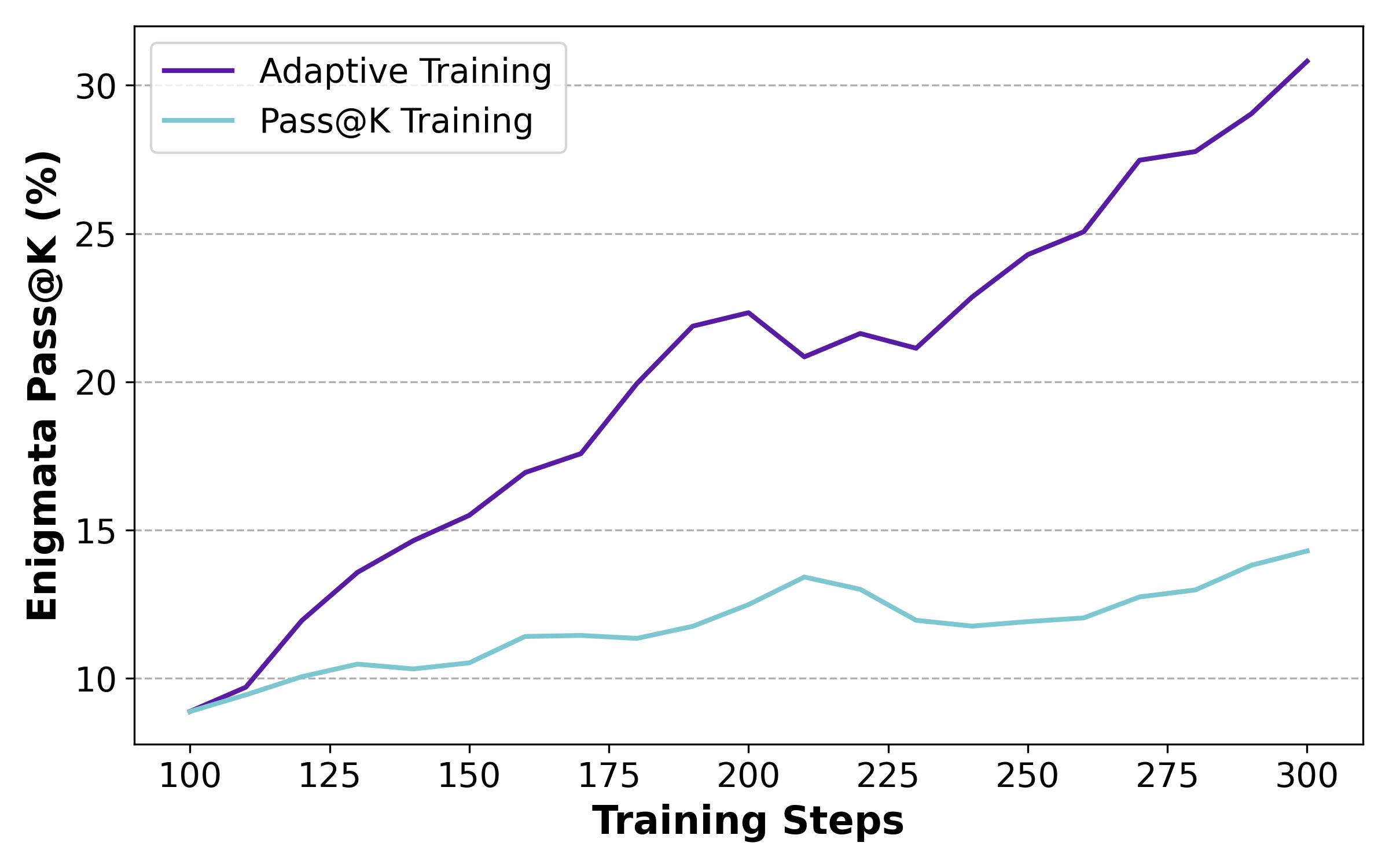}
        \label{fig:sec4.2.3-passk}
    }
    \caption{Training progress of Pass@k Training and Adaptive Training on baseline setting.}
    \label{fig:sec4.2.3}
\end{figure}

Building on the insights from the previous section, we explore whether the training objective can be adaptively adjusted throughout the RLVR process.
Besides, as discussed in previous work~\cite{62:journals/corr/abs-2506-01939,19:journals/corr/abs-2505-22617}, the entropy of policy distribution can indicate its exploration ability.
Thus, we conduct the Pass@k Training based on the guidance of policy entropy (named as \textit{Adaptive Training}).
Concretely, we first compute the average entropy $\bar{E}$ of the sampled responses of each problem, and then rank each problem based on its $\bar{E}$. We designate the top 50\% as \textit{high-exploration problems} and the rest as \textit{low-exploration problems}. 
For high-exploration problems, we use the Pass@1 advantage function to help the model exploit prior exploration. 
For low-exploration problems, we apply the Pass@K advantage function to encourage further exploration.
This approach uses policy entropy to guide advantage computation, allowing us to combine the strengths of different training strategies.

We present the experimental results in Figure~\ref{fig:sec4.2.3}.
Experimental results show that under Adaptive Training, the model achieves effective improvements in both Pass@1 and Pass@K performance, outperforming both Pass@1 Training and Pass@K Training.
This indicates that Pass@1 and Pass@K training are complementary. By designing a proper adaptation mechanism, it is possible to better leverage the strengths of both training methods, enabling the model to achieve improved performance on downstream tasks.
This also confirms that the entropy of the policy distribution can serve as an indicator of the model’s exploration ability, and that it integrates well with Pass@K training. Using entropy as a monitoring signal to adjust RLVR training yields better results than directly using it as a training objective.

\begin{tcolorbox}[
    colframe=takeaway,
    colback=white,
    coltitle=takeawayTitle,
]
\textcolor{takeawayTitle}{\textbf{Takeaway from Section~\ref{sec:implicit_reward_design}}}

Implicit reward design allows for better control over the optimization process, avoiding the complex theoretical derivation. Concretely, increasing the optimization strength for more difficult problems can effectively enhance the model’s ability to solve them (i.e., Pass@k performance), and combining or dynamically adjusting different forms of advantage estimation make it possible to improve both exploration and exploitation capabilities simultaneously.
\end{tcolorbox}

%% file: sections/5-relatedwork.tex
\section{Related Work}

\subsection{Reinforcement Learning with Verifiable Rewards}

To unleash the potential of LLM reasoning ability, DeepSeek directly employs reinforcement learning with verifiable rewards (RLVR) on DeepSeek-V3, obtaining the large reasoning model DeepSeek-R1-Zero~\cite{7:journals/corr/abs-2501-12948}, which can perform the reasoning process with complex reasoning actions (\eg reflection and verification). Given the success of the DeepSeek-R1, a surge of studies~\cite{8:journals/corr/abs-2503-04548,11:journals/corr/abs-2503-18892,12:journals/corr/abs-2503-24290} have explored the effectiveness of RLVR on the popular open-source LLMs, like Qwen~\cite{35:journals/corr/abs-2412-15115}, Mistral~\cite{36:journals/corr/abs-2310-06825}, and LLaMA~\cite{37:journals/corr/abs-2407-21783}. Moreover, the RLVR training paradigm can help LLMs to control their reasoning time~\cite{38:journals/corr/abs-2503-04697}, switch the reasoning pattern~\cite{39:journals/corr/abs-2505-21097,40:journals/corr/abs-2505-20258}, enhance the specific performance metric~\cite{41:journals/corr/abs-2503-19595}, and enhance their abilities without supervision~\cite{43:journals/corr/abs-2504-16084,44:journals/corr/abs-2505-20633}. However, recent work points out that the popular RLVR algorithms (\eg PPO~\cite{45:journals/corr/SchulmanWDRK17} and GRPO~\cite{4:journals/corr/abs-2402-03300}) still face serious challenges, like training instability, model collapse, and reward noise~\cite{42:journals/corr/abs-2505-15201,46:journals/corr/abs-2503-20783,47:journals/corr/abs-2503-14476,48:journals/corr/abs-2505-24864,49:journals/corr/abs-2506-01347}. To mitigate these issues, existing researches propose the optimization on the rollout strategy~\cite{47:journals/corr/abs-2503-14476}, objective function design~\cite{42:journals/corr/abs-2505-15201,46:journals/corr/abs-2503-20783,48:journals/corr/abs-2505-24864}, and data selection~\cite{49:journals/corr/abs-2506-01347}. Specifically, previous work~\cite{42:journals/corr/abs-2505-15201} utilizes Pass@k as the reward on the policy gradient algorithm~\cite{68:journals/ml/Williams92} to encourage models to solve hard problems. However, the intrinsic connection between Pass@k RLVR training and LLM exploration ability has not been fully recognized. Thus, we further adopt the Pass@k metric in GRPO and its variants through three approaches (Figure~\ref{fig:sec2.4}), and derive the analytical solution of advantage values of Pass@k reward in RLVR training. Moreover, according to empirical experiments and theoretical analysis, we discuss the benefits of Pass@k Training  in balancing the exploration and exploitation abilities of LLMs during the RLVR training procedure, showing the huge potential of Pass@k RLVR training and pointing out the promising future research directions.

\subsection{Effective Exploration in Test-time Scaling}

Recently, test-time scaling has been proposed to improve the performance of LLMs by consuming more computational resources at inference time~\cite{54:journals/corr/abs-2503-24235}. Since the LLMs continuously leverages exploration-derived experience to optimize its performance, effective exploration is important and necessary during the test-time scaling process~\cite{14:journals/corr/abs-2501-11651,59:journals/corr/abs-2505-07787}. However, existing work reveals that the exploration ability is limited by the corresponding base model, hindering the continuous scaling of model performance~\cite{55:journals/corr/abs-2504-13837}. To mitigate this issue, previous work proposed several approaches, including achieved by adjusting the sampling hyper-parameters~\cite{8:journals/corr/abs-2503-04548,12:journals/corr/abs-2503-24290,14:journals/corr/abs-2501-11651}, performing self-verification and self-reflection~\cite{56:journals/corr/abs-2506-10406,57:journals/corr/abs-2505-13445,58:journals/corr/abs-2505-04842}, or leveraging external models to verify the reasoning process~\cite{60:journals/corr/abs-2504-02495,61:journals/corr/abs-2505-15034}. Beyond these approaches from an external perspective of the model, it is equally important to explore the model’s exploration capability through its internal mechanisms. Current studies start from the perspective of the entropy of policy distribution, pointing out that entropy can indicate the exploration ability of LLMs~\cite{13:journals/corr/abs-2506-14758,19:journals/corr/abs-2505-22617} and high-entropy tokens are vital for model optimization~\cite{62:journals/corr/abs-2506-01939}. Based on these findings, training the critical tokens~\cite{62:journals/corr/abs-2506-01939} and adding regularization~\cite{14:journals/corr/abs-2501-11651,48:journals/corr/abs-2505-24864} are employed in the RLVR training process to avoid the degradation of the exploration capability of LLMs. Further, several studies focus on enhancing the exploration abilities of LLMs by selecting useful sampled experience~\cite{49:journals/corr/abs-2506-01347,63:journals/corr/abs-2506-09026}, integrating entropy into advantage estimation~\cite{19:journals/corr/abs-2505-22617}. 

%% file: sections/6-conclusion.tex
\section{Conclusion}

In this work, we proposed the Pass@k Training method within the RLVR framework, aiming to enable mutual improvement between the exploration and exploitation capabilities of the LLM, thereby pushing the limits of its overall performance. We first demonstrated that using Pass@k as the reward can effectively enhance the model’s ability to explore diverse outputs, which in turn improves its exploitation capability. Next, to improve training efficiency and effectiveness, we introduce the bootstrap sampling mechanism and analytical derivation of the advantage function to optimize the Pass@k Training procedure. After that, to better understand the inner mechanism of Pass@k Training, we proposed five research questions from different aspects to answer why the Pass@k Training works and what benefits can be brought from the Pass@k Training. 

Moreover, inspired by the effectiveness of Pass@k Training, we further analyzed it from the perspective of implicit reward design. By examining the curves of advantage value, we preliminarily identified two key factors contributing to the success of Pass@k Training, \ie the argmax and the trend of the sum of absolute advantage $\eta$. Building on these insights, we conducted an initial exploration into designing customized advantage functions to further improve model performance. This exploration is relatively preliminary, but it has shown remarkable effectiveness. We consider it a promising direction for future research.

%% file: sections/appendix.tex
\section{Experiment Setup}

\subsection{Details of Downstream Tasks}

In this section, we present detailed information of each downstream evaluation task. 

\begin{figure}[h]
    \centering
    \includegraphics[width=1\linewidth]{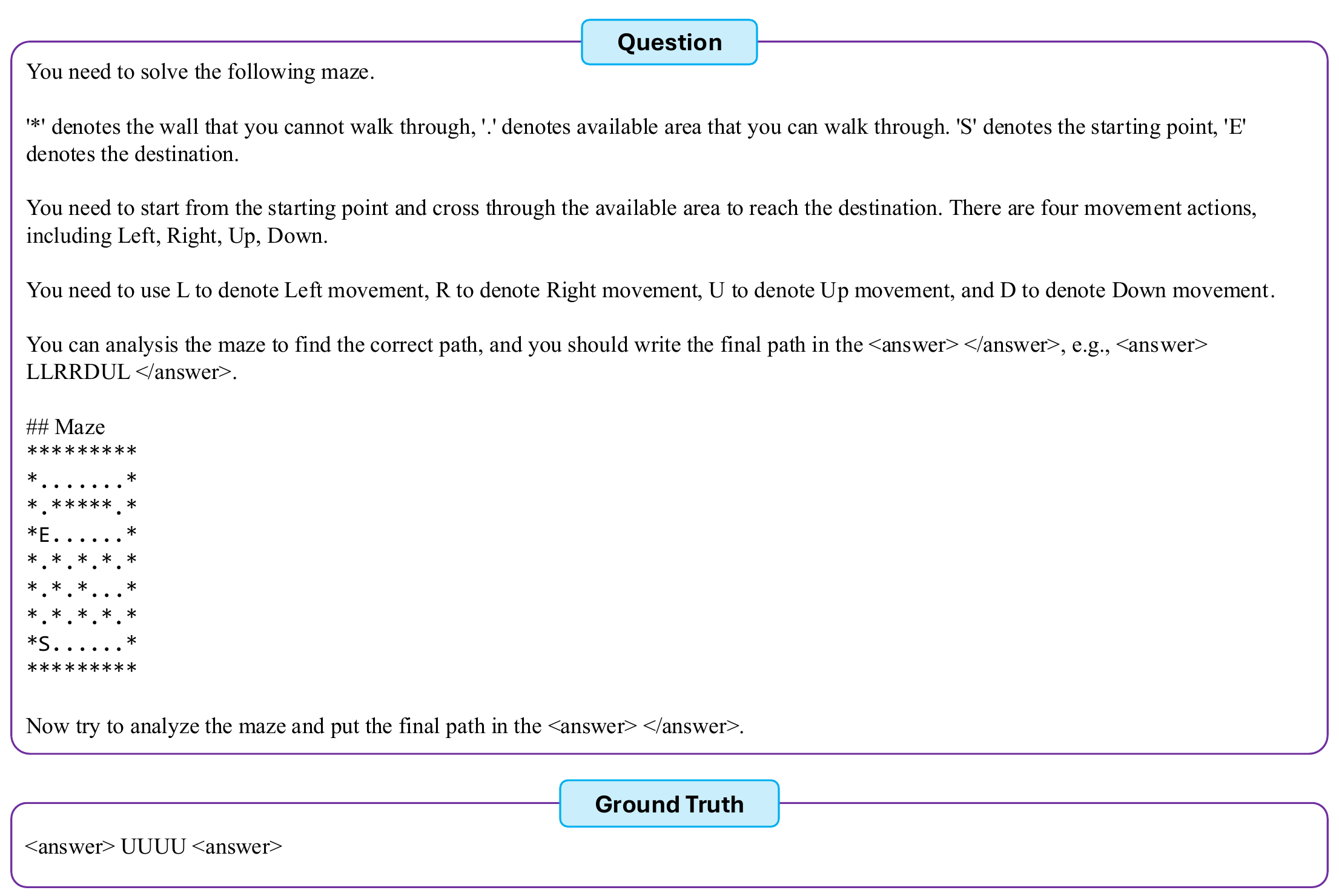}
    \caption{An example of \textbf{Maze} Task.}
    \label{fig:example_maze}
\end{figure}

\paratitle{Maze.}
We follow the framework proposed by previous work to synthesize the different sizes of mazes. Each maze is represented by text, containing $n$ rows and $n$ columns, a total of $n*n$ characters. Concretely, each of them is one of the following four characters ``S'', ``E'', ``*'', and ``.'', denoting the start point, destination, available place, and unavailable place, respectively. Given the maze, LLMs can first generate the thought or reasoning process and then generate the final answer, which includes one of the four actions ``U'', ``D'', ``L'', and ``R'', refering to moving up, down, left, and right, respectively. For training data, we construct the mazes with sizes of $9\times9$, $11\times11$, $13\times13$, and $15\times15$, to increase the diversity of training data. For test data, to evaluate the generalization of the RLVR process, we not only conduct the same sizes of the mazes with the training dataset, but also collect the mazes with sizes of $7\times7$, $17\times17$, $19\times19$, and $21\times21$. To ensure the validity of the experiment, we performed strict deduplication operations after generating the training and test data. The statistical information of the datasets is presented in Table~\ref{tab:maze_data}. For better understanding, we present a test instance in Figure~\ref{fig:example_maze}.
To present the empirical insights more clearly, we only showed the results of the $9\times9$ maze in the above text, and the remaining results are presented in Appendix~\ref{app:maze}.

\begin{table}[h]
    \centering
    \small
    \caption{The statistical information of the Maze task.}
      \begin{tabular}{lcccccccc}
      \toprule
      & {$7 \times 7$} & {$9 \times 9$} & {$11 \times 11$} & {$13 \times 13$} & {$15 \times 15$} & {$17 \times 17$} & {$19 \times 19$} & {$21 \times 21$} \\
      \midrule
      Training Set & - & 10,000 & 10,000 & 10,000 & 10,000 & - & - & -  \\
      Test Set & 75 & 100 & 100 & 100 & 100 & 100 & 100 & 100 \\
      \bottomrule
      \end{tabular}
      \label{tab:maze_data}
\end{table}

\begin{figure}[h]
    \centering
    \includegraphics[width=1\linewidth]{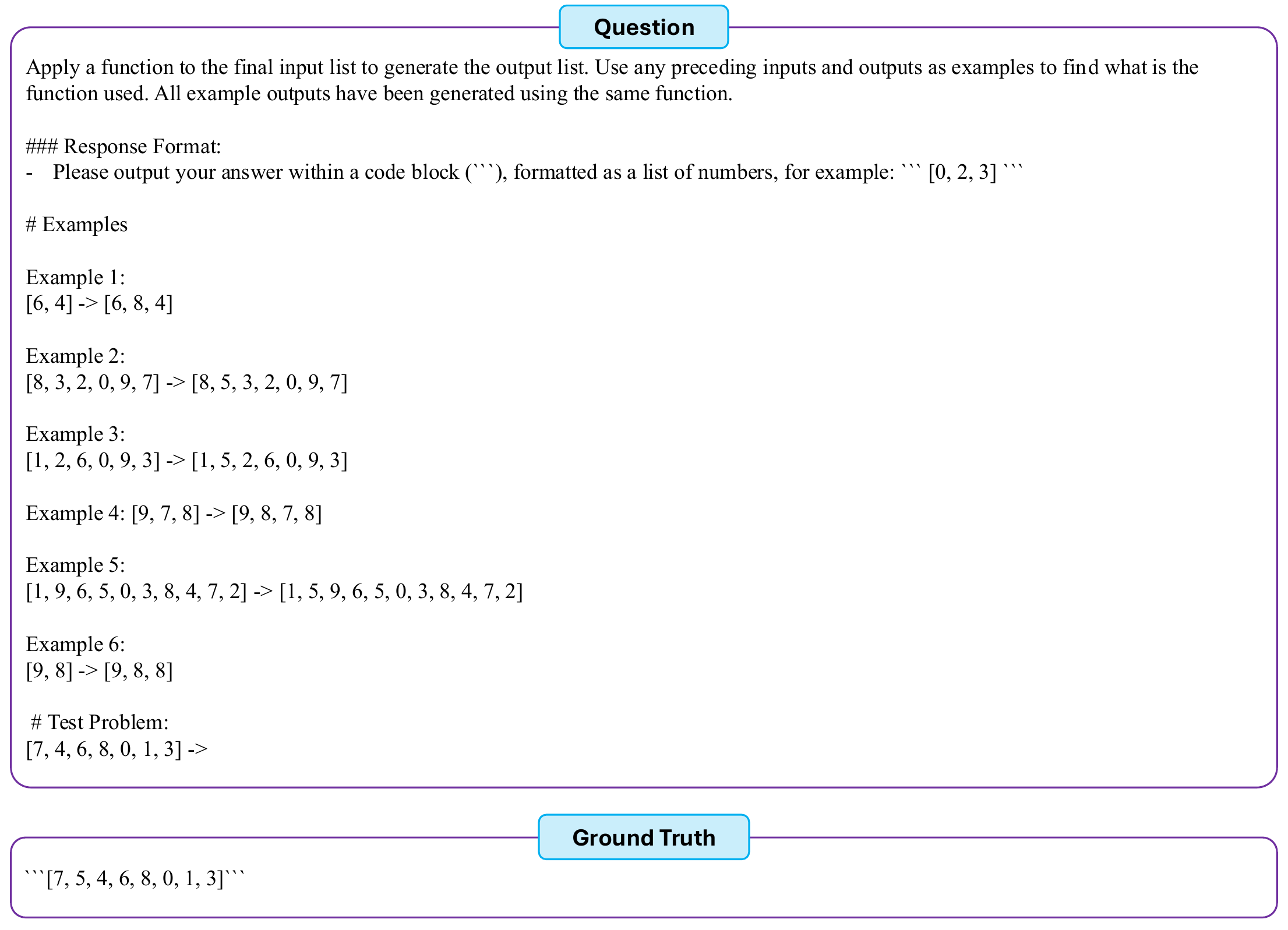}
    \caption{An example of \textbf{Enigmata} Task.}
    \label{fig:example_enigmata}
\end{figure}

\paratitle{Enigmata.}
To assess the reasoning and logical abilities of LLMs, Enigmata proposed a comprehensive benchmark that includes the 36 categories of synthetic verifiable puzzles of 7 primary categories, including Crypto Puzzle, Arithmetic Puzzle, Logic Puzzle, Grid Puzzle, Graph Puzzle, Search Puzzle, and Sequential Puzzle. Each category can assess different abilities of LLMs. For better understanding, we present a test instance in Figure~\ref{fig:example_enigmata}.

\begin{figure}[h]
    \centering
    \includegraphics[width=1\linewidth]{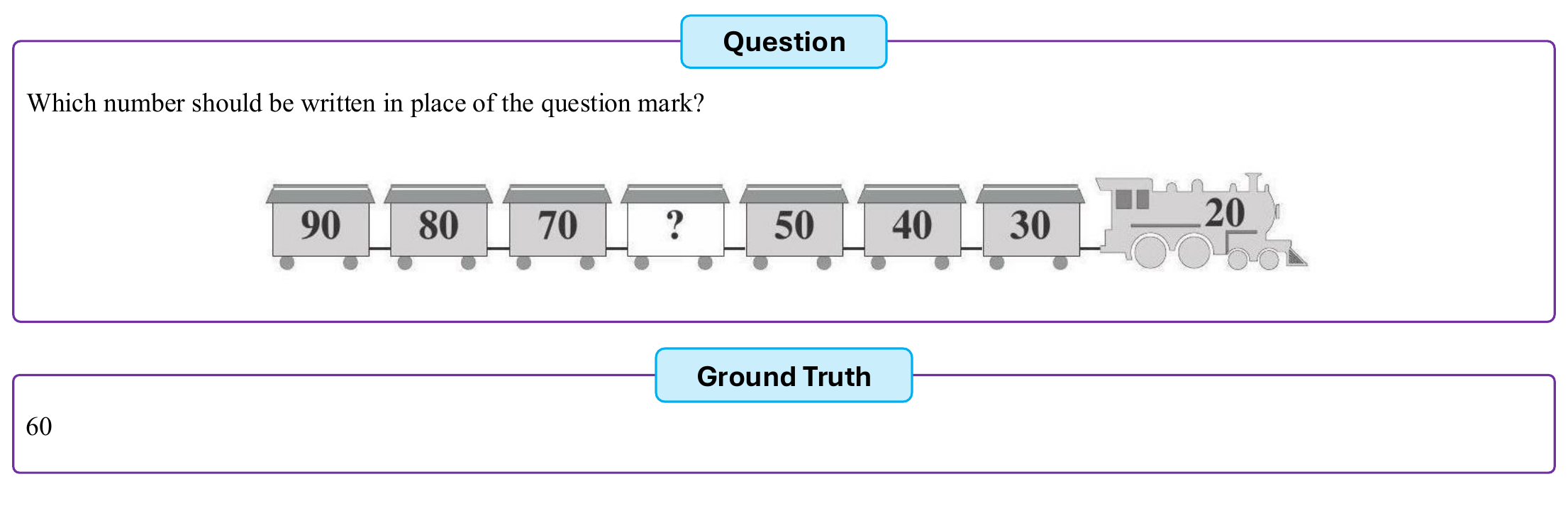}
    \caption{An example of \textbf{MathVision} Task.}
    \label{fig:example_mathvision}
\end{figure}

\paratitle{MathVision.}
MathVision selects 3,040 high-quality problems from human math competitions, each accompanied by relevant images. Solving these problems requires both careful interpretation of the visual information and rigorous mathematical reasoning. MathVision provides a benchmark for assessing a model’s multimodal understanding as well as its ability to perform rigorous mathematical reasoning. For better understanding, we present a test instance in Figure~\ref{fig:example_mathvision}.

\begin{figure}[h]
    \centering
    \includegraphics[width=1\linewidth]{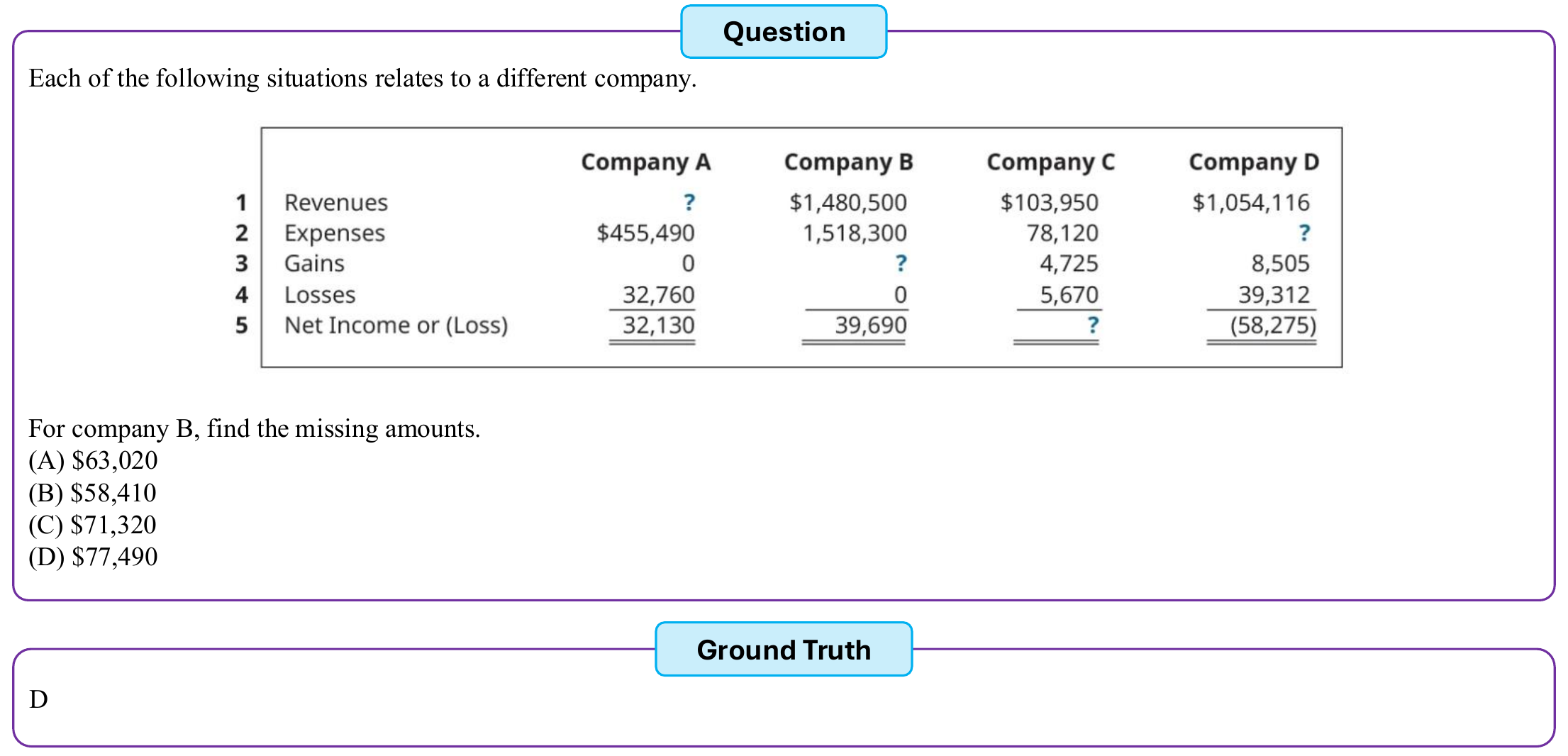}
    \caption{An example of \textbf{MMMU} Tasks.}
    \label{fig:example_mmmu}
\end{figure}

\paratitle{MMMU.}
MMMU includes college-level reasoning and comprehension tasks across six academic subjects, including Art \& Design, Business, Science, Health \& Medicine, Humanities \& Social Science, and Tech \& Engineering. Moreover, MMMU includes a wide range of image types, enabling a comprehensive assessment of a model's capability to process and reason over different forms of visual information. For better understanding, we present a test instance in Figure~\ref{fig:example_mmmu}.

\subsection{Implementation Details}
\label{sec:implementation_details}

\paratitle{Training.}
In our experiment, we adapt Qwen2.5-7B-Instruct and Qwen2.5-32B-Instruct as the backbone model and train it through DAPO. To enhance the efficiency of the training process, we only retain the clip-higher (i.e., $\varepsilon_\text{low}=0.2$ and $\varepsilon_\text{high}=0.28$) and token-level policy gradient loss, and remove other optimizations. For the training hyper-parameters, we set the learning rate for the policy model as $1\times10^{-6}$ with 10 warmup steps, and employ $128$, $32$, and $32$ as prompt batch size $BS_\text{prompt}$, mini-batch size $BS_\text{mini}$, and rollout times $n_\text{rollout}$, respectively. For the reward, the responses that pass the verification (named as positive responses) will be assigned the positive reward $R_{\text{pos}}=1$, while the other responses (named as negative responses) will be endowed with the negative reward $R_{\text{neg}}=0$. Additionally, we do not employ any regularization methods, such as KL or Entropy regularization.

\paratitle{Evaluation.}
To evaluate the performance of LLMs, we adopt 1.0 and 0.95 as the Temperature and Top\_P. For each question, we sample 32 responses from LLMs for the Maze task and sample 8 responses from LLMs for other tasks, and then utilize the sampled response to compute the Pass@1 and Pass@k scores.

\section{Details of Analytical Derivation}

We present the details of the analytical derivation procedure mentioned in Section~\ref{sec:passk_w_analytical_derivation}, including the derivation of the average of the group reward, standard deviation of the group reward, and response-relative advantage.

\subsection{Derivation of the Average of Group Reward}

\begin{align}
    \bar{R}^\text{group}&=\frac{1}{N^\text{group}_\text{total}}\times\left(N^\text{group}_\text{pos} \times R_\text{pos} + N^\text{group}_\text{neg} \times R_\text{neg}\right) \\
    &=\frac{1}{\binom{N_\text{rollout}}{K}}\times\left(\left(\binom{N_\text{rollout}}{K}-\binom{N_\text{neg}}{K}\right)\times 1+\left(\binom{N_\text{neg}}{K}\right)\times0\right) \\
    &=1-\frac{\binom{N_\text{neg}}{K}}{\binom{N_\text{rollout}}{K}}.
\end{align}

\subsection{Derivation of the Standard Deviation of Group Reward}

\begin{align}
    \sigma^\text{group}&=\sqrt{\frac{1}{N_\text{rollout}}\times\left(N^\text{group}_\text{pos}\times\left(\bar{R}^\text{group}-R_\text{pos}\right)^2+N^\text{group}_\text{neg}\times\left(\bar{R}^\text{group}-R_\text{neg}\right)^2\right)} \\
        &=\sqrt{\frac{\left(\binom{N_\text{rollout}}{K}-\binom{N_\text{neg}}{K}\right)\times\left(1-\frac{\binom{N_\text{neg}}{K}}{\binom{N_\text{bootstrap}}{K}}-1\right)^2+\binom{N_\text{neg}}{K}\times\left(1-\frac{\binom{N_\text{neg}}{K}}{\binom{N_\text{rollout}}{K}}-0\right)^2}{\binom{N_\text{rollout}}{K}}} \\
        &=\sqrt{\left(1-\frac{\binom{N_\text{neg}}{K}}{\binom{N_\text{bootstrap}}{K}}\right)\times\left(\frac{\binom{N_\text{neg}}{K}}{\binom{N_\text{rollout}}{K}}\right)^2+\frac{\binom{N_\text{neg}}{K}}{\binom{N_\text{rollout}}{K}}\times\left(1-\frac{\binom{N_\text{neg}}{K}}{\binom{N_\text{rollout}}{K}}\right)^2} \\
        &=\sqrt{\left(1-\frac{\binom{N_\text{neg}}{K}}{\binom{N_\text{rollout}}{K}}\right)\times\left(\frac{\binom{N_\text{neg}}{K}}{\binom{N_\text{rollout}}{K}}\right)\times\left(1-\frac{\binom{N_\text{neg}}{K}}{\binom{N_\text{rollout}}{K}}+\frac{\binom{N_\text{neg}}{K}}{\binom{N_\text{rollout}}{K}}\right)} \\
        &=\sqrt{\left(1-\frac{\binom{N_\text{neg}}{K}}{\binom{N_\text{rollout}}{K}}\right)\times\left(\frac{\binom{N_\text{neg}}{K}}{\binom{N_\text{rollout}}{K}}\right)} \\
&=\sqrt{\bar{R}^\text{group}\times\left(1-\bar{R}^\text{group}\right)}.
\end{align}

\subsection{Derivation of the Response-Relative Advantage}

\begin{align}
    \hat{A}_\text{pos} &= \frac{1}{\binom{N_\text{rollout}-1}{K-1}} \times \left(\binom{N_\text{rollout}-1}{K-1}\times \hat{A}^\text{group}_{pos} + 0\times\hat{A}^\text{group}_{neg}\right)\\
&=\frac{1-\bar{R}^\text{group}}{\sigma^\text{group}}. \\
\hat{A}_\text{neg}&=\frac{1}{\binom{N_\text{rollout}-1}{K-1}}\times \left(\left(\binom{N_\text{rollout}-1}{K-1}-\binom{N_\text{neg}-1}{K-1}\right)\times \hat{A}^\text{group}_{pos} + \binom{N_\text{neg}-1}{K-1}\times\hat{A}^\text{group}_{neg}\right) \\
&=\left(\left(1-\frac{\binom{N_\text{neg}-1}{K-1}}{\binom{N_\text{rollout}-1}{K-1}} \right)\times\hat{A}^\text{group}_\text{pos}+\frac{\binom{N_\text{neg}-1}{K-1}}{\binom{N_\text{rollout}-1}{K-1}}\times \hat{A}^\text{group}_\text{neg} \right)\\
&=\left(\left(1-\frac{\binom{N_\text{neg}-1}{K-1}}{\binom{N_\text{rollout}-1}{K-1}} \right)\times\frac{1-\bar{R}^\text{group}}{\sigma^\text{group}}+\frac{\binom{N_\text{neg}-1}{K-1}}{\binom{N_\text{rollout}-1}{K-1}}\times \left(-\frac{\bar{R}^\text{group}}{\sigma^\text{group}}\right) \right)\\
&=\left(1-\bar{R}^\text{group}-\frac{\binom{N_\text{neg}-1}{K-1}}{\binom{N_\text{rollout}-1}{K-1}}\right)\times\left(\sigma^\text{group}\right)^{-1}.
\end{align}

\section{Pseudo Code for Pass@k Training}
\label{app:code}

We present the pseudo code for Pass@k Training with full sampling (Algorithm~\ref{code:full_sampling}), bootstrap sampling (Algorithm~\ref{code:bootstrap_sampling}), and analytical derivation (Algorithm~\ref{code:analytical_derivation}).

\begin{algorithm*}[h]
\small
\caption{The Pseudo Code for Pass@k Training with Full Sampling.}
\label{code:full_sampling}
\SetKwInOut{Input}{Input}
\SetKwInOut{Output}{Output}

\Input{A tensor of reward $\mathcal{R}\in \mathbb{R}^{N_\text{rollout}}$ of the responses for the problem, the number of the rollouted responses $N_\text{rollout}$, and the $k$ for Pass@k metric.}
\Output{A tensor of estimated advantages of the responses for this problem $\hat{\mathcal{A}}\in \mathbb{R}^{N_\text{rollout}}$.}

\BlankLine
\BlankLine
\texttt{\# Construct the groups and discard the redundant instances.} \\
Separate $\mathcal{R}\in \mathbb{R}^{N_\text{rollout}}$ into $\lfloor \frac{N_\text{rollout}}{K} \rfloor$ group and each group contains $k$ instances. \\
Compute the reward of the groups $\mathcal{R}^\text{group}\in \mathbb{R}^{\lfloor \frac{N_\text{rollout}}{K} \rfloor}$ using Eq.~\ref{eq:passk}. \\

\BlankLine
\BlankLine
\texttt{\# Follow GRPO advantage estimation method to compute group-relative advantage.}\\
Compute the average reward of the groups $\bar{R}^\text{group}$ using Eq.~\ref{eq:grpo_avg}. \\
Compute the standard deviation of the groups $\sigma^\text{group}$ using Eq.~\ref{eq:grpo_std}. \\
Based on $\bar{R}^\text{group}$ and $\sigma^\text{group}$, compute the group-relative advantage $\hat{\mathcal{A}}^\text{group}$ using Eq.~\ref{eq:grpo_adv}. \\

\BlankLine
\BlankLine
\texttt{\# Compute response-relative advantage.} \\
Assign the $\hat{\mathcal{A}}^\text{group}$ to the responses that the group contains, obtaining the response-relative advantage $\hat{\mathcal{A}}$.

\end{algorithm*}

\begin{algorithm*}[h]
\small
\caption{The Pseudo Code for Pass@k Training with Bootstrap Sampling.}
\label{code:bootstrap_sampling}
\SetKwInOut{Input}{Input}
\SetKwInOut{Output}{Output}

\Input{A tensor of reward $\mathcal{R}\in \mathbb{R}^{N_\text{rollout}}$ of the responses for the problem, the number of the rollouted responses $N_\text{rollout}$, and the $k$ for Pass@k metric.}
\Output{A tensor of estimated advantages of the responses for this problem $\hat{\mathcal{A}}\in \mathbb{R}^{N_\text{rollout}}$.}

\BlankLine
\BlankLine

\texttt{\# Construct the groups through bootstrap sampling.} \\
\For{$i$ from $1$ to $N^\text{group}$}{
    Randomly sample $k$ instances from $\mathcal{R}$ to construct the $i$-th group. \\
    Compute the reward of $i$-th group using Eq.~\ref{eq:passk}. \\
}
Obtain the reward of the groups $\mathcal{R}^\text{group}\in \mathbb{R}^{N^\text{group}}$.\\

\BlankLine
\BlankLine
\texttt{\# Follow GRPO advantage estimation method to compute group-relative advantage.}\\
Compute the average reward of the groups $\bar{R}^\text{group}$ using Eq.~\ref{eq:grpo_avg}. \\
Compute the standard deviation of the groups $\sigma^\text{group}$ using Eq.~\ref{eq:grpo_std}. \\
Based on $\bar{R}^\text{group}$ and $\sigma^\text{group}$, compute the group-relative advantage $\hat{\mathcal{A}}^\text{group}$ using Eq.~\ref{eq:grpo_adv}. \\

\BlankLine
\BlankLine
\texttt{\# Calculate response-relative advantage.} \\
Based on $\hat{\mathcal{A}}^\text{group}$, compute response-relative advantage $\hat{\mathcal{A}}$ using Eq.~\ref{eq:bootstrap_sampling_group_to_response}.

\end{algorithm*}

\begin{algorithm*}[h]
\small
\caption{The Pseudo Code for Pass@k Training with Analytical Derivation.}
\label{code:analytical_derivation}
\SetKwInOut{Input}{Input}
\SetKwInOut{Output}{Output}

\Input{A tensor of reward $\mathcal{R}\in \mathbb{R}^{N_\text{rollout}}$ of the responses for the problem, the number of the rollouted responses $N_\text{rollout}$, and the $k$ for Pass@k metric.}
\Output{A tensor of estimated advantages of the responses for this problem $\hat{\mathcal{A}}\in \mathbb{R}^{N_\text{rollout}}$.}

\BlankLine
\BlankLine

\texttt{\# Calculate the average and standard deviation of the group reward scores.} \\
Compute the average reward of the groups $\bar{R}^\text{group}$ using Eq.~\ref{eq:group_avg}. \\
Compute the standard deviation of the groups $\sigma^\text{group}$ using Eq.~\ref{eq:group_std}. \\

\BlankLine
\BlankLine
\texttt{\# Calculate response-relative advantage.}\\
Compute the advantage of the positive responses $\hat{A}_{\text{pos}}$ using Eq.~\ref{eq:response_pos}. \\
Compute the advantage of the negative responses $\hat{A}_{\text{neg}}$ using Eq.~\ref{eq:response_neg}. \\
Based on $\hat{A}_{\text{pos}}$, $\hat{A}_{\text{neg}}$, and $\mathcal{R}$, assign the advantage to each instance, obtaining response-relative advantage $\hat{\mathcal{A}}$. 

\end{algorithm*}

\section{Curves of Advantage Function}

\begin{figure}[h]
    \centering
    \subfloat[Pass@k Training w/o Easy Problems.]{
        \includegraphics[width=0.4\linewidth]{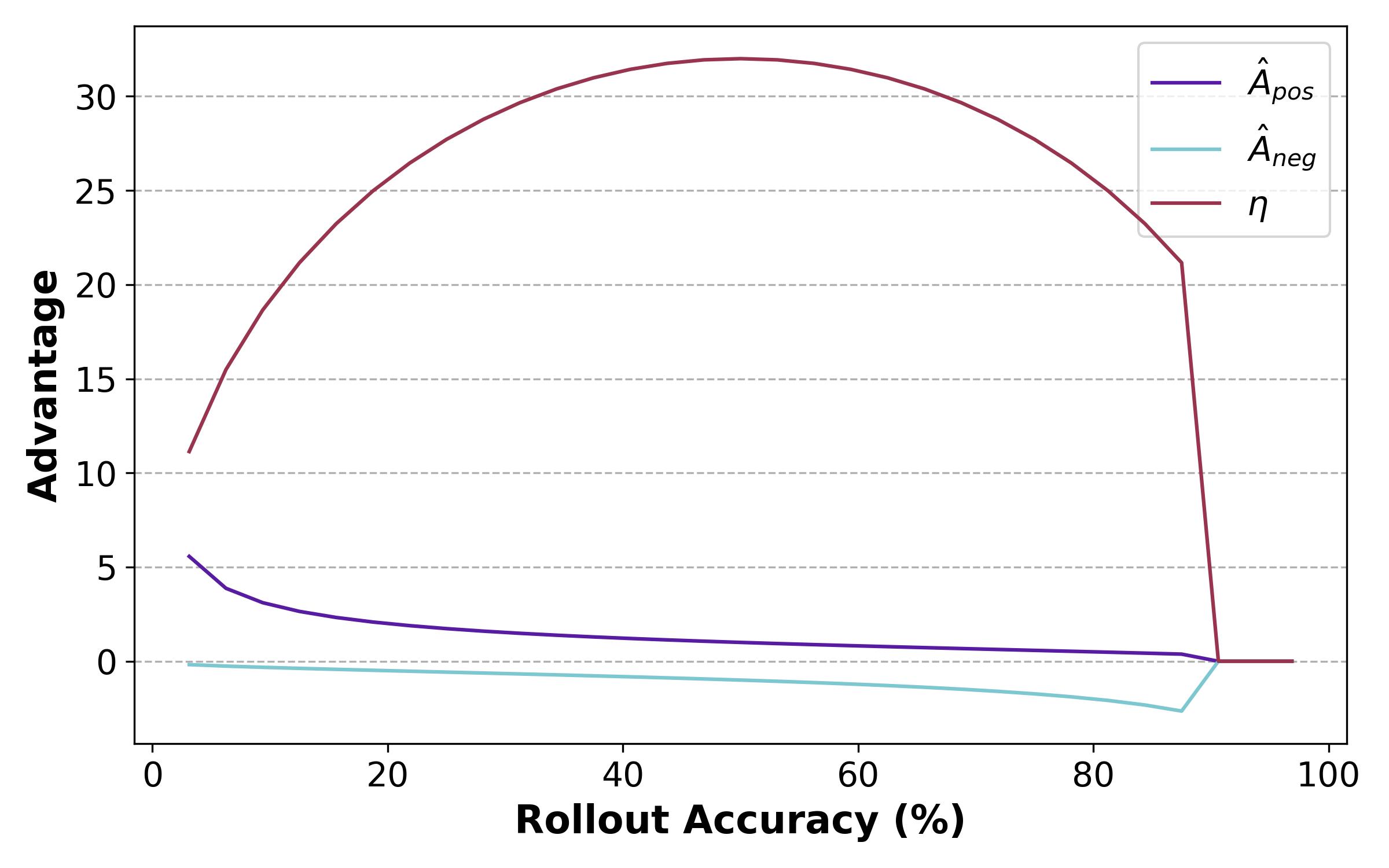}
        \label{fig:adv-wo_easy}
    }
    \subfloat[Pass@k Training w/ Combination.]{
        \includegraphics[width=0.4\linewidth]{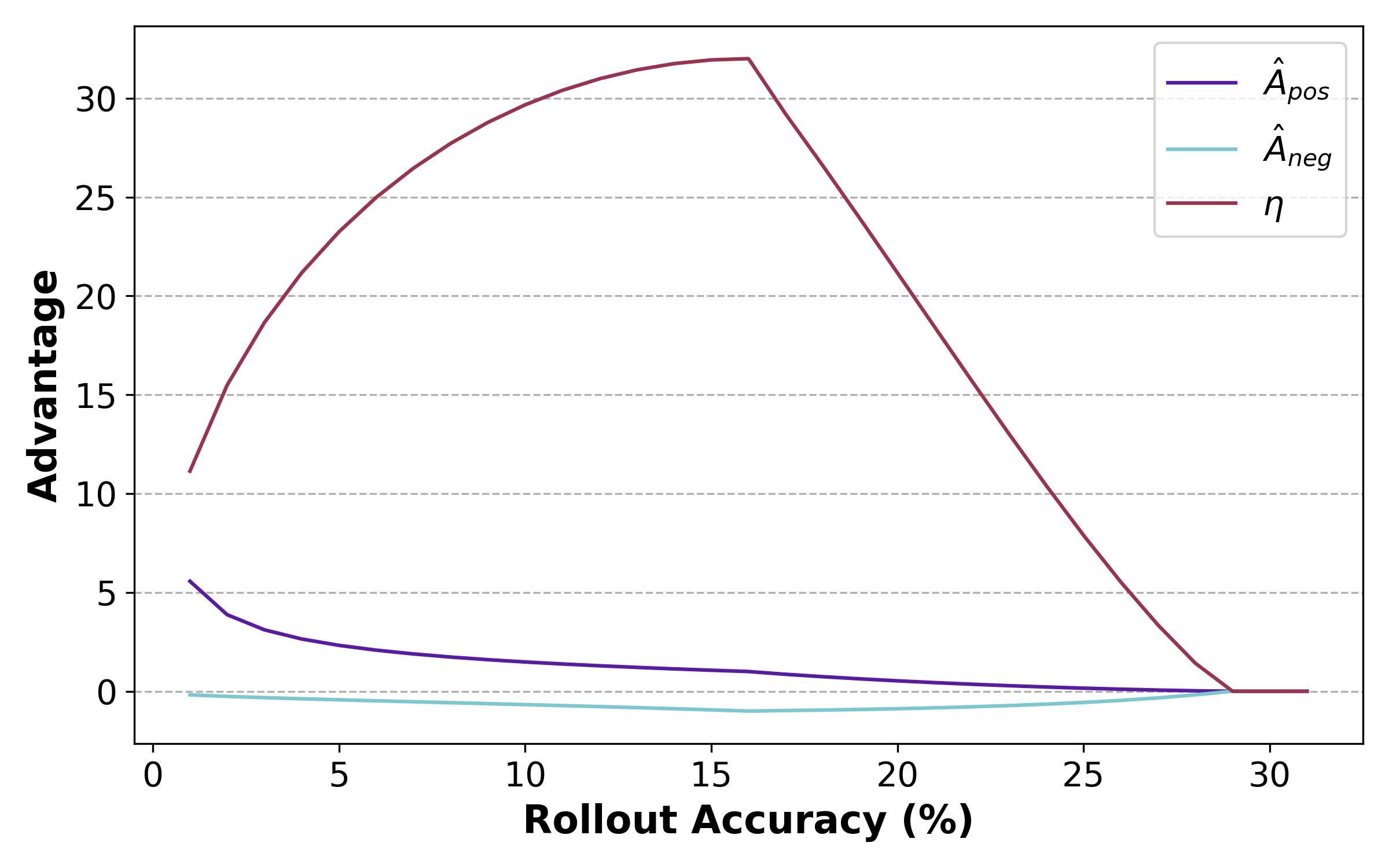}
        \label{fig:adv-segment}
    }
    \\
    \subfloat[Exceeding Pass@k Training.]{
        \includegraphics[width=0.4\linewidth]{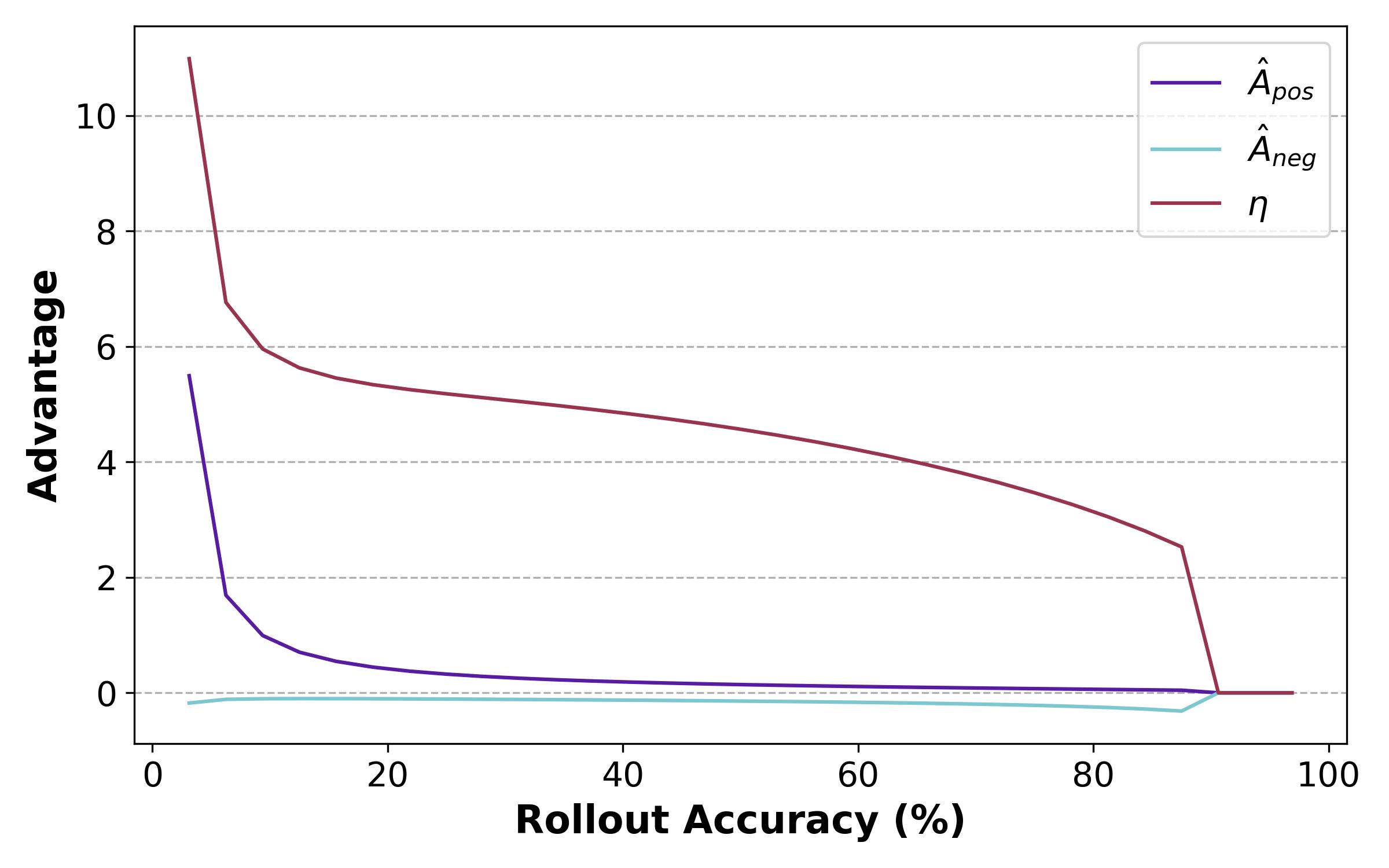}
        \label{fig:adv-log}
    }
    \subfloat[Combination Training.]{
        \includegraphics[width=0.4\linewidth]{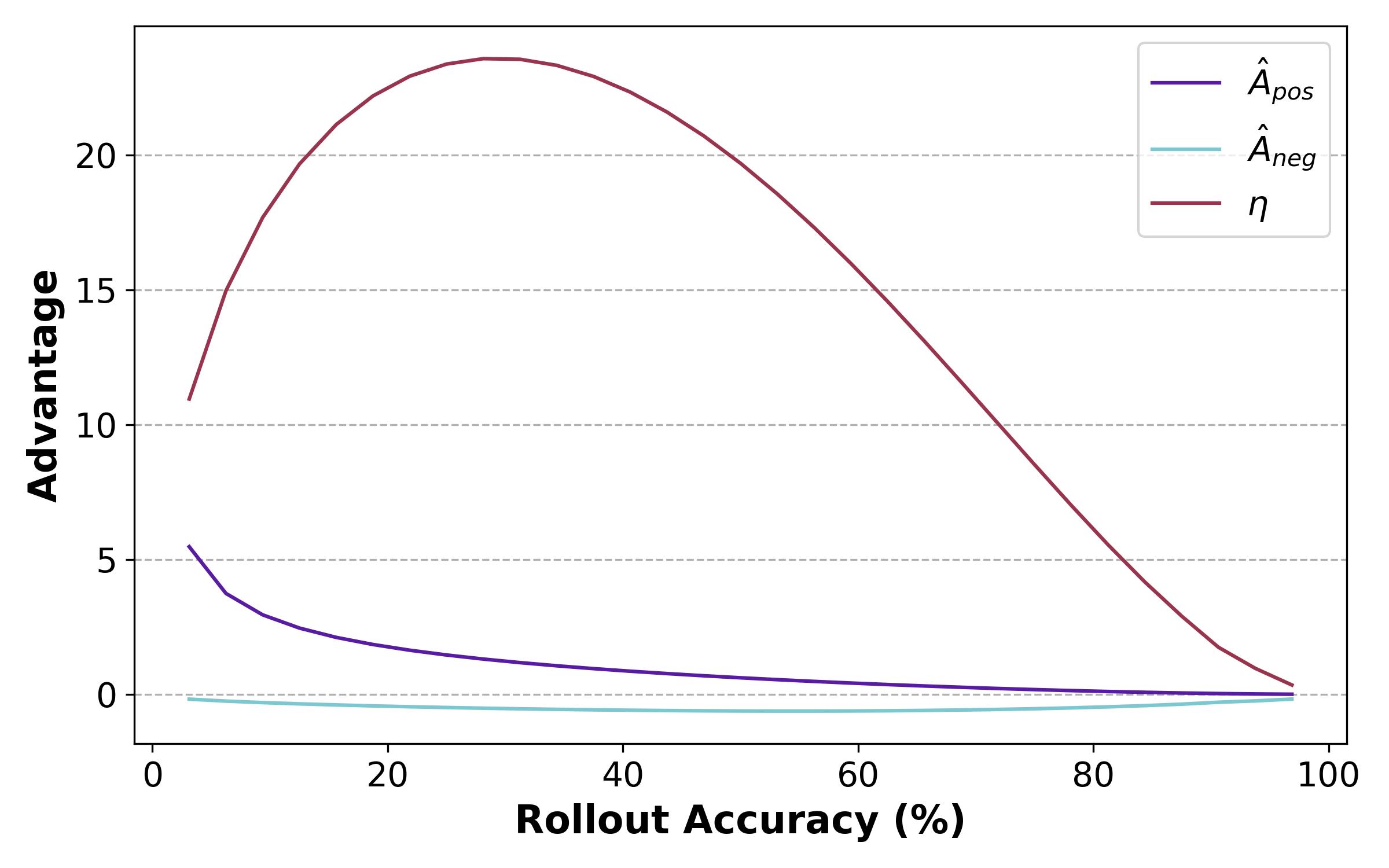}
        \label{fig:adv-comb}
    }
    \caption{The curves of the advantage function on the setting of $N_\text{rollout}=32$.}
    \label{fig:app-adv}
\end{figure}

We present the curves of the advantage function of different training approaches in Figure~\ref{fig:app-adv}, including Pass@k Training w/o easy problems, Pass@k Training w/ combination, exceeding Pass@k Training, and combination training.

\section{Experiments on Various LLMs and Tasks}
\label{sec:app_exp}

In this section, to further verify the effectiveness of Pass@k Training, we provide the performance of various LLMs trained through Pass@k Training on Mathematical Tasks (\ie AIME 2024~\cite{70:aime24}, AIME 2025~\cite{33:aime25}, and OlymMATH~\cite{71:journals/corr/abs-2503-21380}) and Synthetic Puzzle Task (\ie Enigmata~\cite{30:journals/corr/abs-2505-19914}).

\subsection{Pass@k Training on Mathematical Tasks}
We follow the experiment settings described in Appendix~\ref{sec:implementation_details} to perform Pass@k Training on LLaMA models~\cite{37:journals/corr/abs-2407-21783} (\ie LLaMA3.2-3B-Instruct and LLaMA3.1-8B-Instruct) and DeepSeek-R1-Distill-Qwen~\cite{7:journals/corr/abs-2501-12948} (\ie 1.5B and 7B version). For LLaMA models, we set the maximum prompt length and response length as 2048 and 6144, respectively. For DeepSeek-R1-Distill-Qwen, we extend the response length to 10240.
Specifically, to adapt the LLMs to the mathematical tasks, we adopt the training data used in previous work~\cite{8:journals/corr/abs-2503-04548} during the RLVR training process.
Besides, we follow the settings in Appendix~\ref{sec:implementation_details} to perform the evaluation, and present the results in Table~\ref{tab:app_math}.
Since the single turn of Pass@k Training followed by Pass@1 Training can significantly improve the Pass@1 performance of LLMs, we conduct the experiment about multiple turns of the above training process in Table~\ref{tab:app_math}, named as ``(P@k T. + P@1 T.) $\times$ 2''.

\begin{table}[h]
    \centering
    \small
    \setlength{\tabcolsep}{4pt}
    \caption{Pass@1/Pass@k Performance on mathematical tasks of LLaMA and DeepSeek-R1-Distill-Qwen models trained through different RLVR approaches. ``P@1 T.'' and ``P@k T.'' denote the Pass@1 Training and Pass@k Training with analytical derivation, respectively. ``(P@k T. + P@1 T.) $\times$ 2'' refers to that the process of Pass@k Training followed by Pass@1 Training is repeated twice.}
      \begin{tabular}{lccccc}
      \toprule
       & \textbf{AIME 2024} & \textbf{AIME 2025} & \textbf{OlymMATH-Easy} & \textbf{OlymMATH-Hard} & \textbf{Avg.} \\
      \midrule
      \multicolumn{6}{c}{\textbf{RLVR on LLaMA3.2-3B-Instruct (Pass@1/Pass@k)}} \\
      Baseline & 1.5/17.3 & 0.1/2.1 & 1.7/14.4 & 1.1/9.2 & 1.1/10.8 \\
      + P@1 T. & 13.6/26.7 & 1.1/6.6 & 3.8/4.0 & 2.0/6.3 & 5.1/10.9 \\
      + P@k T. & 12.7/32.0 & 1.7/12.9 & 3.7/8.8 & 1.7/7.7 & 5.0/\textbf{15.4} \\
      + P@k T. + P@1 T. & 14.6/32.1 & 1.3/8.6 & 4.1/7.7 & 2.0/7.5 & \textbf{5.5}/14.0 \\
      \midrule
      \multicolumn{6}{c}{\textbf{RLVR on LLaMA3.1-8B-Instruct (Pass@1/Pass@k)}} \\
      Baseline & 3.4/17.9 & 0.2/4.3 & 0.8/7.5 & 0.5/7.6 & 1.0/9.3 \\
      + P@1 T. & 4.4/32.1 & 0.9/7.7 & 1.4/4.1 & 1.1/6.2 & 2.0/12.5 \\
      + P@k T. & 7.1/40.0 & 1.8/10.6 & 1.5/8.9 & 1.4/8.2 & 3.0/\textbf{17.0} \\
      + P@k T. + P@1 T. & 8.7/29.7 & 0.9/8.7 & 1.8/7.9 & 1.6/6.8 & \textbf{3.3}/13.3 \\
      \midrule
      \multicolumn{6}{c}{\textbf{RLVR on DeepSeek-R1-Distill-Qwen-1.5B (Pass@1/Pass@k)}} \\
      Baseline & 22.7/61.4 & 20.5/37.2 & 6.6/36.7 & 0.6/5.2 & 12.6/35.1 \\
      + P@1 T. & 36.7/76.0 & 28.8/49.4 & 16.7/51.7 & 2.5/17.5 & 21.2/48.7 \\
      + P@k T. & 36.5/79.3 & 27.0/55.5 & 17.6/59.3 & 2.4/17.4 & 20.9/52.9 \\
      + P@k T. + P@1 T. & 42.3/71.7 & 30.4/57.8 & 20.7/60.5 & 3.4/18.9 & 24.2/52.2 \\
      + (P@k T. + P@1 T.) $\times$ 2 & 44.2/77.2 & 31.5/57.7 & 22.6/62.7 & 4.4/21.2 & \textbf{25.7}/\textbf{54.7} \\
      \midrule
      \multicolumn{6}{c}{\textbf{RLVR on DeepSeek-R1-Distill-Qwen-7B (Pass@1/Pass@k)}} \\
      Baseline & 43.2/80.9 & 31.5/59.1 & 22.2/66.0 & 1.4/13.7 & 24.6/54.9 \\
      + P@1 T. & 48.5/79.5 & 35.5/59.4 & 27.9/69.1 & 3.1/22.5 & 28.8/57.6 \\
      + P@k T. & 48.2/80.9 & 36.5/66.7 & 28.1/72.7 & 3.3/23.3 & 29.0/\textbf{60.9} \\
      + P@k T. + P@1 T. & 50.3/81.0 & 39.3/61.9 & 32.3/68.9 & 3.5/22.7 & \textbf{31.4}/58.6 \\
      \bottomrule
      \end{tabular}
      \label{tab:app_math}
\end{table}

\subsection{Pass@k Training on Enigmata Task}
We follow the experiment settings described in Appendix~\ref{sec:implementation_details} to perform Pass@k Training on various LLMs (\ie LLaMA3.2-3B-Instruct~\cite{37:journals/corr/abs-2407-21783} and LLaMA3.1-8B-Instruct~\cite{37:journals/corr/abs-2407-21783}), and set the maximum of the prompt length and response length as 4096 and 4096, respectively. The results are presented in Table~\ref{tab:app_enigmata}.
For evaluation, we follow the settings described in Appendix~\ref{sec:implementation_details}.

\begin{table}[h]
    \centering
    \small
    \setlength{\tabcolsep}{4pt}
    \caption{Enigmata Pass@1/Pass@k Performance of LLaMA models trained on different RLVR approaches. ``P@1 T.'' and ``P@k T.'' denote the Pass@1 Training and Pass@k Training with analytical derivation, respectively.}
      \begin{tabular}{lcccccccc}
      \toprule
       & \textbf{Crypto} & \textbf{Arithmetic} & \textbf{Logic} & \textbf{Grid} & \textbf{Graph} & \textbf{Search} & \textbf{Sequential} & \textbf{Overall} \\
      \midrule
      \multicolumn{9}{c}{\textbf{RLVR on LLaMA3.2-3B-Instruct (Pass@1/Pass@k)}} \\
      Baseline & 0.0/0.0 & 0.2/1.6 & 19.3/44.3 & 2.7/4.7 & 1.7/8.8 & 5.4/11.1 & 0.4/1.8 & 3.1/7.3 \\
      + P@1 T. & 0.0/0.0 & 0.2/1.3 & 19.7/27.0 & 17.4/18.0 & 5.3/12.8 & 12.9/14.5 & 9.8/10.7 & 11.1/13.0 \\
      + P@k T. & 0.0/0.0 & 0.2/0.7 & 22.0/31.0 & 17.3/18.1 & 6.1/14.2 & 12.2/16.6 & 10.7/12.0 & 11.5/\textbf{14.1} \\
      + P@k T. + P@1 T. & 0.0/0.0 & 0.4/2.2 & 22.8/27.7 & 16.7/17.4 & 6.5/13.5 & 14.6/16.0 & 12.0/13.0 & \textbf{12.3}/14.0 \\
      \midrule
      \multicolumn{9}{c}{\textbf{RLVR on LLaMA3.1-8B-Instruct (Pass@1/Pass@k)}} \\
      Baseline & 0.0/0.0 & 0.1/1.1 & 21.6/41.7 & 3.7/4.6 & 1.5/7.8 & 6.0/17.0 & 1.2/5.0 & 3.8/9.0 \\
      + P@1 T. & 0.0/0.0 & 0.1/0.9 & 29.2/38.0 & 12.1/13.2 & 3.8/8.5 & 12.1/14.7 & 5.3/7.4 & 8.7/11.1 \\
      + P@k T. & 0.0/0.0 & 0.1/0.9 & 30.5/39.3 & 12.9/14.8 & 5.5/12.2 & 12.2/14.7 & 7.5/10.9 & 9.9/13.0 \\
      + P@k T. + P@1 T. & 0.0/0.0 & 0.2/1.1 & 34.4/44.7 & 12.5/14.2 & 7.5/17.8 & 13.2/15.7 & 8.7/10.3 & \textbf{10.8}/\textbf{13.7} \\
      \bottomrule
      \end{tabular}
      \label{tab:app_enigmata}
\end{table}

\subsection{Pass@k Training on Maze Task}
\label{app:maze}

In this part, we present the full results of Pass@k Training on the Maze task in Table~\ref{tab:maze_results}.
Without any RLVR training, it is really difficult for the model to solve the Maze task.
Thus, we do not report the performance of the backbone model.

\begin{table}[h]
    \centering
    \small
    \setlength{\tabcolsep}{3.5pt}
    \caption{The Pass@1/Pass@k performance of Qwen2.5-7b-Instruct trained on different approaches on various Maze sizes. ``P@1 T.'' and ``P@k T.'' denote the Pass@1 Training and Pass@k Training with analytical derivation, respectively. ``FS'', ``BS'', and ``AD'' denote the full sampling, bootstrap sampling, and analytical derivation, respectively.}
      \begin{tabular}{lccccccccc}
      \toprule
      & {$7 \times 7$} & {$9 \times 9$} & {$11 \times 11$} & {$13 \times 13$} & {$15 \times 15$} & {$17 \times 17$} & {$19 \times 19$} & {$21 \times 21$} & Avg. \\
      \midrule
      + P@1 T. & 36.0/38.2 & 32.4/33.0 & 10.6/11.0 & 14.0/14.0 & 8.1/9.0 & 5.0/5.0 & 2.0/2.0 & 3.0/3.0 & 13.9/14.4 \\
      + P@k T. w/ FS & 34.6/67.4 & 26.4/47.6 & 13.7/26.0 & 11.0/18.5 & 8.6/17.6 & 3.0/7.6 & 2.2/7.9 & 1.9/5.6 & 12.7/24.7 \\
      + P@k T. w/ BS & 45.3/70.6 & 37.8/51.0 & 15.4/27.0 & 12.8/20.7 & 12.3/19.9 & 3.3/8.8 & 4.8/9.0 & 2.3/6.3 & 16.8/26.7 \\
      + P@k T. w/ AD & 86.8/98.2 & 94.6/100.0 & 75.2/98.3 & 55.2/84.6 & 39.2/72.0 & 10.5/29.2 & 16.7/28.3 & 3.5/9.7 & \textbf{47.7}/\textbf{65.0} \\
      \bottomrule
      \end{tabular}
      \label{tab:maze_results}
\end{table}

